%% file: paper.tex
\documentclass[letterpaper, 10 pt, conference]{ieeeconf}  

\IEEEoverridecommandlockouts                              

\usepackage{tikz}
\usepackage{pgfplots}
\usepackage{graphicx}
\usepackage{amsmath} 
\usepackage{amssymb}  
\usepackage{mathtools}
\usepackage{booktabs}
\usepackage{nicefrac}

\usetikzlibrary{calc}
\definecolor{myblue}{RGB}{21, 96, 130} 
\definecolor{mygreen}{RGB}{78, 167, 46} 
\definecolor{mygray}{RGB}{127, 127, 127}

\usetikzlibrary{arrows.meta}
\usetikzlibrary{backgrounds}
\usepgfplotslibrary{patchplots}
\usepgfplotslibrary{fillbetween}
\pgfplotsset{%
    layers/standard/.define layer set={%
        background,axis background,axis grid,axis ticks,axis lines,axis tick labels,pre main,main,axis descriptions,axis foreground%
    }{
        grid style={/pgfplots/on layer=axis grid},%
        tick style={/pgfplots/on layer=axis ticks},%
        axis line style={/pgfplots/on layer=axis lines},%
        label style={/pgfplots/on layer=axis descriptions},%
        legend style={/pgfplots/on layer=axis descriptions},%
        title style={/pgfplots/on layer=axis descriptions},%
        colorbar style={/pgfplots/on layer=axis descriptions},%
        ticklabel style={/pgfplots/on layer=axis tick labels},%
        axis background@ style={/pgfplots/on layer=axis background},%
        3d box foreground style={/pgfplots/on layer=axis foreground},%
    },
    compat=1.14
}
\include{def_color}

\newcommand{\bs}[1]{\boldsymbol{#1}}
\newcommand{\hT}{^\mathsf{T}}

\DeclareMathOperator*{\expectation}{\mathbb{E}}
\DeclareMathOperator*{\argmax}{arg\,max}
\DeclareMathOperator*{\Log}{Log}
\DeclareMathOperator*{\Exp}{Exp}

\title{\LARGE \bf
Reinforcement Learning with Lie Group Orientations for Robotics
}

\author{Martin Schuck, Jan Br\"udigam, Sandra Hirche, and Angela Schoellig
\thanks{All authors are with Technical University of Munich}
}

\begin{document}

\maketitle
\thispagestyle{empty}
\pagestyle{empty}

\begin{abstract}
Handling orientations of robots and objects is a crucial aspect of many applications. Yet, ever so often, there is a lack of mathematical correctness when dealing with orientations, especially in learning pipelines involving, for example, artificial neural networks. In this paper, we investigate reinforcement learning with orientations and propose a simple modification of the network's input and output that adheres to the Lie group structure of orientations. As a result, we obtain an easy and efficient implementation that is directly usable with existing learning libraries and achieves significantly better performance than other common orientation representations. We briefly introduce Lie theory specifically for orientations in robotics to motivate and outline our approach. Subsequently, a thorough empirical evaluation of different combinations of orientation representations for states and actions demonstrates the superior performance of our proposed approach in different scenarios, including: direct orientation control, end effector orientation control, and pick-and-place tasks.
\end{abstract}


\section{Introduction}
The orientation of robots, end effectors, or objects is a central state in almost all robotics applications, such as drone flying \cite{brunner2019urban, panerati2021learning, hanover2024autonomous}, highly dynamic locomotion \cite{hwangbo2019learning, yang2023cerberus}, or manipulation of objects \cite{andrychowicz2020learning, chen2022system, brudigam2024jacta}. While positions can be trivially described with Cartesian coordinates in Euclidean space, orientations require more attention. Since orientations form a mathematical group, they have favorable structural properties that can simplify and improve calculations when respected \cite{sola2017quaternion, sola2018micro}. However, in practice, this group structure is often only partially considered or even entirely ignored, leading to, for example, issues with singularities in Euler angles. Moreover, learning orientations with artificial neural networks often conflicts with the group structure because such networks typically operate in Euclidean space, i.e., $\mathbb{R}^n$, since most practical learning implementations provide full support only for such a representation \cite{ansel2024pytorch, tensorflow2015whitepaper}. Another complexity is that orientations can be expressed by many different representations, which have different advantages and disadvantages, such as computational speed, smoothness, multi-cover, and singularities \cite{kim2023rotation}.

This paper focuses on treating orientations mathematically consistently in reinforcement learning (RL). Our central claim is that in an RL setting, adhering to the Lie group structure of orientations where possible results in mathematically sound expressions and practically superior performance in learning progress, computational speed, and policy performance. Specifically, our contributions are:
\begin{itemize}
    \item A modification of network inputs and outputs in reinforcement learning based on the Lie algebra of orientations that is mathematically sound, practically efficient to implement, and leads to improved policy performance compared to other common implementations.
    \item A practical introduction to Lie theory for orientations in robotics to motivate and outline the modified network architecture.
    \item An empirical evaluation of different combinations of orientation representations for observations and actions to determine the most suitable representation for RL with orientations.
\end{itemize}

\begin{figure}[!tb]
    \centering
     \input{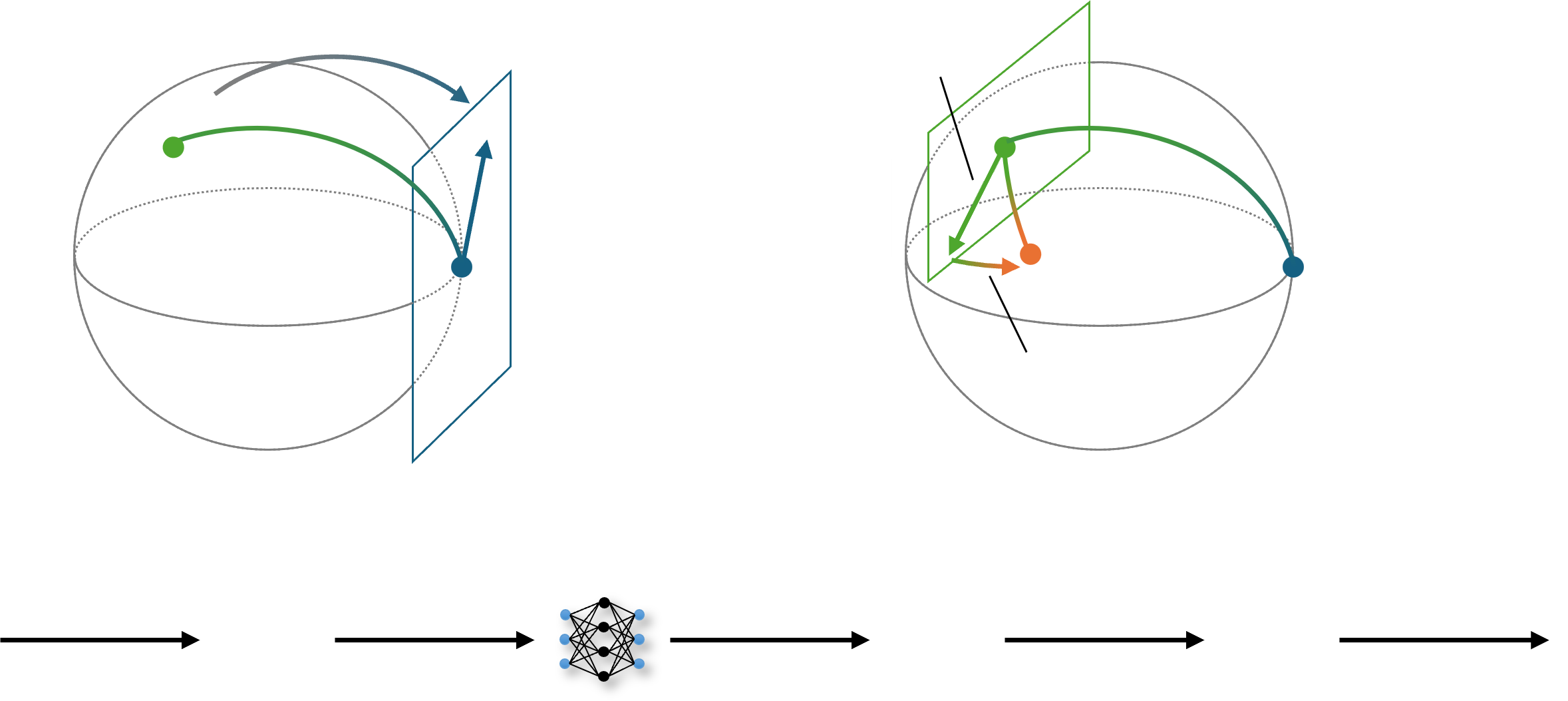}
     \vspace{-1.0\intextsep}
    \caption{\textbf{Top}: Our proposed learning architecture. Starting with an orientation state $\bs{s}\in\mathcal{M}$, we take the $\Log$ to obtain a vector $\prescript{\bs{\mathcal{E}}}{}{\bs{\tau}}_{\bs{s}}\in\mathbb{R}^3$ in the tangent space. This vector is passed into the neural network to obtain an action vector $\prescript{\bs{s}}{}{\bs{\tau}}_{\bs{a}}\in\mathbb{R}^3$ relative to $\bs{s}$. By taking the $\Exp$, we obtain a relative action $\bs{a}\in\mathcal{M}$ which is composed with the original state $\bs{s}$ to obtain the new state $\bs{s}'=\bs{s}\cdot\bs{a}\in\mathcal{M}$. \textbf{Bottom}: The hardware setup for the pick-and-place task. A cube is moved from an initial pose to a goal pose in the air.}
    \label{fig:nn_structure}
    \vspace{-1.0\intextsep}
\end{figure}

\subsection{Problem setting}
We consider reinforcement learning (RL) for a goal-conditioned Markov decision process with state $\bs{s}\in\mathcal{S}$, goal $\bs{g}\in\mathcal{G}_{\bs{s}}$, action $\bs{a}\in\mathcal{A}$, and sparse reward $r\in\{-1,0\}$. The state can be composed of positions and orientations. The actions are position or orientation commands relative to the current state, resulting in the dynamics $\bs{s}'=\bs{f}(\bs{s},\bs{a})$.

The objective in goal-conditioned RL is to find a policy $\bs{a} = \bs{\pi}^\star(\bs{s},\bs{g})$ that maximizes the expected future reward:
\begin{equation}
    \bs{\pi}^\star = \argmax_{\bs{\pi}}\expectation_{\bs{\pi}}\left[\sum_{t=0}^\infty \gamma^t r(\bs{s}_t,\bs{g}_t,\bs{\pi}(\bs{s}_t,\bs{g}_t))\right],
\end{equation}
with discount factor $\gamma\in[0,1)$.

For the reward, we consider the distance between the state and the goal. For Euclidean quantities, i.e., positions, the distance is the norm of the state-to-goal difference:
\begin{equation}
    d_\mathrm{E}(\bs{s},\bs{g}) = \lVert \bs{s} - \bs{g}\rVert\in\mathbb{R}.
\end{equation} 
For orientations, the distance is the norm of the orientation difference in the tangent space between state and goal orientations: 
\begin{equation}
    d_\mathcal{M}(\bs{s},\bs{g}) = \lVert \bs{s}\ominus \bs{g}\rVert\in\mathbb{R}.
\end{equation}
This distance is simply the angle between the two orientations, independent of the representation, and, therefore, comparable between different representations. We explain how this difference is computed in Sec. \ref{sec:lie_alg} and refer to \cite{sola2018micro} for further details and notation.

The sparse reward is 0 if the Euclidean distance is smaller than a threshold $\epsilon_\mathrm{E}\in\mathbb{R}^+$ and the orientation distance is smaller than a threshold $\epsilon_\mathcal{M}\in\mathbb{R}^+$, and $-1$ otherwise, i.e.,
\begin{equation}
    r(\bs{s},\bs{g},\bs{a}) = 
    \begin{cases}
      0 & \text{if } d_\mathrm{E}\leq\epsilon_\mathrm{E} ~\text{and}~ d_\mathcal{M}\leq\epsilon_\mathcal{M}\\
      -1 & \text{otherwise}
    \end{cases}.
\end{equation}

For our experiments, we use the deep deterministic policy gradient (DDPG) learning algorithm \cite{lillicrap2015continuous} with hindsight experience replay \cite{andrychowicz2017hindsight}.

\subsection{Related work}
The correct treatment of orientations in robotics is gaining increasing interest to improve performance. Learning methods based on artificial neural networks require careful attention due to their Euclidean structure in practice, which is not directly compatible with the manifold structure of orientations. Accordingly, \cite{zhou2019continuity} investigates which orientation representations exhibit discontinuous maps to the rotation group $SO(3)$ and how these discontinuities lead to theoretical and practical learning deficiencies. Continuing in this direction, \cite{geist2024learning} investigates supervised learning with orientations both at the input and output of networks. While both works address differences in rotation representations for supervised learning, they still treat the representations as Euclidean vectors, requiring, for example, the projection of the network output onto the orientation manifold. In contrast, we propose a method that adheres to the orientation structure both at the network input and output.

One approach to treat the orientation manifold structure correctly in reinforcement learning can be found in \cite{alhousani2023geometric}. Here, the focus is on orientations in the action space, which are treated as a Riemannian manifold at the network output. This view can be complex and, if not handled correctly, is subject to subtle issues described in detail in \cite{jaquier2024unraveling}. Initial ideas in \cite{alhousani2023reinforcement}, where instead of the Riemannian manifold view, the Lie group perspective is taken, simplify computations for orientations. However, the real-world experiments in \cite{alhousani2023reinforcement} are limited, and no study of the mutual influence of representation combinations in the action and observation space is considered. In our paper, we conduct a thorough study across many different parametrizations and show results for complex, real-world robotic tasks.

A fundamentally different approach is to adapt the entire network architecture to adhere to the group structure of orientations. Several authors have investigated such methods \cite{finzi2020generalizing, masci2015geodesic, chakraborty2020manifoldnet, li2018deep}, and there are a number of works specifically on quaternion representations \cite{garcia2020quaternion, parcollet2020survey}. However, since most state-of-the-art learning libraries have limited support for such network structures, we focus on modifying the network input and output to leverage existing packages and provide a practically beneficial method.

\section{Background on Lie theory for orientations}
Unlike the position of an object, its orientation is a non-Euclidean state. An intuitive example of this fact is that incrementally rotating an object will decrease the global orientation once the orientation exceeds $\pi$. Nonetheless, the set of all possible orientations can be associated with a mathematical group on a smooth manifold, which has the same shape everywhere, making it a Lie group. In this section, we often follow the excellent introduction to Lie groups for robotics by \cite{sola2018micro} and refer to this work for further details. In our paper, we only provide the background necessary for learning in robotics with orientations.

First, we provide representations that are commonly used to describe orientations. Then, we explain how distances between orientations can be computed and how relative orientations can be composed. Finally, we give insights into how these representations are used in learning with neural networks.

\subsection{Lie groups as orientation representations}
The orientation of an object is a geometric property. In order to work with orientations, we need a mathematical representation. Since there are different applications with distinct requirements, various orientation representations are commonly used, and a summary of common representations is given in Tab. \ref{tab:orientation_reps}.

\begin{table}[!htb]
\vspace{-0.0\intextsep}
\centering
\begin{tabular}{l c c c c}
 \toprule
  Representation & Structure & Size & Dim. & Cover \\
 \midrule
    Rotation matrices & Lie Group $SO(3)$ & 9 & 3 & single\\
    $3\times2$ matrices & Lie Group $SO_{1:2}(3)$ & 6 & 3 & single\\
    Quaternions & Lie Group $S^3$ & 4 & 3 & double\\
    Quaternions$^+$ & Group & 4 & 3 & single\\
    Euler angles & Vector space $\mathbb{R}^3$ & 3 & 3 & multi\\
    Axis-angle & Lie Algebra $\mathfrak{m}\cong \mathbb{R}^3$ & 3 & 3 & multi\\
 \bottomrule
\end{tabular}
 \caption{Orientation representations}\label{tab:orientation_reps}
\vspace{-1.0\intextsep}
\end{table}

Some of these representations are a group. Citing \cite{sola2018micro}, a group $(\mathcal{G}, \cdot)$ is a set, $\mathcal{A}$, with a composition operation, $\cdot$, that, for elements $\bs{x}, \bs{y}, \bs{z} \in \mathcal{G}$, satisfies the following axioms:
\begin{subequations}\label{eqn:group_ops}
    \begin{align}
        \text{Closure under $\cdot$}&:~~ \bs{x}\cdot \bs{y} \in \mathcal{G}\\
        \text{Identity $\bs{\mathcal{E}}$}&:~~ \bs{\mathcal{E}}\cdot \bs{x} = \bs{x}\cdot \bs{\mathcal{E}} = \bs{x}\\
        \text{Inverse $\bs{x}^{-1}$}&:~~ \bs{x}^{-1}\cdot \bs{x} = \bs{x}\cdot \bs{x}^{-1} = \bs{\mathcal{E}}\\
        \text{Associativity}&:~~ (\bs{x}\cdot \bs{y})\cdot z = \bs{x}\cdot (\bs{y}\cdot \bs{z})
    \end{align}
\end{subequations}
As mentioned above, if the elements of a group exist on a smooth manifold that looks the same everywhere, it is a Lie group. An example is the group of rotation matrices $SO(3)$ with the composition being matrix-matrix multiplication, the identity element being the identity matrix, $\bs{\mathcal{E}}=\bs{I}_{3\times3}$, and the inverse defined as the matrix transpose, $\bs{R}^{-1} = \bs{R}\hT$.

Since the $3\times3$ rotation matrices $\bs{R}\in SO(3)$ evolve smoothly on their manifold, they form a Lie group. There is a bijective (one-to-one) mapping between the elements of the $SO(3)$ group and the elements of the set of all possible orientations. Therefore, $SO(3)$ is typically referred to as \textit{the} orientation (or rotation) group.

A related and less common yet practically useful representation are the $3\times2$ matrices $\bs{R}\in SO_{1:2}(3)$ composed of the first two columns of the matrices in $SO(3)$ \cite{zhou2019continuity}. Given a matrix in $SO_{1:2}(3)$, the third column can be uniquely reconstructed from the first two with the cross product, making $SO_{1:2}(3)$ isomorphic (same structure but different representation) to $SO(3)$, i.e., $SO_{1:2}(3)\cong SO(3)$. The advantage of fewer parameters in the $SO_{1:2}(3)$ representation comes at the cost of losing direct matrix-matrix or matrix-vector multiplication.

Another way to represent orientations is with the computationally efficient four-parameter unit quaternions $\bs{q}\in S^3$, with quaternions $\bs{q}\in\mathbb{H}$, $\lVert \bs{q} \rVert = 1$. Unit quaternions also form a Lie group, but there is a surjective (here many-to-one) mapping from $S^3$ to $SO(3)$. Specifically, there is a double cover, i.e., both $q$ and $-q$ represent the same orientation. Yet, the group of unit quaternions, as well as unit-quaternion distances and incremental changes, can be formulated smoothly.

There are also orientation representations that are not Lie groups. In order to avoid non-uniqueness issues with the double cover, unit quaternions can be limited to a single cover $S^{3+}$ by negating quaternions with negative real part. While $S^{3+}$ still satisfies the group axioms \eqref{eqn:group_ops} by modifying the composition operation to always negate negative-real-part quaternions, this modification breaks the Lie group structure because of the loss of a smooth manifold.

\begin{figure}[!tb]
\vspace{-0.0\intextsep}
    \centering
    \input{figures/orientation_comp.tex}
     \vspace{-1.0\intextsep}
    \caption{A comparison of continuous and discontinuous orientation representations. \textbf{Top}: the initial frame (1) is first turned $-90^\circ$ around the local $y$-axis to obtain frame (2), and then turned $90^\circ$ around the local $z$-axis to obtain frame (3). \textbf{Bottom}: Representing this frame transformation with a rotation matrix or a quaternion (with double cover) evolves continuously on their respective manifolds. Using Euler angles results in a discontinuity at the singularity.}
    \label{fig:orientation_comp}
    \vspace{-1.0\intextsep}
\end{figure}
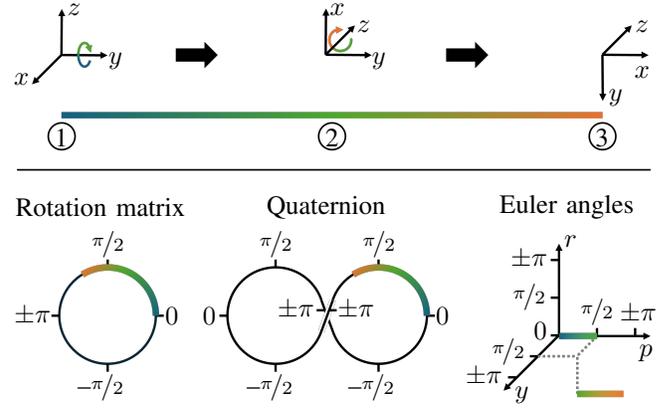

Another widespread representation is three-parameter Euler angles $\bs{e}\in\mathbb{R}^3$. Euler angles have a non-smooth mapping to orientations at specific configurations, i.e., singularities. In such configurations, a small local change of orientation can only be achieved by a discontinuity in Euler angles, as shown in Fig. \ref{fig:orientation_comp}. Euler angles also have a surjective mapping to orientations, where in singularities, infinitely many Euler angles represent the same orientation.

Yet another representation is axis-angle $\bs{\theta}\in\mathbb{R}^3$. This representation builds on the fact that any representation can be expressed as a three-dimensional unit-norm rotation axis $\bs{u}\in\mathbb{R}^3$ and a rotation angle $\theta\in\mathbb{R}$, resulting in $\bs{\theta}=\theta \bs{u}$. There is a surjective mapping from axis-angle to orientations with a multi-cover. A different perspective is to consider this representation as an isomorphism to the Lie algebras for the Lie groups $SO(3)$, $SO_{1:2}(3)$, and $S^3$, which we discuss in the next section.

\subsection{Lie algebra for distances and increments}\label{sec:lie_alg}
\begin{figure}[!tb]
\vspace{-0.0\intextsep}
    \centering
     \input{figures/group_tangent}
    \caption{Visualization of a Lie group on manifold $\mathcal{M}$, its Lie algebra $\mathfrak{m}$, i.e., the tangent space at the identity $T_{\bs{\mathcal{E}}}\mathcal{M}$, and a tangent space at $T_{\bs{x}}\mathcal{M}$ at $\bs{x}\in\mathcal{M}$, where the following holds: $\bs{y} = \bs{x}\cdot\Exp(\prescript{\bs{x}}{}{\bs{\tau}})$.}
    \label{fig:group_tangent}
    \vspace{-1.0\intextsep}
\end{figure}
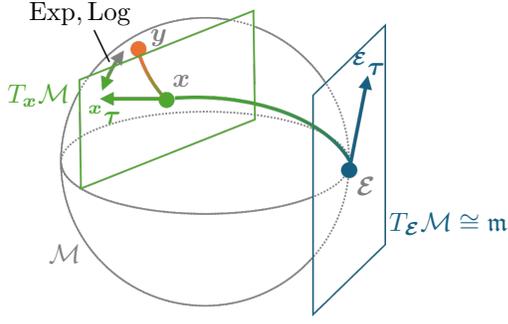

Every Lie group has an associated Lie algebra closely related to the group. The Lie algebra is a vector space tangent to the group at the identity element, and its elements can be represented by vectors. That is, the Lie algebra $\mathfrak{m}$ is the tangent space $T_{\bs{\mathcal{E}}}\mathcal{M}$ at the identity $\bs{\mathcal{E}}$ of the Lie group $\mathcal{M}$. The relationship between Lie group and algebra is visualized in Fig. \ref{fig:group_tangent}. Elements $\bs{\tau}^\wedge\in\mathfrak{m}$ of the Lie algebra $\mathfrak{m}$ can be uniquely represented by vectors $\bs{\tau}\in\mathbb{R}^m$, where $m$ is the dimension of the group, i.e., $\mathfrak{m}\cong \mathbb{R}^m$. Table \ref{tab:lie_alg} contains orientation Lie groups, Lie algebras, and the Lie algebra vector representations.

\begin{table}[!htb]
\vspace{-0.0\intextsep}
\centering
\begin{tabular}{l c c}
 \toprule
  Lie Group $\mathcal{M}$ & Lie Algebra $\mathfrak{m}$ & Vector Representation $\mathbb{R}^m$ \\
 \midrule
    $SO(3)$ & $\bs{\theta}^\times\in\mathfrak{so}(3)$ & $\bs{\theta}\in\mathbb{R}^3$\\
   $SO_{1:2}(3)$ & $\cong\bs{\theta}^\times\in\mathfrak{so}(3)$ & $\bs{\theta}\in\mathbb{R}^3$\\
   $S^3$ & $\bs{\theta}/2\in\mathbb{H}_\mathrm{p}$ & $\bs{\theta}\in\mathbb{R}^3$\\
 \bottomrule
\end{tabular}
 \caption{Lie groups and algebras}\label{tab:lie_alg}
 \vspace{-1.0\intextsep}
\end{table}
The $^\times$ operation in $\bs{\theta}^\times$ creates a $3\times3$ skew-symmetric matrix from $\bs{\theta}$. For orientation Lie groups such as $SO(3)$, $SO_{1:2}(3)$, and $S^3$, the elements of the Lie algebra represented as vectors are axis-angles $\bs{\theta}\in\mathbb{R}^3$. 

There exists a simple relation between a Lie group element $\bs{x}\in\mathcal{M}$ and its associated Lie algebra element $\bs{\tau}^\wedge\in\mathfrak{m}$ (and $\bs{\tau}\in\mathbb{R}^m$):
\begin{subequations}\label{eqn:lie_map}
    \begin{align}
        \bs{x} &= \exp(\bs{\tau}^\wedge) = \Exp(\bs{\tau}),\\
        \bs{\tau}^\wedge &= \log(\bs{x}),\\
        \bs{\tau} &= \Log(\bs{x}).
    \end{align}
\end{subequations}

The Lie algebra allows us to easily compose relative orientations and compute orientation differences. An incremental orientation $\prescript{\bs{x}}{}{\bs{\tau}}\in T_{\bs{x}}\mathcal{M}$ expressed in the local tangent space at the orientation $\bs{x}\in\mathcal{M}$ is achieved by the following composition:
\begin{equation}
    \bs{y} = \bs{x}\oplus\prescript{\bs{x}}{}{\bs{\tau}} = \bs{x}\cdot\Exp(\prescript{\bs{x}}{}{\bs{\tau}}).
\end{equation}
Because of the group composition and exponential map, adding a relative orientation change remains on the group manifold and does not require any projection. Similarly, the difference between two orientations can be computed as
\begin{equation}
    \prescript{\bs{x}}{}{\bs{\tau}} = \bs{y}\ominus\bs{x} = \Log(\bs{x}^{-1}\bs{y}).
\end{equation}

With the group operations \eqref{eqn:group_ops} and the Lie algebra, we can also define a distance metric $d:\mathcal{M}\times\mathcal{M}\rightarrow\mathbb{R}$ with the following properties for elements $\bs{x}, \bs{y}, \bs{z} \in \mathcal{M}$:
\begin{subequations}\label{eqn:distance_props}
    \begin{align}
        \text{Zero self distance}&:~~ d(\bs{x},\bs{x}) = 0\\
        \text{Positive for $\bs{x}\neq \bs{y}$}&:~~ d(\bs{x},\bs{y}) > 0\\
        \text{Symmetry}&:~~ d(\bs{x},\bs{y}) = d(\bs{y},\bs{x})\\
        \text{Triangle inequality}&:~~ d(\bs{x},\bs{z})\leq d(\bs{x},\bs{y})+d(\bs{y},\bs{z}).
    \end{align}
\end{subequations}
Defining such a metric is not trivial: if we want to know the distance of an orientation to the origin, i.e., the identity, we cannot simply take the norm of the orientation representation element as we would do for Euclidean vectors since, for example, any rotation matrix, including the identity matrix, has a norm $\lVert \bs{R}\rVert_\mathrm{F}=\sqrt{3}$. Accordingly, several different metrics exist that measure the distance between orientations \cite{huynh2009metrics}. However, we consider the geodesic distance to be the geometrically ``correct'' metric. The geodesic distance is simply the rotation angle between two orientations. 

The Lie algebra is an axis-angle representation and can be used directly as a distance measure. As a result, the distance to the origin for an orientation $\bs{x}\in\mathcal{M}$ is the norm of its Lie algebra element: $\lVert \log(\bs{x}) \rVert = \lVert \bs{\theta} \rVert = \theta$. The distance between two orientations $\bs{x},\bs{y}\in\mathcal{M}$ is the distance of the relative orientation between the two to the origin: $\lVert\log(\bs{y}^{-1}\bs{x})\rVert$.

Since Euler angles do not constitute a Lie group, we treat them as a vector space and compose orientations by adding Euler angles. Distances are computed by first converting Euler angles to a Lie group representation and then computing the Lie algebra distance.

\section{Learning with orientations}
Neural networks can be interpreted as learning a nonlinear feature representation for an input and producing a weighted output of these features. Most network architectures operate in Euclidean vector spaces, i.e., they have vector inputs and outputs, as well as addition and scalar multiplication as operations. Accordingly, the feature representation for orientations happens in Euclidean space by representing orientation representations as Euclidean vectors at the inputs and outputs.

When using Lie group representations and directly passing them into a network, they are mathematically incorrectly treated as vectors at the input, for example, by flattening a rotation matrix $\bs{R}_\mathrm{in}\in SO(3)$ into a vector $\bs{r}_\mathrm{in}\in\mathbb{R}^9$. More problematically, the network's output is also a vector and not on the orientation manifold $SO(3)$. Therefore, some form of projection back onto the manifold is required, for example, based on singular-value decomposition, to obtain the ``closest'' rotation matrix $\bs{R}_\mathrm{out}\in SO(3)$ from a non-$SO(3)$ matrix $\bs{R}_\mathrm{out}'\in\mathbb{R}^{3\times3}$.

As an alternative, we propose to properly consider the Lie group structure of orientations when learning with neural networks. To this end, instead of directly feeding an orientation state $\bs{s}$ as a vector at the input, we pass in the associated Lie algebra element $\prescript{\bs{\mathcal{E}}}{}{\bs{\tau}}_{\bs{s}} = \Log(\bs{s})\in\mathbb{R}^3$ which is, in fact, a vector. The network then performs operations that are within the algebraic structure of the Lie algebra and also provides an output that is a Lie algebra element $\prescript{\bs{\bs{s}}}{}{\bs{\tau}}_{\bs{a}}\in\mathbb{R}^{3}$. An orientation action is recovered from this element by taking the exponential $\bs{a} = \Exp(\prescript{\bs{\bs{s}}}{}{\bs{\tau}}_{\bs{a}})\in\mathcal{M}$. Finally, the new state $\bs{s}'$ is obtained by composing the current state $\bs{s}$ and action $\bs{a}$: $\bs{s}'=\bs{s}\cdot\bs{a}\in\mathcal{M}$. The full process is visualized in Fig. \ref{fig:nn_structure}.

In practice, the exponential and logarithm can either be manually implemented in closed form (cf. \cite{sola2018micro}) or by using the matrix and quaternion exponential and logarithm existing in most linear algebra packages.

\begin{figure*}[!htb]
    \centering
    \input{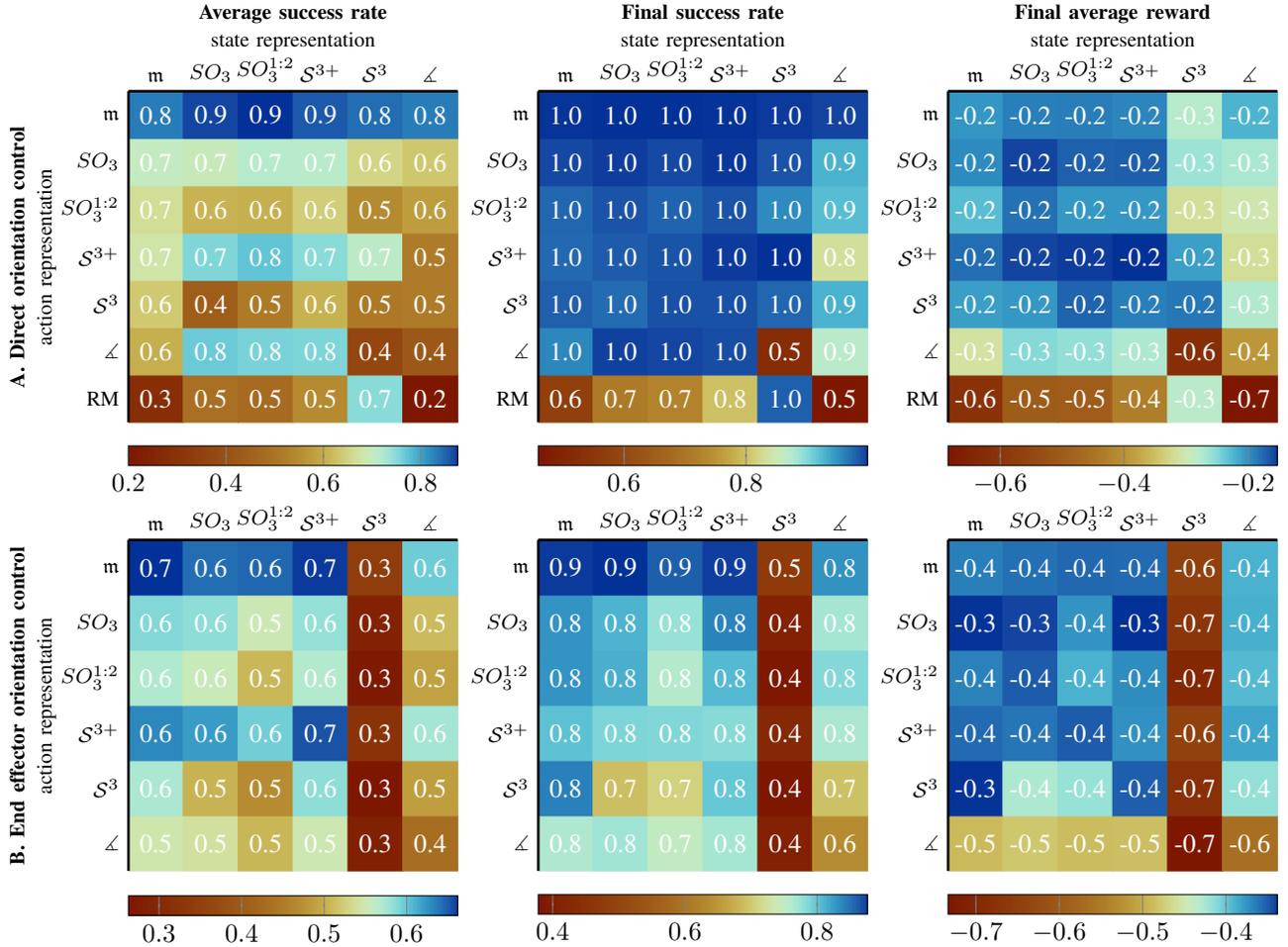}
    \vspace{-1.0\intextsep}
    \caption{Comparison of different orientation representations for state and action: Lie algebra $\mathfrak{m}$, rotation matrices $SO(3)$ ($SO_3$), two-column rotation matrices $SO_{1:2}(3)$ ($SO_3^{1:2}$), positive-real-part quaternions $\mathcal{S}^{3+}$, quaternions $\mathcal{S}^3$, Euler angles $\measuredangle$, and Riemannian manifold action RM. \textbf{Top row}: Results for direct orientation control. \textbf{Bottom row}: Results for end effector orientation control. \textbf{Left column}: Average success rate during training to measure convergence speed and overall success. Higher (blue) is better. \textbf{Center column}: Final success rate to measure best task success. Higher (blue) is better. \textbf{Right column}: Average reward per step of the final policy to measure policy performance. Closer to zero (blue) is better.}
    \label{fig:results}
    \vspace{-1.0\intextsep}
\end{figure*}

\section{Experiments}
We investigate three settings with increasing complexity to provide empirical results on reinforcement learning with orientations:
\begin{enumerate}
    \item[A.] Directly learning a policy for orienting a frame with incremental rotation actions without any robot embodiment to obtain a ``clean'' baseline.
    \item[B.] Moving the end effector of a robot arm to a desired orientation to investigate the learning of embodied orientation.
    \item[C.] Picking and placing a cube with the robot arm where the orientation of the unactuated cube cannot be controlled directly.
\end{enumerate}

We use MuJoCo \cite{todorov2012mujoco} for the robot simulation, and DDPG for learning, although we repeated some of the experiments with TD3 for verification and obtained similar results. 

\subsection{Direct orientation control}\label{sec:doc}
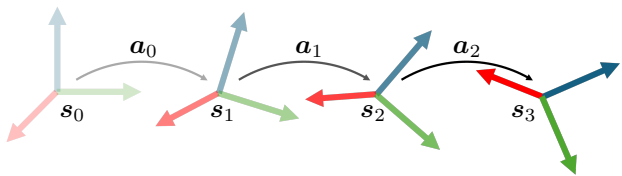
\begin{figure}[!b]
\vspace{-1.0\intextsep}
    \centering
    \input{figures/orientation_control}
    \vspace{-2.0\intextsep}
    \caption{Task progression for direct orientation control. From the initial state $\bs{s}_0$, relative rotation actions $\bs{a}_i$ are taken to move toward the goal (not shown).}
    \label{fig:orientation_control}
\end{figure}

We compare 36 orientation representation combinations by evaluating the six representations listed in Tab. \ref{tab:orientation_reps} for the state and action. The task is to rotate an initial frame into a goal frame by applying incremental rotation actions, visualized in Fig. \ref{fig:orientation_control}. The initial state $\bs{s}$ and goal $\bs{g}$ are sampled uniformly in $S^3$ and then converted to other representations. For all representations, the action is limited to a relative rotation angle of at most $0.1\pi$. We also repeated the experiments with $0.05\pi$ and $0.2\pi$ and obtained similar results. During each episode, the agent can take 50 steps to reach the goal. We train the agent for 200,000 environment steps.

We also compare to the Riemannian manifold approach in \cite{alhousani2023geometric}. Note that the method in \cite{alhousani2023geometric} only provides a different way of treating actions, but not states.

The results are shown on the top row of Fig. \ref{fig:results}. We show three metrics: The average success rate during the training process indicates how quickly and consistently training progress is made, and higher values represent faster training progress. The final success rate at the end of training indicates if the policy is at all able to solve the task. And the average reward per step during 160 rollouts of the final policy shows how well the policy performs, as reaching the goal faster yields a higher reward. For all results, we take the average of 100 runs.

A comparison of the required time for a full training run is provided in Tab. \ref{tab:times}. We use the same representation for state and action and show the time spent on network training and policy rollout.
\begin{table}[!htb]
\vspace{-0.0\intextsep}
\centering
\begin{tabular}{l c c c c c c c}
 \toprule
 Time (s) & $\mathfrak{m}$ & $SO_3$ & $SO_3^{1:2}$ & $\mathcal{S}^{3+}$ & $\mathcal{S}^{3}$ & $\measuredangle$ & RM \\
 \midrule
   Train & \textbf{18.8} & 29.0 & 20.5 & 27.0 & 25.9 & 45.1 & 28.9 \\
   Rollout & \textbf{19.5} & 38.6 & 24.8 & 21.3 & 19.9 & 20.0 & 32.7 \\
 \bottomrule
\end{tabular}
 \caption{Time spent during training (see Fig. \ref{fig:results} for labels)}\label{tab:times}
 \vspace{-1.0\intextsep}
\end{table}

\subsection{End effector orientation control}
We evaluate the applicability of the baseline results to robotic systems by requiring the end effector of a robot arm to reach a desired goal frame from the single initial robot configuration. We compare the same 36 orientation representation combinations for state and action as in Sec. \ref{sec:doc}. The task is to rotate the end effector from the initial frame into a goal frame without restrictions on the position. As before, the goal $\bs{g}$ is sampled uniformly in $S^3$ and then converted to other representations, and the action is limited to a relative rotation angle of at most $0.1\pi$. During each episode, the agent can take 100 steps to reach the goal. We train the agent for 2,000,000 environment steps. 

The results are shown on the bottom of Fig. \ref{fig:results}. As before, we provide results for the average success rate, the final success rate, and the average reward per step of the resulting policy. For all results, we take the average of ten runs.

\subsection{Pick-and-place task}
Finally, we perform a pick-and-place task, where the robot arm needs to grasp and move a cube to a desired position and orientation. For this scenario, we use the same representation for state and action and showcase three orientation representations: Lie algebra $\mathfrak{m}$, rotation matrix $SO(3)$, and positive-real-part quaternion $\mathcal{S}^{3+}$. The robot arm is initialized in a single configuration. The cube's position is initialized in a plane on the ground, and its orientation is randomized and then projected to lie flat on the ground. The goal position is in the air, and the orientation is uniformly varied up to $\frac{\pi}{2}$ around a random axis from the initial orientation. The action is limited to a relative rotation angle of at most $0.2\pi$. During each episode, the agent can take 100 steps to reach the goal. We train the agent for 10,000,000 environment steps. We use an expert policy to guide learning similar to the process described in \cite{brudigam2024jacta}.

\begin{figure}[!tb]
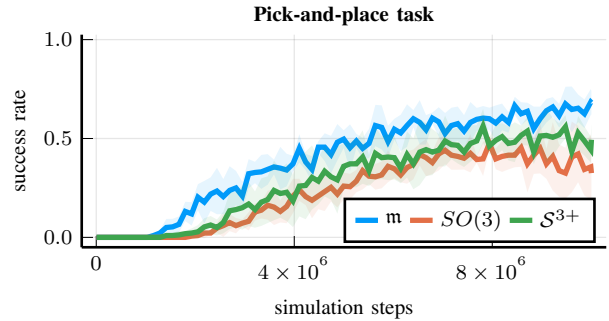

\vspace{-0.0\intextsep}
    \centering
    \include{figures/pickplaceres}
    \vspace{-3.0\intextsep}
    \caption{Comparison of the training performance of selected orientation representations for the pick-and-place task with the same representation for state and action: Lie algebra $\mathfrak{m}$ (blue), rotation matrices $SO(3)$ (red), and positive-real-part quaternion $\mathcal{S}^+$ (green).}
    \label{fig:pick_and_place}
    \vspace{-1.0\intextsep}
\end{figure}

The hardware setup is shown in Fig. \ref{fig:nn_structure} and the training results for three best representations, Lie algebra, rotation matrix, and positive-real-part quaternion, are shown in Fig. \ref{fig:pick_and_place}. For the results, we take the average of five runs. 

We conduct two experiments on hardware. For the first experiment, we place the cube in the center of the table and use eight uniformly spaced starting orientations. The goals are sampled randomly, as in the simulation. We record four trials for each orientation. The second experiment has four different initial cube positions on a rectangle, and we use four uniformly spaced staring orientations. The goals are sampled randomly, as in the simulation. We record two trials for each orientation. The results for evaluation on hardware are shown in Tab. \ref{tab:hardware_res}. 

\begin{table}[!htb]
\vspace{-0.0\intextsep}
\centering
\begin{tabular}{l c c c}
 \toprule
 Experiment & Trials & Success \\
 \midrule
   Center position & 32 & 46.9\% \\
   Rectangle positions & 32 & 50.0\% \\
 \bottomrule
\end{tabular}
 \caption{Success rates for pick-and-place hardware experiments}\label{tab:hardware_res}
 \vspace{-1.0\intextsep}
\end{table}

\section{Discussion}
The results for direct and end effector orientation control, as well as picking-and-placing, show that, overall, the action representation has a larger effect on training performance, and the best training is achieved with our proposed Lie algebra actions, matching the theoretical motivation outlined in the paper. Rotation matrices ($SO(3)$ and $SO_{1:2}(3)$) and positive-real-part quaternions ($\mathcal{S}^{3+}$) achieve good results, and the discontinuity in $\mathcal{S}^{3+}$ does not appear to deteriorate the performance. The results for the state representation are less conclusive. A potential reason could be that networks are able to extract relevant features from all representations, even without adhering to the theoretical manifold at the input. Nonetheless, Euler angles and quaternions perform consistently worse than other representations. A possible reason for this result could be the multi-cover of orientations in these representations. Using the Riemannian manifold actions proposed in \cite{alhousani2023geometric} does not provide good performance. We refer to \cite{jaquier2024unraveling} for potential reasons.

The resulting policies with Lie algebra actions perform slightly worse than certain other representations despite their better training performance. We attribute this result to the fixed box-bound action scale, which results in a smaller maximum rotation in certain directions for Lie algebra actions compared to other representations. Accordingly, the final average reward is lower since it takes longer to reach the goal. Scaling the action dynamically to achieve the full range considerably deteriorates training performance, possibly because now the action scale also has to be learned.

Due to the computationally faster projection, the Lie algebra representation also has the lowest training and rollout times. For the same reason, the Lie algebra representation is also faster than the Riemannian manifold action due to the simpler mapping from manifold to tangent space and back.

\section{Conclusions}
We presented a novel state and action orientation representation for learning with neural networks in reinforcement learning based on the Lie algebra of orientations. Our results show that specifically using the Lie algebra actions results in better training performance and faster training and policy rollout, making them generally a suitable choice for learning with orientations in robotics. Our findings are consistent across tasks with increasing complexity, indicating that the results generally apply to different robotics settings.

\addtolength{\textheight}{-7.5cm}   




\section*{Acknowledgment}
The authors would like to thank Petar Bevanda for the technical discussions and his help in preparing the manuscript.


\bibliography{references}
\bibliographystyle{ieeetr}

\end{document}

%% file: def_color.tex
\pgfplotsset{
colormap={plots1}{rgb(0.00000000)=(0.49230000,0.09080000,0.00010000)
rgb(0.01506375)=(0.49670000,0.10280000,0.00370000)
rgb(0.02962247)=(0.50110000,0.11400000,0.00710000)
rgb(0.04370893)=(0.50550000,0.12470000,0.01040000)
rgb(0.05735281)=(0.50980000,0.13490000,0.01380000)
rgb(0.07058107)=(0.51410000,0.14460000,0.01680000)
rgb(0.08341830)=(0.51840000,0.15400000,0.01970000)
rgb(0.09588694)=(0.52260000,0.16320000,0.02250000)
rgb(0.10800757)=(0.52680000,0.17200000,0.02500000)
rgb(0.11979910)=(0.53100000,0.18070000,0.02750000)
rgb(0.13127891)=(0.53510000,0.18910000,0.03010000)
rgb(0.14246308)=(0.53920000,0.19740000,0.03290000)
rgb(0.15336644)=(0.54330000,0.20550000,0.03590000)
rgb(0.16400276)=(0.54730000,0.21350000,0.03890000)
rgb(0.17438480)=(0.55120000,0.22130000,0.04200000)
rgb(0.18452443)=(0.55510000,0.22900000,0.04500000)
rgb(0.19443272)=(0.55900000,0.23660000,0.04800000)
rgb(0.20411998)=(0.56280000,0.24410000,0.05100000)
rgb(0.21359587)=(0.56650000,0.25150000,0.05400000)
rgb(0.22286942)=(0.57030000,0.25870000,0.05700000)
rgb(0.23194908)=(0.57390000,0.26590000,0.06010000)
rgb(0.24084279)=(0.57760000,0.27300000,0.06300000)
rgb(0.24955802)=(0.58120000,0.28000000,0.06590000)
rgb(0.25810180)=(0.58470000,0.28700000,0.06890000)
rgb(0.26648073)=(0.58820000,0.29390000,0.07170000)
rgb(0.27470106)=(0.59170000,0.30070000,0.07460000)
rgb(0.28276868)=(0.59520000,0.30740000,0.07740000)
rgb(0.29068916)=(0.59860000,0.31420000,0.08030000)
rgb(0.29846778)=(0.60190000,0.32080000,0.08310000)
rgb(0.30610952)=(0.60530000,0.32740000,0.08590000)
rgb(0.31361912)=(0.60860000,0.33390000,0.08870000)
rgb(0.32100108)=(0.61190000,0.34040000,0.09150000)
rgb(0.32825965)=(0.61510000,0.34680000,0.09430000)
rgb(0.33539890)=(0.61830000,0.35320000,0.09700000)
rgb(0.34242268)=(0.62150000,0.35960000,0.09980000)
rgb(0.34933468)=(0.62460000,0.36590000,0.10250000)
rgb(0.35613838)=(0.62780000,0.37220000,0.10520000)
rgb(0.36283715)=(0.63090000,0.37840000,0.10800000)
rgb(0.36943415)=(0.63400000,0.38470000,0.11070000)
rgb(0.37593244)=(0.63710000,0.39090000,0.11350000)
rgb(0.38233494)=(0.64010000,0.39710000,0.11620000)
rgb(0.38864441)=(0.64320000,0.40330000,0.11900000)
rgb(0.39486353)=(0.64620000,0.40940000,0.12170000)
rgb(0.40099485)=(0.64920000,0.41560000,0.12450000)
rgb(0.40704081)=(0.65230000,0.42180000,0.12730000)
rgb(0.41300376)=(0.65530000,0.42800000,0.13020000)
rgb(0.41888594)=(0.65830000,0.43410000,0.13300000)
rgb(0.42468951)=(0.66130000,0.44030000,0.13580000)
rgb(0.43041656)=(0.66430000,0.44650000,0.13880000)
rgb(0.43606906)=(0.66740000,0.45280000,0.14180000)
rgb(0.44164894)=(0.67040000,0.45900000,0.14480000)
rgb(0.44715803)=(0.67340000,0.46530000,0.14780000)
rgb(0.45259812)=(0.67650000,0.47160000,0.15100000)
rgb(0.45797090)=(0.67960000,0.47800000,0.15420000)
rgb(0.46327803)=(0.68260000,0.48430000,0.15740000)
rgb(0.46852108)=(0.68570000,0.49080000,0.16080000)
rgb(0.47370160)=(0.68880000,0.49720000,0.16420000)
rgb(0.47882104)=(0.69200000,0.50370000,0.16770000)
rgb(0.48388084)=(0.69510000,0.51030000,0.17140000)
rgb(0.48888237)=(0.69830000,0.51690000,0.17510000)
rgb(0.49382695)=(0.70150000,0.52360000,0.17900000)
rgb(0.49871587)=(0.70470000,0.53030000,0.18290000)
rgb(0.50355037)=(0.70790000,0.53700000,0.18700000)
rgb(0.50833164)=(0.71120000,0.54380000,0.19120000)
rgb(0.51306084)=(0.71440000,0.55070000,0.19560000)
rgb(0.51773911)=(0.71770000,0.55760000,0.20010000)
rgb(0.52236751)=(0.72100000,0.56460000,0.20480000)
rgb(0.52694711)=(0.72440000,0.57160000,0.20970000)
rgb(0.53147892)=(0.72770000,0.57870000,0.21470000)
rgb(0.53596393)=(0.73110000,0.58590000,0.21990000)
rgb(0.54040309)=(0.73450000,0.59310000,0.22530000)
rgb(0.54479734)=(0.73790000,0.60040000,0.23100000)
rgb(0.54914757)=(0.74130000,0.60770000,0.23680000)
rgb(0.55345466)=(0.74470000,0.61510000,0.24280000)
rgb(0.55771945)=(0.74820000,0.62250000,0.24910000)
rgb(0.56194277)=(0.75160000,0.63000000,0.25560000)
rgb(0.56612541)=(0.75510000,0.63760000,0.26230000)
rgb(0.57026816)=(0.75850000,0.64510000,0.26930000)
rgb(0.57437176)=(0.76200000,0.65280000,0.27650000)
rgb(0.57843695)=(0.76540000,0.66050000,0.28400000)
rgb(0.58246444)=(0.76880000,0.66820000,0.29180000)
rgb(0.58645492)=(0.77220000,0.67590000,0.29970000)
rgb(0.59040907)=(0.77560000,0.68370000,0.30800000)
rgb(0.59432754)=(0.77890000,0.69150000,0.31650000)
rgb(0.59821098)=(0.78220000,0.69940000,0.32530000)
rgb(0.60205999)=(0.78550000,0.70720000,0.33430000)
rgb(0.60587519)=(0.78870000,0.71500000,0.34360000)
rgb(0.60965717)=(0.79180000,0.72280000,0.35310000)
rgb(0.61340650)=(0.79490000,0.73070000,0.36280000)
rgb(0.61712374)=(0.79780000,0.73840000,0.37280000)
rgb(0.62080943)=(0.80070000,0.74620000,0.38310000)
rgb(0.62446410)=(0.80350000,0.75390000,0.39350000)
rgb(0.62808828)=(0.80610000,0.76150000,0.40410000)
rgb(0.63168246)=(0.80860000,0.76910000,0.41500000)
rgb(0.63524714)=(0.81100000,0.77660000,0.42590000)
rgb(0.63878280)=(0.81320000,0.78400000,0.43710000)
rgb(0.64228991)=(0.81530000,0.79130000,0.44840000)
rgb(0.64576892)=(0.81720000,0.79850000,0.45980000)
rgb(0.64922028)=(0.81890000,0.80560000,0.47130000)
rgb(0.65264444)=(0.82040000,0.81250000,0.48280000)
rgb(0.65604180)=(0.82170000,0.81920000,0.49440000)
rgb(0.65941280)=(0.82280000,0.82580000,0.50610000)
rgb(0.66275783)=(0.82360000,0.83230000,0.51780000)
rgb(0.66607730)=(0.82420000,0.83850000,0.52940000)
rgb(0.66937158)=(0.82460000,0.84460000,0.54100000)
rgb(0.67264107)=(0.82480000,0.85040000,0.55260000)
rgb(0.67588612)=(0.82460000,0.85600000,0.56400000)
rgb(0.67910711)=(0.82430000,0.86140000,0.57540000)
rgb(0.68230438)=(0.82360000,0.86660000,0.58670000)
rgb(0.68547829)=(0.82270000,0.87160000,0.59780000)
rgb(0.68862917)=(0.82150000,0.87630000,0.60870000)
rgb(0.69175736)=(0.82000000,0.88070000,0.61950000)
rgb(0.69486317)=(0.81830000,0.88500000,0.63010000)
rgb(0.69794693)=(0.81630000,0.88900000,0.64050000)
rgb(0.70100895)=(0.81400000,0.89270000,0.65070000)
rgb(0.70404953)=(0.81140000,0.89620000,0.66060000)
rgb(0.70706897)=(0.80850000,0.89940000,0.67030000)
rgb(0.71006756)=(0.80530000,0.90240000,0.67980000)
rgb(0.71304559)=(0.80190000,0.90510000,0.68900000)
rgb(0.71600334)=(0.79820000,0.90760000,0.69790000)
rgb(0.71894109)=(0.79420000,0.90980000,0.70660000)
rgb(0.72185909)=(0.79000000,0.91180000,0.71500000)
rgb(0.72475762)=(0.78540000,0.91350000,0.72310000)
rgb(0.72763693)=(0.78060000,0.91500000,0.73100000)
rgb(0.73049727)=(0.77560000,0.91620000,0.73850000)
rgb(0.73333891)=(0.77020000,0.91720000,0.74580000)
rgb(0.73616207)=(0.76460000,0.91800000,0.75280000)
rgb(0.73896699)=(0.75880000,0.91850000,0.75950000)
rgb(0.74175392)=(0.75270000,0.91880000,0.76600000)
rgb(0.74452307)=(0.74630000,0.91890000,0.77210000)
rgb(0.74727468)=(0.73970000,0.91870000,0.77800000)
rgb(0.75000897)=(0.73290000,0.91840000,0.78360000)
rgb(0.75272615)=(0.72580000,0.91770000,0.78890000)
rgb(0.75542644)=(0.71850000,0.91690000,0.79390000)
rgb(0.75811004)=(0.71100000,0.91580000,0.79870000)
rgb(0.76077715)=(0.70330000,0.91460000,0.80320000)
rgb(0.76342799)=(0.69540000,0.91310000,0.80750000)
rgb(0.76606275)=(0.68720000,0.91130000,0.81140000)
rgb(0.76868162)=(0.67890000,0.90940000,0.81520000)
rgb(0.77128479)=(0.67040000,0.90730000,0.81860000)
rgb(0.77387245)=(0.66160000,0.90490000,0.82190000)
rgb(0.77644479)=(0.65270000,0.90240000,0.82490000)
rgb(0.77900197)=(0.64370000,0.89960000,0.82760000)
rgb(0.78154419)=(0.63450000,0.89670000,0.83010000)
rgb(0.78407162)=(0.62510000,0.89350000,0.83240000)
rgb(0.78658442)=(0.61560000,0.89010000,0.83450000)
rgb(0.78908276)=(0.60590000,0.88660000,0.83630000)
rgb(0.79156682)=(0.59610000,0.88290000,0.83800000)
rgb(0.79403675)=(0.58630000,0.87890000,0.83940000)
rgb(0.79649271)=(0.57630000,0.87480000,0.84060000)
rgb(0.79893486)=(0.56620000,0.87050000,0.84160000)
rgb(0.80136335)=(0.55600000,0.86610000,0.84250000)
rgb(0.80377834)=(0.54580000,0.86140000,0.84310000)
rgb(0.80617997)=(0.53550000,0.85660000,0.84350000)
rgb(0.80856840)=(0.52520000,0.85170000,0.84380000)
rgb(0.81094376)=(0.51490000,0.84650000,0.84390000)
rgb(0.81330621)=(0.50450000,0.84130000,0.84380000)
rgb(0.81565587)=(0.49420000,0.83590000,0.84360000)
rgb(0.81799288)=(0.48380000,0.83040000,0.84320000)
rgb(0.82031739)=(0.47350000,0.82470000,0.84270000)
rgb(0.82262952)=(0.46330000,0.81890000,0.84200000)
rgb(0.82492941)=(0.45310000,0.81300000,0.84120000)
rgb(0.82721718)=(0.44300000,0.80700000,0.84020000)
rgb(0.82949297)=(0.43300000,0.80090000,0.83910000)
rgb(0.83175689)=(0.42310000,0.79470000,0.83790000)
rgb(0.83400907)=(0.41330000,0.78850000,0.83660000)
rgb(0.83624963)=(0.40360000,0.78210000,0.83510000)
rgb(0.83847869)=(0.39410000,0.77570000,0.83360000)
rgb(0.84069637)=(0.38470000,0.76930000,0.83200000)
rgb(0.84290278)=(0.37550000,0.76270000,0.83020000)
rgb(0.84509804)=(0.36640000,0.75620000,0.82840000)
rgb(0.84728226)=(0.35760000,0.74960000,0.82650000)
rgb(0.84945555)=(0.34890000,0.74300000,0.82450000)
rgb(0.85161801)=(0.34040000,0.73630000,0.82250000)
rgb(0.85376977)=(0.33220000,0.72960000,0.82040000)
rgb(0.85591091)=(0.32410000,0.72290000,0.81820000)
rgb(0.85804155)=(0.31630000,0.71620000,0.81600000)
rgb(0.86016179)=(0.30870000,0.70950000,0.81370000)
rgb(0.86227172)=(0.30120000,0.70280000,0.81140000)
rgb(0.86437146)=(0.29400000,0.69610000,0.80910000)
rgb(0.86646109)=(0.28710000,0.68940000,0.80670000)
rgb(0.86854072)=(0.28030000,0.68270000,0.80430000)
rgb(0.87061043)=(0.27380000,0.67600000,0.80180000)
rgb(0.87267033)=(0.26750000,0.66940000,0.79930000)
rgb(0.87472051)=(0.26140000,0.66270000,0.79680000)
rgb(0.87676105)=(0.25550000,0.65610000,0.79430000)
rgb(0.87879205)=(0.24990000,0.64950000,0.79180000)
rgb(0.88081359)=(0.24440000,0.64290000,0.78920000)
rgb(0.88282577)=(0.23920000,0.63640000,0.78670000)
rgb(0.88482867)=(0.23410000,0.62990000,0.78410000)
rgb(0.88682237)=(0.22920000,0.62340000,0.78150000)
rgb(0.88880697)=(0.22450000,0.61690000,0.77890000)
rgb(0.89078253)=(0.22010000,0.61050000,0.77630000)
rgb(0.89274915)=(0.21570000,0.60410000,0.77370000)
rgb(0.89470691)=(0.21160000,0.59770000,0.77110000)
rgb(0.89665588)=(0.20760000,0.59130000,0.76850000)
rgb(0.89859614)=(0.20370000,0.58500000,0.76590000)
rgb(0.90052777)=(0.20000000,0.57870000,0.76330000)
rgb(0.90245085)=(0.19640000,0.57240000,0.76070000)
rgb(0.90436545)=(0.19300000,0.56610000,0.75810000)
rgb(0.90627165)=(0.18970000,0.55990000,0.75550000)
rgb(0.90816951)=(0.18650000,0.55370000,0.75290000)
rgb(0.91005912)=(0.18340000,0.54750000,0.75030000)
rgb(0.91194054)=(0.18040000,0.54130000,0.74770000)
rgb(0.91381385)=(0.17760000,0.53520000,0.74510000)
rgb(0.91567911)=(0.17480000,0.52900000,0.74250000)
rgb(0.91753640)=(0.17210000,0.52290000,0.73990000)
rgb(0.91938578)=(0.16950000,0.51680000,0.73730000)
rgb(0.92122731)=(0.16700000,0.51060000,0.73470000)
rgb(0.92306107)=(0.16450000,0.50450000,0.73210000)
rgb(0.92488712)=(0.16210000,0.49840000,0.72950000)
rgb(0.92670552)=(0.15970000,0.49230000,0.72680000)
rgb(0.92851634)=(0.15740000,0.48610000,0.72420000)
rgb(0.93031964)=(0.15520000,0.48000000,0.72160000)
rgb(0.93211549)=(0.15300000,0.47380000,0.71890000)
rgb(0.93390393)=(0.15080000,0.46770000,0.71630000)
rgb(0.93568505)=(0.14870000,0.46150000,0.71360000)
rgb(0.93745889)=(0.14650000,0.45530000,0.71100000)
rgb(0.93922551)=(0.14450000,0.44900000,0.70830000)
rgb(0.94098498)=(0.14240000,0.44280000,0.70560000)
rgb(0.94273735)=(0.14030000,0.43650000,0.70290000)
rgb(0.94448267)=(0.13830000,0.43010000,0.70020000)
rgb(0.94622101)=(0.13620000,0.42380000,0.69740000)
rgb(0.94795242)=(0.13420000,0.41740000,0.69470000)
rgb(0.94967695)=(0.13210000,0.41100000,0.69190000)
rgb(0.95139467)=(0.13010000,0.40450000,0.68910000)
rgb(0.95310561)=(0.12800000,0.39800000,0.68630000)
rgb(0.95480984)=(0.12590000,0.39150000,0.68350000)
rgb(0.95650741)=(0.12370000,0.38490000,0.68060000)
rgb(0.95819837)=(0.12150000,0.37840000,0.67770000)
rgb(0.95988278)=(0.11930000,0.37170000,0.67490000)
rgb(0.96156067)=(0.11700000,0.36500000,0.67200000)
rgb(0.96323211)=(0.11470000,0.35830000,0.66900000)
rgb(0.96489714)=(0.11230000,0.35160000,0.66610000)
rgb(0.96655581)=(0.10990000,0.34480000,0.66320000)
rgb(0.96820817)=(0.10730000,0.33800000,0.66020000)
rgb(0.96985426)=(0.10460000,0.33110000,0.65720000)
rgb(0.97149414)=(0.10200000,0.32420000,0.65420000)
rgb(0.97312785)=(0.09910000,0.31730000,0.65120000)
rgb(0.97475544)=(0.09610000,0.31030000,0.64810000)
rgb(0.97637695)=(0.09300000,0.30330000,0.64500000)
rgb(0.97799244)=(0.08980000,0.29620000,0.64200000)
rgb(0.97960193)=(0.08640000,0.28910000,0.63880000)
rgb(0.98120548)=(0.08280000,0.28200000,0.63570000)
rgb(0.98280313)=(0.07890000,0.27480000,0.63260000)
rgb(0.98439493)=(0.07480000,0.26750000,0.62940000)
rgb(0.98598091)=(0.07040000,0.26020000,0.62620000)
rgb(0.98756112)=(0.06570000,0.25290000,0.62300000)
rgb(0.98913560)=(0.06060000,0.24550000,0.61980000)
rgb(0.99070440)=(0.05500000,0.23800000,0.61650000)
rgb(0.99226755)=(0.04890000,0.23050000,0.61330000)
rgb(0.99382509)=(0.04200000,0.22300000,0.61000000)
rgb(0.99537707)=(0.03410000,0.21530000,0.60670000)
rgb(0.99692352)=(0.02620000,0.20770000,0.60340000)
rgb(0.99846448)=(0.01820000,0.19990000,0.60000000)
rgb(1.00000000)=(0.00980000,0.19210000,0.59670000)},
}
\pgfplotsset{
colormap={plots2}{rgb(0.00000000)=(0.49230000,0.09080000,0.00010000)
rgb(0.01506375)=(0.49670000,0.10280000,0.00370000)
rgb(0.02962247)=(0.50110000,0.11400000,0.00710000)
rgb(0.04370893)=(0.50550000,0.12470000,0.01040000)
rgb(0.05735281)=(0.50980000,0.13490000,0.01380000)
rgb(0.07058107)=(0.51410000,0.14460000,0.01680000)
rgb(0.08341830)=(0.51840000,0.15400000,0.01970000)
rgb(0.09588694)=(0.52260000,0.16320000,0.02250000)
rgb(0.10800757)=(0.52680000,0.17200000,0.02500000)
rgb(0.11979910)=(0.53100000,0.18070000,0.02750000)
rgb(0.13127891)=(0.53510000,0.18910000,0.03010000)
rgb(0.14246308)=(0.53920000,0.19740000,0.03290000)
rgb(0.15336644)=(0.54330000,0.20550000,0.03590000)
rgb(0.16400276)=(0.54730000,0.21350000,0.03890000)
rgb(0.17438480)=(0.55120000,0.22130000,0.04200000)
rgb(0.18452443)=(0.55510000,0.22900000,0.04500000)
rgb(0.19443272)=(0.55900000,0.23660000,0.04800000)
rgb(0.20411998)=(0.56280000,0.24410000,0.05100000)
rgb(0.21359587)=(0.56650000,0.25150000,0.05400000)
rgb(0.22286942)=(0.57030000,0.25870000,0.05700000)
rgb(0.23194908)=(0.57390000,0.26590000,0.06010000)
rgb(0.24084279)=(0.57760000,0.27300000,0.06300000)
rgb(0.24955802)=(0.58120000,0.28000000,0.06590000)
rgb(0.25810180)=(0.58470000,0.28700000,0.06890000)
rgb(0.26648073)=(0.58820000,0.29390000,0.07170000)
rgb(0.27470106)=(0.59170000,0.30070000,0.07460000)
rgb(0.28276868)=(0.59520000,0.30740000,0.07740000)
rgb(0.29068916)=(0.59860000,0.31420000,0.08030000)
rgb(0.29846778)=(0.60190000,0.32080000,0.08310000)
rgb(0.30610952)=(0.60530000,0.32740000,0.08590000)
rgb(0.31361912)=(0.60860000,0.33390000,0.08870000)
rgb(0.32100108)=(0.61190000,0.34040000,0.09150000)
rgb(0.32825965)=(0.61510000,0.34680000,0.09430000)
rgb(0.33539890)=(0.61830000,0.35320000,0.09700000)
rgb(0.34242268)=(0.62150000,0.35960000,0.09980000)
rgb(0.34933468)=(0.62460000,0.36590000,0.10250000)
rgb(0.35613838)=(0.62780000,0.37220000,0.10520000)
rgb(0.36283715)=(0.63090000,0.37840000,0.10800000)
rgb(0.36943415)=(0.63400000,0.38470000,0.11070000)
rgb(0.37593244)=(0.63710000,0.39090000,0.11350000)
rgb(0.38233494)=(0.64010000,0.39710000,0.11620000)
rgb(0.38864441)=(0.64320000,0.40330000,0.11900000)
rgb(0.39486353)=(0.64620000,0.40940000,0.12170000)
rgb(0.40099485)=(0.64920000,0.41560000,0.12450000)
rgb(0.40704081)=(0.65230000,0.42180000,0.12730000)
rgb(0.41300376)=(0.65530000,0.42800000,0.13020000)
rgb(0.41888594)=(0.65830000,0.43410000,0.13300000)
rgb(0.42468951)=(0.66130000,0.44030000,0.13580000)
rgb(0.43041656)=(0.66430000,0.44650000,0.13880000)
rgb(0.43606906)=(0.66740000,0.45280000,0.14180000)
rgb(0.44164894)=(0.67040000,0.45900000,0.14480000)
rgb(0.44715803)=(0.67340000,0.46530000,0.14780000)
rgb(0.45259812)=(0.67650000,0.47160000,0.15100000)
rgb(0.45797090)=(0.67960000,0.47800000,0.15420000)
rgb(0.46327803)=(0.68260000,0.48430000,0.15740000)
rgb(0.46852108)=(0.68570000,0.49080000,0.16080000)
rgb(0.47370160)=(0.68880000,0.49720000,0.16420000)
rgb(0.47882104)=(0.69200000,0.50370000,0.16770000)
rgb(0.48388084)=(0.69510000,0.51030000,0.17140000)
rgb(0.48888237)=(0.69830000,0.51690000,0.17510000)
rgb(0.49382695)=(0.70150000,0.52360000,0.17900000)
rgb(0.49871587)=(0.70470000,0.53030000,0.18290000)
rgb(0.50355037)=(0.70790000,0.53700000,0.18700000)
rgb(0.50833164)=(0.71120000,0.54380000,0.19120000)
rgb(0.51306084)=(0.71440000,0.55070000,0.19560000)
rgb(0.51773911)=(0.71770000,0.55760000,0.20010000)
rgb(0.52236751)=(0.72100000,0.56460000,0.20480000)
rgb(0.52694711)=(0.72440000,0.57160000,0.20970000)
rgb(0.53147892)=(0.72770000,0.57870000,0.21470000)
rgb(0.53596393)=(0.73110000,0.58590000,0.21990000)
rgb(0.54040309)=(0.73450000,0.59310000,0.22530000)
rgb(0.54479734)=(0.73790000,0.60040000,0.23100000)
rgb(0.54914757)=(0.74130000,0.60770000,0.23680000)
rgb(0.55345466)=(0.74470000,0.61510000,0.24280000)
rgb(0.55771945)=(0.74820000,0.62250000,0.24910000)
rgb(0.56194277)=(0.75160000,0.63000000,0.25560000)
rgb(0.56612541)=(0.75510000,0.63760000,0.26230000)
rgb(0.57026816)=(0.75850000,0.64510000,0.26930000)
rgb(0.57437176)=(0.76200000,0.65280000,0.27650000)
rgb(0.57843695)=(0.76540000,0.66050000,0.28400000)
rgb(0.58246444)=(0.76880000,0.66820000,0.29180000)
rgb(0.58645492)=(0.77220000,0.67590000,0.29970000)
rgb(0.59040907)=(0.77560000,0.68370000,0.30800000)
rgb(0.59432754)=(0.77890000,0.69150000,0.31650000)
rgb(0.59821098)=(0.78220000,0.69940000,0.32530000)
rgb(0.60205999)=(0.78550000,0.70720000,0.33430000)
rgb(0.60587519)=(0.78870000,0.71500000,0.34360000)
rgb(0.60965717)=(0.79180000,0.72280000,0.35310000)
rgb(0.61340650)=(0.79490000,0.73070000,0.36280000)
rgb(0.61712374)=(0.79780000,0.73840000,0.37280000)
rgb(0.62080943)=(0.80070000,0.74620000,0.38310000)
rgb(0.62446410)=(0.80350000,0.75390000,0.39350000)
rgb(0.62808828)=(0.80610000,0.76150000,0.40410000)
rgb(0.63168246)=(0.80860000,0.76910000,0.41500000)
rgb(0.63524714)=(0.81100000,0.77660000,0.42590000)
rgb(0.63878280)=(0.81320000,0.78400000,0.43710000)
rgb(0.64228991)=(0.81530000,0.79130000,0.44840000)
rgb(0.64576892)=(0.81720000,0.79850000,0.45980000)
rgb(0.64922028)=(0.81890000,0.80560000,0.47130000)
rgb(0.65264444)=(0.82040000,0.81250000,0.48280000)
rgb(0.65604180)=(0.82170000,0.81920000,0.49440000)
rgb(0.65941280)=(0.82280000,0.82580000,0.50610000)
rgb(0.66275783)=(0.82360000,0.83230000,0.51780000)
rgb(0.66607730)=(0.82420000,0.83850000,0.52940000)
rgb(0.66937158)=(0.82460000,0.84460000,0.54100000)
rgb(0.67264107)=(0.82480000,0.85040000,0.55260000)
rgb(0.67588612)=(0.82460000,0.85600000,0.56400000)
rgb(0.67910711)=(0.82430000,0.86140000,0.57540000)
rgb(0.68230438)=(0.82360000,0.86660000,0.58670000)
rgb(0.68547829)=(0.82270000,0.87160000,0.59780000)
rgb(0.68862917)=(0.82150000,0.87630000,0.60870000)
rgb(0.69175736)=(0.82000000,0.88070000,0.61950000)
rgb(0.69486317)=(0.81830000,0.88500000,0.63010000)
rgb(0.69794693)=(0.81630000,0.88900000,0.64050000)
rgb(0.70100895)=(0.81400000,0.89270000,0.65070000)
rgb(0.70404953)=(0.81140000,0.89620000,0.66060000)
rgb(0.70706897)=(0.80850000,0.89940000,0.67030000)
rgb(0.71006756)=(0.80530000,0.90240000,0.67980000)
rgb(0.71304559)=(0.80190000,0.90510000,0.68900000)
rgb(0.71600334)=(0.79820000,0.90760000,0.69790000)
rgb(0.71894109)=(0.79420000,0.90980000,0.70660000)
rgb(0.72185909)=(0.79000000,0.91180000,0.71500000)
rgb(0.72475762)=(0.78540000,0.91350000,0.72310000)
rgb(0.72763693)=(0.78060000,0.91500000,0.73100000)
rgb(0.73049727)=(0.77560000,0.91620000,0.73850000)
rgb(0.73333891)=(0.77020000,0.91720000,0.74580000)
rgb(0.73616207)=(0.76460000,0.91800000,0.75280000)
rgb(0.73896699)=(0.75880000,0.91850000,0.75950000)
rgb(0.74175392)=(0.75270000,0.91880000,0.76600000)
rgb(0.74452307)=(0.74630000,0.91890000,0.77210000)
rgb(0.74727468)=(0.73970000,0.91870000,0.77800000)
rgb(0.75000897)=(0.73290000,0.91840000,0.78360000)
rgb(0.75272615)=(0.72580000,0.91770000,0.78890000)
rgb(0.75542644)=(0.71850000,0.91690000,0.79390000)
rgb(0.75811004)=(0.71100000,0.91580000,0.79870000)
rgb(0.76077715)=(0.70330000,0.91460000,0.80320000)
rgb(0.76342799)=(0.69540000,0.91310000,0.80750000)
rgb(0.76606275)=(0.68720000,0.91130000,0.81140000)
rgb(0.76868162)=(0.67890000,0.90940000,0.81520000)
rgb(0.77128479)=(0.67040000,0.90730000,0.81860000)
rgb(0.77387245)=(0.66160000,0.90490000,0.82190000)
rgb(0.77644479)=(0.65270000,0.90240000,0.82490000)
rgb(0.77900197)=(0.64370000,0.89960000,0.82760000)
rgb(0.78154419)=(0.63450000,0.89670000,0.83010000)
rgb(0.78407162)=(0.62510000,0.89350000,0.83240000)
rgb(0.78658442)=(0.61560000,0.89010000,0.83450000)
rgb(0.78908276)=(0.60590000,0.88660000,0.83630000)
rgb(0.79156682)=(0.59610000,0.88290000,0.83800000)
rgb(0.79403675)=(0.58630000,0.87890000,0.83940000)
rgb(0.79649271)=(0.57630000,0.87480000,0.84060000)
rgb(0.79893486)=(0.56620000,0.87050000,0.84160000)
rgb(0.80136335)=(0.55600000,0.86610000,0.84250000)
rgb(0.80377834)=(0.54580000,0.86140000,0.84310000)
rgb(0.80617997)=(0.53550000,0.85660000,0.84350000)
rgb(0.80856840)=(0.52520000,0.85170000,0.84380000)
rgb(0.81094376)=(0.51490000,0.84650000,0.84390000)
rgb(0.81330621)=(0.50450000,0.84130000,0.84380000)
rgb(0.81565587)=(0.49420000,0.83590000,0.84360000)
rgb(0.81799288)=(0.48380000,0.83040000,0.84320000)
rgb(0.82031739)=(0.47350000,0.82470000,0.84270000)
rgb(0.82262952)=(0.46330000,0.81890000,0.84200000)
rgb(0.82492941)=(0.45310000,0.81300000,0.84120000)
rgb(0.82721718)=(0.44300000,0.80700000,0.84020000)
rgb(0.82949297)=(0.43300000,0.80090000,0.83910000)
rgb(0.83175689)=(0.42310000,0.79470000,0.83790000)
rgb(0.83400907)=(0.41330000,0.78850000,0.83660000)
rgb(0.83624963)=(0.40360000,0.78210000,0.83510000)
rgb(0.83847869)=(0.39410000,0.77570000,0.83360000)
rgb(0.84069637)=(0.38470000,0.76930000,0.83200000)
rgb(0.84290278)=(0.37550000,0.76270000,0.83020000)
rgb(0.84509804)=(0.36640000,0.75620000,0.82840000)
rgb(0.84728226)=(0.35760000,0.74960000,0.82650000)
rgb(0.84945555)=(0.34890000,0.74300000,0.82450000)
rgb(0.85161801)=(0.34040000,0.73630000,0.82250000)
rgb(0.85376977)=(0.33220000,0.72960000,0.82040000)
rgb(0.85591091)=(0.32410000,0.72290000,0.81820000)
rgb(0.85804155)=(0.31630000,0.71620000,0.81600000)
rgb(0.86016179)=(0.30870000,0.70950000,0.81370000)
rgb(0.86227172)=(0.30120000,0.70280000,0.81140000)
rgb(0.86437146)=(0.29400000,0.69610000,0.80910000)
rgb(0.86646109)=(0.28710000,0.68940000,0.80670000)
rgb(0.86854072)=(0.28030000,0.68270000,0.80430000)
rgb(0.87061043)=(0.27380000,0.67600000,0.80180000)
rgb(0.87267033)=(0.26750000,0.66940000,0.79930000)
rgb(0.87472051)=(0.26140000,0.66270000,0.79680000)
rgb(0.87676105)=(0.25550000,0.65610000,0.79430000)
rgb(0.87879205)=(0.24990000,0.64950000,0.79180000)
rgb(0.88081359)=(0.24440000,0.64290000,0.78920000)
rgb(0.88282577)=(0.23920000,0.63640000,0.78670000)
rgb(0.88482867)=(0.23410000,0.62990000,0.78410000)
rgb(0.88682237)=(0.22920000,0.62340000,0.78150000)
rgb(0.88880697)=(0.22450000,0.61690000,0.77890000)
rgb(0.89078253)=(0.22010000,0.61050000,0.77630000)
rgb(0.89274915)=(0.21570000,0.60410000,0.77370000)
rgb(0.89470691)=(0.21160000,0.59770000,0.77110000)
rgb(0.89665588)=(0.20760000,0.59130000,0.76850000)
rgb(0.89859614)=(0.20370000,0.58500000,0.76590000)
rgb(0.90052777)=(0.20000000,0.57870000,0.76330000)
rgb(0.90245085)=(0.19640000,0.57240000,0.76070000)
rgb(0.90436545)=(0.19300000,0.56610000,0.75810000)
rgb(0.90627165)=(0.18970000,0.55990000,0.75550000)
rgb(0.90816951)=(0.18650000,0.55370000,0.75290000)
rgb(0.91005912)=(0.18340000,0.54750000,0.75030000)
rgb(0.91194054)=(0.18040000,0.54130000,0.74770000)
rgb(0.91381385)=(0.17760000,0.53520000,0.74510000)
rgb(0.91567911)=(0.17480000,0.52900000,0.74250000)
rgb(0.91753640)=(0.17210000,0.52290000,0.73990000)
rgb(0.91938578)=(0.16950000,0.51680000,0.73730000)
rgb(0.92122731)=(0.16700000,0.51060000,0.73470000)
rgb(0.92306107)=(0.16450000,0.50450000,0.73210000)
rgb(0.92488712)=(0.16210000,0.49840000,0.72950000)
rgb(0.92670552)=(0.15970000,0.49230000,0.72680000)
rgb(0.92851634)=(0.15740000,0.48610000,0.72420000)
rgb(0.93031964)=(0.15520000,0.48000000,0.72160000)
rgb(0.93211549)=(0.15300000,0.47380000,0.71890000)
rgb(0.93390393)=(0.15080000,0.46770000,0.71630000)
rgb(0.93568505)=(0.14870000,0.46150000,0.71360000)
rgb(0.93745889)=(0.14650000,0.45530000,0.71100000)
rgb(0.93922551)=(0.14450000,0.44900000,0.70830000)
rgb(0.94098498)=(0.14240000,0.44280000,0.70560000)
rgb(0.94273735)=(0.14030000,0.43650000,0.70290000)
rgb(0.94448267)=(0.13830000,0.43010000,0.70020000)
rgb(0.94622101)=(0.13620000,0.42380000,0.69740000)
rgb(0.94795242)=(0.13420000,0.41740000,0.69470000)
rgb(0.94967695)=(0.13210000,0.41100000,0.69190000)
rgb(0.95139467)=(0.13010000,0.40450000,0.68910000)
rgb(0.95310561)=(0.12800000,0.39800000,0.68630000)
rgb(0.95480984)=(0.12590000,0.39150000,0.68350000)
rgb(0.95650741)=(0.12370000,0.38490000,0.68060000)
rgb(0.95819837)=(0.12150000,0.37840000,0.67770000)
rgb(0.95988278)=(0.11930000,0.37170000,0.67490000)
rgb(0.96156067)=(0.11700000,0.36500000,0.67200000)
rgb(0.96323211)=(0.11470000,0.35830000,0.66900000)
rgb(0.96489714)=(0.11230000,0.35160000,0.66610000)
rgb(0.96655581)=(0.10990000,0.34480000,0.66320000)
rgb(0.96820817)=(0.10730000,0.33800000,0.66020000)
rgb(0.96985426)=(0.10460000,0.33110000,0.65720000)
rgb(0.97149414)=(0.10200000,0.32420000,0.65420000)
rgb(0.97312785)=(0.09910000,0.31730000,0.65120000)
rgb(0.97475544)=(0.09610000,0.31030000,0.64810000)
rgb(0.97637695)=(0.09300000,0.30330000,0.64500000)
rgb(0.97799244)=(0.08980000,0.29620000,0.64200000)
rgb(0.97960193)=(0.08640000,0.28910000,0.63880000)
rgb(0.98120548)=(0.08280000,0.28200000,0.63570000)
rgb(0.98280313)=(0.07890000,0.27480000,0.63260000)
rgb(0.98439493)=(0.07480000,0.26750000,0.62940000)
rgb(0.98598091)=(0.07040000,0.26020000,0.62620000)
rgb(0.98756112)=(0.06570000,0.25290000,0.62300000)
rgb(0.98913560)=(0.06060000,0.24550000,0.61980000)
rgb(0.99070440)=(0.05500000,0.23800000,0.61650000)
rgb(0.99226755)=(0.04890000,0.23050000,0.61330000)
rgb(0.99382509)=(0.04200000,0.22300000,0.61000000)
rgb(0.99537707)=(0.03410000,0.21530000,0.60670000)
rgb(0.99692352)=(0.02620000,0.20770000,0.60340000)
rgb(0.99846448)=(0.01820000,0.19990000,0.60000000)
rgb(1.00000000)=(0.00980000,0.19210000,0.59670000)},
}
\pgfplotsset{
colormap={plots3}{rgb(0.00000000)=(0.49230000,0.09080000,0.00010000)
rgb(0.01506375)=(0.49670000,0.10280000,0.00370000)
rgb(0.02962247)=(0.50110000,0.11400000,0.00710000)
rgb(0.04370893)=(0.50550000,0.12470000,0.01040000)
rgb(0.05735281)=(0.50980000,0.13490000,0.01380000)
rgb(0.07058107)=(0.51410000,0.14460000,0.01680000)
rgb(0.08341830)=(0.51840000,0.15400000,0.01970000)
rgb(0.09588694)=(0.52260000,0.16320000,0.02250000)
rgb(0.10800757)=(0.52680000,0.17200000,0.02500000)
rgb(0.11979910)=(0.53100000,0.18070000,0.02750000)
rgb(0.13127891)=(0.53510000,0.18910000,0.03010000)
rgb(0.14246308)=(0.53920000,0.19740000,0.03290000)
rgb(0.15336644)=(0.54330000,0.20550000,0.03590000)
rgb(0.16400276)=(0.54730000,0.21350000,0.03890000)
rgb(0.17438480)=(0.55120000,0.22130000,0.04200000)
rgb(0.18452443)=(0.55510000,0.22900000,0.04500000)
rgb(0.19443272)=(0.55900000,0.23660000,0.04800000)
rgb(0.20411998)=(0.56280000,0.24410000,0.05100000)
rgb(0.21359587)=(0.56650000,0.25150000,0.05400000)
rgb(0.22286942)=(0.57030000,0.25870000,0.05700000)
rgb(0.23194908)=(0.57390000,0.26590000,0.06010000)
rgb(0.24084279)=(0.57760000,0.27300000,0.06300000)
rgb(0.24955802)=(0.58120000,0.28000000,0.06590000)
rgb(0.25810180)=(0.58470000,0.28700000,0.06890000)
rgb(0.26648073)=(0.58820000,0.29390000,0.07170000)
rgb(0.27470106)=(0.59170000,0.30070000,0.07460000)
rgb(0.28276868)=(0.59520000,0.30740000,0.07740000)
rgb(0.29068916)=(0.59860000,0.31420000,0.08030000)
rgb(0.29846778)=(0.60190000,0.32080000,0.08310000)
rgb(0.30610952)=(0.60530000,0.32740000,0.08590000)
rgb(0.31361912)=(0.60860000,0.33390000,0.08870000)
rgb(0.32100108)=(0.61190000,0.34040000,0.09150000)
rgb(0.32825965)=(0.61510000,0.34680000,0.09430000)
rgb(0.33539890)=(0.61830000,0.35320000,0.09700000)
rgb(0.34242268)=(0.62150000,0.35960000,0.09980000)
rgb(0.34933468)=(0.62460000,0.36590000,0.10250000)
rgb(0.35613838)=(0.62780000,0.37220000,0.10520000)
rgb(0.36283715)=(0.63090000,0.37840000,0.10800000)
rgb(0.36943415)=(0.63400000,0.38470000,0.11070000)
rgb(0.37593244)=(0.63710000,0.39090000,0.11350000)
rgb(0.38233494)=(0.64010000,0.39710000,0.11620000)
rgb(0.38864441)=(0.64320000,0.40330000,0.11900000)
rgb(0.39486353)=(0.64620000,0.40940000,0.12170000)
rgb(0.40099485)=(0.64920000,0.41560000,0.12450000)
rgb(0.40704081)=(0.65230000,0.42180000,0.12730000)
rgb(0.41300376)=(0.65530000,0.42800000,0.13020000)
rgb(0.41888594)=(0.65830000,0.43410000,0.13300000)
rgb(0.42468951)=(0.66130000,0.44030000,0.13580000)
rgb(0.43041656)=(0.66430000,0.44650000,0.13880000)
rgb(0.43606906)=(0.66740000,0.45280000,0.14180000)
rgb(0.44164894)=(0.67040000,0.45900000,0.14480000)
rgb(0.44715803)=(0.67340000,0.46530000,0.14780000)
rgb(0.45259812)=(0.67650000,0.47160000,0.15100000)
rgb(0.45797090)=(0.67960000,0.47800000,0.15420000)
rgb(0.46327803)=(0.68260000,0.48430000,0.15740000)
rgb(0.46852108)=(0.68570000,0.49080000,0.16080000)
rgb(0.47370160)=(0.68880000,0.49720000,0.16420000)
rgb(0.47882104)=(0.69200000,0.50370000,0.16770000)
rgb(0.48388084)=(0.69510000,0.51030000,0.17140000)
rgb(0.48888237)=(0.69830000,0.51690000,0.17510000)
rgb(0.49382695)=(0.70150000,0.52360000,0.17900000)
rgb(0.49871587)=(0.70470000,0.53030000,0.18290000)
rgb(0.50355037)=(0.70790000,0.53700000,0.18700000)
rgb(0.50833164)=(0.71120000,0.54380000,0.19120000)
rgb(0.51306084)=(0.71440000,0.55070000,0.19560000)
rgb(0.51773911)=(0.71770000,0.55760000,0.20010000)
rgb(0.52236751)=(0.72100000,0.56460000,0.20480000)
rgb(0.52694711)=(0.72440000,0.57160000,0.20970000)
rgb(0.53147892)=(0.72770000,0.57870000,0.21470000)
rgb(0.53596393)=(0.73110000,0.58590000,0.21990000)
rgb(0.54040309)=(0.73450000,0.59310000,0.22530000)
rgb(0.54479734)=(0.73790000,0.60040000,0.23100000)
rgb(0.54914757)=(0.74130000,0.60770000,0.23680000)
rgb(0.55345466)=(0.74470000,0.61510000,0.24280000)
rgb(0.55771945)=(0.74820000,0.62250000,0.24910000)
rgb(0.56194277)=(0.75160000,0.63000000,0.25560000)
rgb(0.56612541)=(0.75510000,0.63760000,0.26230000)
rgb(0.57026816)=(0.75850000,0.64510000,0.26930000)
rgb(0.57437176)=(0.76200000,0.65280000,0.27650000)
rgb(0.57843695)=(0.76540000,0.66050000,0.28400000)
rgb(0.58246444)=(0.76880000,0.66820000,0.29180000)
rgb(0.58645492)=(0.77220000,0.67590000,0.29970000)
rgb(0.59040907)=(0.77560000,0.68370000,0.30800000)
rgb(0.59432754)=(0.77890000,0.69150000,0.31650000)
rgb(0.59821098)=(0.78220000,0.69940000,0.32530000)
rgb(0.60205999)=(0.78550000,0.70720000,0.33430000)
rgb(0.60587519)=(0.78870000,0.71500000,0.34360000)
rgb(0.60965717)=(0.79180000,0.72280000,0.35310000)
rgb(0.61340650)=(0.79490000,0.73070000,0.36280000)
rgb(0.61712374)=(0.79780000,0.73840000,0.37280000)
rgb(0.62080943)=(0.80070000,0.74620000,0.38310000)
rgb(0.62446410)=(0.80350000,0.75390000,0.39350000)
rgb(0.62808828)=(0.80610000,0.76150000,0.40410000)
rgb(0.63168246)=(0.80860000,0.76910000,0.41500000)
rgb(0.63524714)=(0.81100000,0.77660000,0.42590000)
rgb(0.63878280)=(0.81320000,0.78400000,0.43710000)
rgb(0.64228991)=(0.81530000,0.79130000,0.44840000)
rgb(0.64576892)=(0.81720000,0.79850000,0.45980000)
rgb(0.64922028)=(0.81890000,0.80560000,0.47130000)
rgb(0.65264444)=(0.82040000,0.81250000,0.48280000)
rgb(0.65604180)=(0.82170000,0.81920000,0.49440000)
rgb(0.65941280)=(0.82280000,0.82580000,0.50610000)
rgb(0.66275783)=(0.82360000,0.83230000,0.51780000)
rgb(0.66607730)=(0.82420000,0.83850000,0.52940000)
rgb(0.66937158)=(0.82460000,0.84460000,0.54100000)
rgb(0.67264107)=(0.82480000,0.85040000,0.55260000)
rgb(0.67588612)=(0.82460000,0.85600000,0.56400000)
rgb(0.67910711)=(0.82430000,0.86140000,0.57540000)
rgb(0.68230438)=(0.82360000,0.86660000,0.58670000)
rgb(0.68547829)=(0.82270000,0.87160000,0.59780000)
rgb(0.68862917)=(0.82150000,0.87630000,0.60870000)
rgb(0.69175736)=(0.82000000,0.88070000,0.61950000)
rgb(0.69486317)=(0.81830000,0.88500000,0.63010000)
rgb(0.69794693)=(0.81630000,0.88900000,0.64050000)
rgb(0.70100895)=(0.81400000,0.89270000,0.65070000)
rgb(0.70404953)=(0.81140000,0.89620000,0.66060000)
rgb(0.70706897)=(0.80850000,0.89940000,0.67030000)
rgb(0.71006756)=(0.80530000,0.90240000,0.67980000)
rgb(0.71304559)=(0.80190000,0.90510000,0.68900000)
rgb(0.71600334)=(0.79820000,0.90760000,0.69790000)
rgb(0.71894109)=(0.79420000,0.90980000,0.70660000)
rgb(0.72185909)=(0.79000000,0.91180000,0.71500000)
rgb(0.72475762)=(0.78540000,0.91350000,0.72310000)
rgb(0.72763693)=(0.78060000,0.91500000,0.73100000)
rgb(0.73049727)=(0.77560000,0.91620000,0.73850000)
rgb(0.73333891)=(0.77020000,0.91720000,0.74580000)
rgb(0.73616207)=(0.76460000,0.91800000,0.75280000)
rgb(0.73896699)=(0.75880000,0.91850000,0.75950000)
rgb(0.74175392)=(0.75270000,0.91880000,0.76600000)
rgb(0.74452307)=(0.74630000,0.91890000,0.77210000)
rgb(0.74727468)=(0.73970000,0.91870000,0.77800000)
rgb(0.75000897)=(0.73290000,0.91840000,0.78360000)
rgb(0.75272615)=(0.72580000,0.91770000,0.78890000)
rgb(0.75542644)=(0.71850000,0.91690000,0.79390000)
rgb(0.75811004)=(0.71100000,0.91580000,0.79870000)
rgb(0.76077715)=(0.70330000,0.91460000,0.80320000)
rgb(0.76342799)=(0.69540000,0.91310000,0.80750000)
rgb(0.76606275)=(0.68720000,0.91130000,0.81140000)
rgb(0.76868162)=(0.67890000,0.90940000,0.81520000)
rgb(0.77128479)=(0.67040000,0.90730000,0.81860000)
rgb(0.77387245)=(0.66160000,0.90490000,0.82190000)
rgb(0.77644479)=(0.65270000,0.90240000,0.82490000)
rgb(0.77900197)=(0.64370000,0.89960000,0.82760000)
rgb(0.78154419)=(0.63450000,0.89670000,0.83010000)
rgb(0.78407162)=(0.62510000,0.89350000,0.83240000)
rgb(0.78658442)=(0.61560000,0.89010000,0.83450000)
rgb(0.78908276)=(0.60590000,0.88660000,0.83630000)
rgb(0.79156682)=(0.59610000,0.88290000,0.83800000)
rgb(0.79403675)=(0.58630000,0.87890000,0.83940000)
rgb(0.79649271)=(0.57630000,0.87480000,0.84060000)
rgb(0.79893486)=(0.56620000,0.87050000,0.84160000)
rgb(0.80136335)=(0.55600000,0.86610000,0.84250000)
rgb(0.80377834)=(0.54580000,0.86140000,0.84310000)
rgb(0.80617997)=(0.53550000,0.85660000,0.84350000)
rgb(0.80856840)=(0.52520000,0.85170000,0.84380000)
rgb(0.81094376)=(0.51490000,0.84650000,0.84390000)
rgb(0.81330621)=(0.50450000,0.84130000,0.84380000)
rgb(0.81565587)=(0.49420000,0.83590000,0.84360000)
rgb(0.81799288)=(0.48380000,0.83040000,0.84320000)
rgb(0.82031739)=(0.47350000,0.82470000,0.84270000)
rgb(0.82262952)=(0.46330000,0.81890000,0.84200000)
rgb(0.82492941)=(0.45310000,0.81300000,0.84120000)
rgb(0.82721718)=(0.44300000,0.80700000,0.84020000)
rgb(0.82949297)=(0.43300000,0.80090000,0.83910000)
rgb(0.83175689)=(0.42310000,0.79470000,0.83790000)
rgb(0.83400907)=(0.41330000,0.78850000,0.83660000)
rgb(0.83624963)=(0.40360000,0.78210000,0.83510000)
rgb(0.83847869)=(0.39410000,0.77570000,0.83360000)
rgb(0.84069637)=(0.38470000,0.76930000,0.83200000)
rgb(0.84290278)=(0.37550000,0.76270000,0.83020000)
rgb(0.84509804)=(0.36640000,0.75620000,0.82840000)
rgb(0.84728226)=(0.35760000,0.74960000,0.82650000)
rgb(0.84945555)=(0.34890000,0.74300000,0.82450000)
rgb(0.85161801)=(0.34040000,0.73630000,0.82250000)
rgb(0.85376977)=(0.33220000,0.72960000,0.82040000)
rgb(0.85591091)=(0.32410000,0.72290000,0.81820000)
rgb(0.85804155)=(0.31630000,0.71620000,0.81600000)
rgb(0.86016179)=(0.30870000,0.70950000,0.81370000)
rgb(0.86227172)=(0.30120000,0.70280000,0.81140000)
rgb(0.86437146)=(0.29400000,0.69610000,0.80910000)
rgb(0.86646109)=(0.28710000,0.68940000,0.80670000)
rgb(0.86854072)=(0.28030000,0.68270000,0.80430000)
rgb(0.87061043)=(0.27380000,0.67600000,0.80180000)
rgb(0.87267033)=(0.26750000,0.66940000,0.79930000)
rgb(0.87472051)=(0.26140000,0.66270000,0.79680000)
rgb(0.87676105)=(0.25550000,0.65610000,0.79430000)
rgb(0.87879205)=(0.24990000,0.64950000,0.79180000)
rgb(0.88081359)=(0.24440000,0.64290000,0.78920000)
rgb(0.88282577)=(0.23920000,0.63640000,0.78670000)
rgb(0.88482867)=(0.23410000,0.62990000,0.78410000)
rgb(0.88682237)=(0.22920000,0.62340000,0.78150000)
rgb(0.88880697)=(0.22450000,0.61690000,0.77890000)
rgb(0.89078253)=(0.22010000,0.61050000,0.77630000)
rgb(0.89274915)=(0.21570000,0.60410000,0.77370000)
rgb(0.89470691)=(0.21160000,0.59770000,0.77110000)
rgb(0.89665588)=(0.20760000,0.59130000,0.76850000)
rgb(0.89859614)=(0.20370000,0.58500000,0.76590000)
rgb(0.90052777)=(0.20000000,0.57870000,0.76330000)
rgb(0.90245085)=(0.19640000,0.57240000,0.76070000)
rgb(0.90436545)=(0.19300000,0.56610000,0.75810000)
rgb(0.90627165)=(0.18970000,0.55990000,0.75550000)
rgb(0.90816951)=(0.18650000,0.55370000,0.75290000)
rgb(0.91005912)=(0.18340000,0.54750000,0.75030000)
rgb(0.91194054)=(0.18040000,0.54130000,0.74770000)
rgb(0.91381385)=(0.17760000,0.53520000,0.74510000)
rgb(0.91567911)=(0.17480000,0.52900000,0.74250000)
rgb(0.91753640)=(0.17210000,0.52290000,0.73990000)
rgb(0.91938578)=(0.16950000,0.51680000,0.73730000)
rgb(0.92122731)=(0.16700000,0.51060000,0.73470000)
rgb(0.92306107)=(0.16450000,0.50450000,0.73210000)
rgb(0.92488712)=(0.16210000,0.49840000,0.72950000)
rgb(0.92670552)=(0.15970000,0.49230000,0.72680000)
rgb(0.92851634)=(0.15740000,0.48610000,0.72420000)
rgb(0.93031964)=(0.15520000,0.48000000,0.72160000)
rgb(0.93211549)=(0.15300000,0.47380000,0.71890000)
rgb(0.93390393)=(0.15080000,0.46770000,0.71630000)
rgb(0.93568505)=(0.14870000,0.46150000,0.71360000)
rgb(0.93745889)=(0.14650000,0.45530000,0.71100000)
rgb(0.93922551)=(0.14450000,0.44900000,0.70830000)
rgb(0.94098498)=(0.14240000,0.44280000,0.70560000)
rgb(0.94273735)=(0.14030000,0.43650000,0.70290000)
rgb(0.94448267)=(0.13830000,0.43010000,0.70020000)
rgb(0.94622101)=(0.13620000,0.42380000,0.69740000)
rgb(0.94795242)=(0.13420000,0.41740000,0.69470000)
rgb(0.94967695)=(0.13210000,0.41100000,0.69190000)
rgb(0.95139467)=(0.13010000,0.40450000,0.68910000)
rgb(0.95310561)=(0.12800000,0.39800000,0.68630000)
rgb(0.95480984)=(0.12590000,0.39150000,0.68350000)
rgb(0.95650741)=(0.12370000,0.38490000,0.68060000)
rgb(0.95819837)=(0.12150000,0.37840000,0.67770000)
rgb(0.95988278)=(0.11930000,0.37170000,0.67490000)
rgb(0.96156067)=(0.11700000,0.36500000,0.67200000)
rgb(0.96323211)=(0.11470000,0.35830000,0.66900000)
rgb(0.96489714)=(0.11230000,0.35160000,0.66610000)
rgb(0.96655581)=(0.10990000,0.34480000,0.66320000)
rgb(0.96820817)=(0.10730000,0.33800000,0.66020000)
rgb(0.96985426)=(0.10460000,0.33110000,0.65720000)
rgb(0.97149414)=(0.10200000,0.32420000,0.65420000)
rgb(0.97312785)=(0.09910000,0.31730000,0.65120000)
rgb(0.97475544)=(0.09610000,0.31030000,0.64810000)
rgb(0.97637695)=(0.09300000,0.30330000,0.64500000)
rgb(0.97799244)=(0.08980000,0.29620000,0.64200000)
rgb(0.97960193)=(0.08640000,0.28910000,0.63880000)
rgb(0.98120548)=(0.08280000,0.28200000,0.63570000)
rgb(0.98280313)=(0.07890000,0.27480000,0.63260000)
rgb(0.98439493)=(0.07480000,0.26750000,0.62940000)
rgb(0.98598091)=(0.07040000,0.26020000,0.62620000)
rgb(0.98756112)=(0.06570000,0.25290000,0.62300000)
rgb(0.98913560)=(0.06060000,0.24550000,0.61980000)
rgb(0.99070440)=(0.05500000,0.23800000,0.61650000)
rgb(0.99226755)=(0.04890000,0.23050000,0.61330000)
rgb(0.99382509)=(0.04200000,0.22300000,0.61000000)
rgb(0.99537707)=(0.03410000,0.21530000,0.60670000)
rgb(0.99692352)=(0.02620000,0.20770000,0.60340000)
rgb(0.99846448)=(0.01820000,0.19990000,0.60000000)
rgb(1.00000000)=(0.00980000,0.19210000,0.59670000)},
}
\pgfplotsset{
colormap={plots4}{rgb(0.00000000)=(0.49230000,0.09080000,0.00010000)
rgb(0.01506375)=(0.49670000,0.10280000,0.00370000)
rgb(0.02962247)=(0.50110000,0.11400000,0.00710000)
rgb(0.04370893)=(0.50550000,0.12470000,0.01040000)
rgb(0.05735281)=(0.50980000,0.13490000,0.01380000)
rgb(0.07058107)=(0.51410000,0.14460000,0.01680000)
rgb(0.08341830)=(0.51840000,0.15400000,0.01970000)
rgb(0.09588694)=(0.52260000,0.16320000,0.02250000)
rgb(0.10800757)=(0.52680000,0.17200000,0.02500000)
rgb(0.11979910)=(0.53100000,0.18070000,0.02750000)
rgb(0.13127891)=(0.53510000,0.18910000,0.03010000)
rgb(0.14246308)=(0.53920000,0.19740000,0.03290000)
rgb(0.15336644)=(0.54330000,0.20550000,0.03590000)
rgb(0.16400276)=(0.54730000,0.21350000,0.03890000)
rgb(0.17438480)=(0.55120000,0.22130000,0.04200000)
rgb(0.18452443)=(0.55510000,0.22900000,0.04500000)
rgb(0.19443272)=(0.55900000,0.23660000,0.04800000)
rgb(0.20411998)=(0.56280000,0.24410000,0.05100000)
rgb(0.21359587)=(0.56650000,0.25150000,0.05400000)
rgb(0.22286942)=(0.57030000,0.25870000,0.05700000)
rgb(0.23194908)=(0.57390000,0.26590000,0.06010000)
rgb(0.24084279)=(0.57760000,0.27300000,0.06300000)
rgb(0.24955802)=(0.58120000,0.28000000,0.06590000)
rgb(0.25810180)=(0.58470000,0.28700000,0.06890000)
rgb(0.26648073)=(0.58820000,0.29390000,0.07170000)
rgb(0.27470106)=(0.59170000,0.30070000,0.07460000)
rgb(0.28276868)=(0.59520000,0.30740000,0.07740000)
rgb(0.29068916)=(0.59860000,0.31420000,0.08030000)
rgb(0.29846778)=(0.60190000,0.32080000,0.08310000)
rgb(0.30610952)=(0.60530000,0.32740000,0.08590000)
rgb(0.31361912)=(0.60860000,0.33390000,0.08870000)
rgb(0.32100108)=(0.61190000,0.34040000,0.09150000)
rgb(0.32825965)=(0.61510000,0.34680000,0.09430000)
rgb(0.33539890)=(0.61830000,0.35320000,0.09700000)
rgb(0.34242268)=(0.62150000,0.35960000,0.09980000)
rgb(0.34933468)=(0.62460000,0.36590000,0.10250000)
rgb(0.35613838)=(0.62780000,0.37220000,0.10520000)
rgb(0.36283715)=(0.63090000,0.37840000,0.10800000)
rgb(0.36943415)=(0.63400000,0.38470000,0.11070000)
rgb(0.37593244)=(0.63710000,0.39090000,0.11350000)
rgb(0.38233494)=(0.64010000,0.39710000,0.11620000)
rgb(0.38864441)=(0.64320000,0.40330000,0.11900000)
rgb(0.39486353)=(0.64620000,0.40940000,0.12170000)
rgb(0.40099485)=(0.64920000,0.41560000,0.12450000)
rgb(0.40704081)=(0.65230000,0.42180000,0.12730000)
rgb(0.41300376)=(0.65530000,0.42800000,0.13020000)
rgb(0.41888594)=(0.65830000,0.43410000,0.13300000)
rgb(0.42468951)=(0.66130000,0.44030000,0.13580000)
rgb(0.43041656)=(0.66430000,0.44650000,0.13880000)
rgb(0.43606906)=(0.66740000,0.45280000,0.14180000)
rgb(0.44164894)=(0.67040000,0.45900000,0.14480000)
rgb(0.44715803)=(0.67340000,0.46530000,0.14780000)
rgb(0.45259812)=(0.67650000,0.47160000,0.15100000)
rgb(0.45797090)=(0.67960000,0.47800000,0.15420000)
rgb(0.46327803)=(0.68260000,0.48430000,0.15740000)
rgb(0.46852108)=(0.68570000,0.49080000,0.16080000)
rgb(0.47370160)=(0.68880000,0.49720000,0.16420000)
rgb(0.47882104)=(0.69200000,0.50370000,0.16770000)
rgb(0.48388084)=(0.69510000,0.51030000,0.17140000)
rgb(0.48888237)=(0.69830000,0.51690000,0.17510000)
rgb(0.49382695)=(0.70150000,0.52360000,0.17900000)
rgb(0.49871587)=(0.70470000,0.53030000,0.18290000)
rgb(0.50355037)=(0.70790000,0.53700000,0.18700000)
rgb(0.50833164)=(0.71120000,0.54380000,0.19120000)
rgb(0.51306084)=(0.71440000,0.55070000,0.19560000)
rgb(0.51773911)=(0.71770000,0.55760000,0.20010000)
rgb(0.52236751)=(0.72100000,0.56460000,0.20480000)
rgb(0.52694711)=(0.72440000,0.57160000,0.20970000)
rgb(0.53147892)=(0.72770000,0.57870000,0.21470000)
rgb(0.53596393)=(0.73110000,0.58590000,0.21990000)
rgb(0.54040309)=(0.73450000,0.59310000,0.22530000)
rgb(0.54479734)=(0.73790000,0.60040000,0.23100000)
rgb(0.54914757)=(0.74130000,0.60770000,0.23680000)
rgb(0.55345466)=(0.74470000,0.61510000,0.24280000)
rgb(0.55771945)=(0.74820000,0.62250000,0.24910000)
rgb(0.56194277)=(0.75160000,0.63000000,0.25560000)
rgb(0.56612541)=(0.75510000,0.63760000,0.26230000)
rgb(0.57026816)=(0.75850000,0.64510000,0.26930000)
rgb(0.57437176)=(0.76200000,0.65280000,0.27650000)
rgb(0.57843695)=(0.76540000,0.66050000,0.28400000)
rgb(0.58246444)=(0.76880000,0.66820000,0.29180000)
rgb(0.58645492)=(0.77220000,0.67590000,0.29970000)
rgb(0.59040907)=(0.77560000,0.68370000,0.30800000)
rgb(0.59432754)=(0.77890000,0.69150000,0.31650000)
rgb(0.59821098)=(0.78220000,0.69940000,0.32530000)
rgb(0.60205999)=(0.78550000,0.70720000,0.33430000)
rgb(0.60587519)=(0.78870000,0.71500000,0.34360000)
rgb(0.60965717)=(0.79180000,0.72280000,0.35310000)
rgb(0.61340650)=(0.79490000,0.73070000,0.36280000)
rgb(0.61712374)=(0.79780000,0.73840000,0.37280000)
rgb(0.62080943)=(0.80070000,0.74620000,0.38310000)
rgb(0.62446410)=(0.80350000,0.75390000,0.39350000)
rgb(0.62808828)=(0.80610000,0.76150000,0.40410000)
rgb(0.63168246)=(0.80860000,0.76910000,0.41500000)
rgb(0.63524714)=(0.81100000,0.77660000,0.42590000)
rgb(0.63878280)=(0.81320000,0.78400000,0.43710000)
rgb(0.64228991)=(0.81530000,0.79130000,0.44840000)
rgb(0.64576892)=(0.81720000,0.79850000,0.45980000)
rgb(0.64922028)=(0.81890000,0.80560000,0.47130000)
rgb(0.65264444)=(0.82040000,0.81250000,0.48280000)
rgb(0.65604180)=(0.82170000,0.81920000,0.49440000)
rgb(0.65941280)=(0.82280000,0.82580000,0.50610000)
rgb(0.66275783)=(0.82360000,0.83230000,0.51780000)
rgb(0.66607730)=(0.82420000,0.83850000,0.52940000)
rgb(0.66937158)=(0.82460000,0.84460000,0.54100000)
rgb(0.67264107)=(0.82480000,0.85040000,0.55260000)
rgb(0.67588612)=(0.82460000,0.85600000,0.56400000)
rgb(0.67910711)=(0.82430000,0.86140000,0.57540000)
rgb(0.68230438)=(0.82360000,0.86660000,0.58670000)
rgb(0.68547829)=(0.82270000,0.87160000,0.59780000)
rgb(0.68862917)=(0.82150000,0.87630000,0.60870000)
rgb(0.69175736)=(0.82000000,0.88070000,0.61950000)
rgb(0.69486317)=(0.81830000,0.88500000,0.63010000)
rgb(0.69794693)=(0.81630000,0.88900000,0.64050000)
rgb(0.70100895)=(0.81400000,0.89270000,0.65070000)
rgb(0.70404953)=(0.81140000,0.89620000,0.66060000)
rgb(0.70706897)=(0.80850000,0.89940000,0.67030000)
rgb(0.71006756)=(0.80530000,0.90240000,0.67980000)
rgb(0.71304559)=(0.80190000,0.90510000,0.68900000)
rgb(0.71600334)=(0.79820000,0.90760000,0.69790000)
rgb(0.71894109)=(0.79420000,0.90980000,0.70660000)
rgb(0.72185909)=(0.79000000,0.91180000,0.71500000)
rgb(0.72475762)=(0.78540000,0.91350000,0.72310000)
rgb(0.72763693)=(0.78060000,0.91500000,0.73100000)
rgb(0.73049727)=(0.77560000,0.91620000,0.73850000)
rgb(0.73333891)=(0.77020000,0.91720000,0.74580000)
rgb(0.73616207)=(0.76460000,0.91800000,0.75280000)
rgb(0.73896699)=(0.75880000,0.91850000,0.75950000)
rgb(0.74175392)=(0.75270000,0.91880000,0.76600000)
rgb(0.74452307)=(0.74630000,0.91890000,0.77210000)
rgb(0.74727468)=(0.73970000,0.91870000,0.77800000)
rgb(0.75000897)=(0.73290000,0.91840000,0.78360000)
rgb(0.75272615)=(0.72580000,0.91770000,0.78890000)
rgb(0.75542644)=(0.71850000,0.91690000,0.79390000)
rgb(0.75811004)=(0.71100000,0.91580000,0.79870000)
rgb(0.76077715)=(0.70330000,0.91460000,0.80320000)
rgb(0.76342799)=(0.69540000,0.91310000,0.80750000)
rgb(0.76606275)=(0.68720000,0.91130000,0.81140000)
rgb(0.76868162)=(0.67890000,0.90940000,0.81520000)
rgb(0.77128479)=(0.67040000,0.90730000,0.81860000)
rgb(0.77387245)=(0.66160000,0.90490000,0.82190000)
rgb(0.77644479)=(0.65270000,0.90240000,0.82490000)
rgb(0.77900197)=(0.64370000,0.89960000,0.82760000)
rgb(0.78154419)=(0.63450000,0.89670000,0.83010000)
rgb(0.78407162)=(0.62510000,0.89350000,0.83240000)
rgb(0.78658442)=(0.61560000,0.89010000,0.83450000)
rgb(0.78908276)=(0.60590000,0.88660000,0.83630000)
rgb(0.79156682)=(0.59610000,0.88290000,0.83800000)
rgb(0.79403675)=(0.58630000,0.87890000,0.83940000)
rgb(0.79649271)=(0.57630000,0.87480000,0.84060000)
rgb(0.79893486)=(0.56620000,0.87050000,0.84160000)
rgb(0.80136335)=(0.55600000,0.86610000,0.84250000)
rgb(0.80377834)=(0.54580000,0.86140000,0.84310000)
rgb(0.80617997)=(0.53550000,0.85660000,0.84350000)
rgb(0.80856840)=(0.52520000,0.85170000,0.84380000)
rgb(0.81094376)=(0.51490000,0.84650000,0.84390000)
rgb(0.81330621)=(0.50450000,0.84130000,0.84380000)
rgb(0.81565587)=(0.49420000,0.83590000,0.84360000)
rgb(0.81799288)=(0.48380000,0.83040000,0.84320000)
rgb(0.82031739)=(0.47350000,0.82470000,0.84270000)
rgb(0.82262952)=(0.46330000,0.81890000,0.84200000)
rgb(0.82492941)=(0.45310000,0.81300000,0.84120000)
rgb(0.82721718)=(0.44300000,0.80700000,0.84020000)
rgb(0.82949297)=(0.43300000,0.80090000,0.83910000)
rgb(0.83175689)=(0.42310000,0.79470000,0.83790000)
rgb(0.83400907)=(0.41330000,0.78850000,0.83660000)
rgb(0.83624963)=(0.40360000,0.78210000,0.83510000)
rgb(0.83847869)=(0.39410000,0.77570000,0.83360000)
rgb(0.84069637)=(0.38470000,0.76930000,0.83200000)
rgb(0.84290278)=(0.37550000,0.76270000,0.83020000)
rgb(0.84509804)=(0.36640000,0.75620000,0.82840000)
rgb(0.84728226)=(0.35760000,0.74960000,0.82650000)
rgb(0.84945555)=(0.34890000,0.74300000,0.82450000)
rgb(0.85161801)=(0.34040000,0.73630000,0.82250000)
rgb(0.85376977)=(0.33220000,0.72960000,0.82040000)
rgb(0.85591091)=(0.32410000,0.72290000,0.81820000)
rgb(0.85804155)=(0.31630000,0.71620000,0.81600000)
rgb(0.86016179)=(0.30870000,0.70950000,0.81370000)
rgb(0.86227172)=(0.30120000,0.70280000,0.81140000)
rgb(0.86437146)=(0.29400000,0.69610000,0.80910000)
rgb(0.86646109)=(0.28710000,0.68940000,0.80670000)
rgb(0.86854072)=(0.28030000,0.68270000,0.80430000)
rgb(0.87061043)=(0.27380000,0.67600000,0.80180000)
rgb(0.87267033)=(0.26750000,0.66940000,0.79930000)
rgb(0.87472051)=(0.26140000,0.66270000,0.79680000)
rgb(0.87676105)=(0.25550000,0.65610000,0.79430000)
rgb(0.87879205)=(0.24990000,0.64950000,0.79180000)
rgb(0.88081359)=(0.24440000,0.64290000,0.78920000)
rgb(0.88282577)=(0.23920000,0.63640000,0.78670000)
rgb(0.88482867)=(0.23410000,0.62990000,0.78410000)
rgb(0.88682237)=(0.22920000,0.62340000,0.78150000)
rgb(0.88880697)=(0.22450000,0.61690000,0.77890000)
rgb(0.89078253)=(0.22010000,0.61050000,0.77630000)
rgb(0.89274915)=(0.21570000,0.60410000,0.77370000)
rgb(0.89470691)=(0.21160000,0.59770000,0.77110000)
rgb(0.89665588)=(0.20760000,0.59130000,0.76850000)
rgb(0.89859614)=(0.20370000,0.58500000,0.76590000)
rgb(0.90052777)=(0.20000000,0.57870000,0.76330000)
rgb(0.90245085)=(0.19640000,0.57240000,0.76070000)
rgb(0.90436545)=(0.19300000,0.56610000,0.75810000)
rgb(0.90627165)=(0.18970000,0.55990000,0.75550000)
rgb(0.90816951)=(0.18650000,0.55370000,0.75290000)
rgb(0.91005912)=(0.18340000,0.54750000,0.75030000)
rgb(0.91194054)=(0.18040000,0.54130000,0.74770000)
rgb(0.91381385)=(0.17760000,0.53520000,0.74510000)
rgb(0.91567911)=(0.17480000,0.52900000,0.74250000)
rgb(0.91753640)=(0.17210000,0.52290000,0.73990000)
rgb(0.91938578)=(0.16950000,0.51680000,0.73730000)
rgb(0.92122731)=(0.16700000,0.51060000,0.73470000)
rgb(0.92306107)=(0.16450000,0.50450000,0.73210000)
rgb(0.92488712)=(0.16210000,0.49840000,0.72950000)
rgb(0.92670552)=(0.15970000,0.49230000,0.72680000)
rgb(0.92851634)=(0.15740000,0.48610000,0.72420000)
rgb(0.93031964)=(0.15520000,0.48000000,0.72160000)
rgb(0.93211549)=(0.15300000,0.47380000,0.71890000)
rgb(0.93390393)=(0.15080000,0.46770000,0.71630000)
rgb(0.93568505)=(0.14870000,0.46150000,0.71360000)
rgb(0.93745889)=(0.14650000,0.45530000,0.71100000)
rgb(0.93922551)=(0.14450000,0.44900000,0.70830000)
rgb(0.94098498)=(0.14240000,0.44280000,0.70560000)
rgb(0.94273735)=(0.14030000,0.43650000,0.70290000)
rgb(0.94448267)=(0.13830000,0.43010000,0.70020000)
rgb(0.94622101)=(0.13620000,0.42380000,0.69740000)
rgb(0.94795242)=(0.13420000,0.41740000,0.69470000)
rgb(0.94967695)=(0.13210000,0.41100000,0.69190000)
rgb(0.95139467)=(0.13010000,0.40450000,0.68910000)
rgb(0.95310561)=(0.12800000,0.39800000,0.68630000)
rgb(0.95480984)=(0.12590000,0.39150000,0.68350000)
rgb(0.95650741)=(0.12370000,0.38490000,0.68060000)
rgb(0.95819837)=(0.12150000,0.37840000,0.67770000)
rgb(0.95988278)=(0.11930000,0.37170000,0.67490000)
rgb(0.96156067)=(0.11700000,0.36500000,0.67200000)
rgb(0.96323211)=(0.11470000,0.35830000,0.66900000)
rgb(0.96489714)=(0.11230000,0.35160000,0.66610000)
rgb(0.96655581)=(0.10990000,0.34480000,0.66320000)
rgb(0.96820817)=(0.10730000,0.33800000,0.66020000)
rgb(0.96985426)=(0.10460000,0.33110000,0.65720000)
rgb(0.97149414)=(0.10200000,0.32420000,0.65420000)
rgb(0.97312785)=(0.09910000,0.31730000,0.65120000)
rgb(0.97475544)=(0.09610000,0.31030000,0.64810000)
rgb(0.97637695)=(0.09300000,0.30330000,0.64500000)
rgb(0.97799244)=(0.08980000,0.29620000,0.64200000)
rgb(0.97960193)=(0.08640000,0.28910000,0.63880000)
rgb(0.98120548)=(0.08280000,0.28200000,0.63570000)
rgb(0.98280313)=(0.07890000,0.27480000,0.63260000)
rgb(0.98439493)=(0.07480000,0.26750000,0.62940000)
rgb(0.98598091)=(0.07040000,0.26020000,0.62620000)
rgb(0.98756112)=(0.06570000,0.25290000,0.62300000)
rgb(0.98913560)=(0.06060000,0.24550000,0.61980000)
rgb(0.99070440)=(0.05500000,0.23800000,0.61650000)
rgb(0.99226755)=(0.04890000,0.23050000,0.61330000)
rgb(0.99382509)=(0.04200000,0.22300000,0.61000000)
rgb(0.99537707)=(0.03410000,0.21530000,0.60670000)
rgb(0.99692352)=(0.02620000,0.20770000,0.60340000)
rgb(0.99846448)=(0.01820000,0.19990000,0.60000000)
rgb(1.00000000)=(0.00980000,0.19210000,0.59670000)},
}
\pgfplotsset{
colormap={plots5}{rgb(0.00000000)=(0.49230000,0.09080000,0.00010000)
rgb(0.01506375)=(0.49670000,0.10280000,0.00370000)
rgb(0.02962247)=(0.50110000,0.11400000,0.00710000)
rgb(0.04370893)=(0.50550000,0.12470000,0.01040000)
rgb(0.05735281)=(0.50980000,0.13490000,0.01380000)
rgb(0.07058107)=(0.51410000,0.14460000,0.01680000)
rgb(0.08341830)=(0.51840000,0.15400000,0.01970000)
rgb(0.09588694)=(0.52260000,0.16320000,0.02250000)
rgb(0.10800757)=(0.52680000,0.17200000,0.02500000)
rgb(0.11979910)=(0.53100000,0.18070000,0.02750000)
rgb(0.13127891)=(0.53510000,0.18910000,0.03010000)
rgb(0.14246308)=(0.53920000,0.19740000,0.03290000)
rgb(0.15336644)=(0.54330000,0.20550000,0.03590000)
rgb(0.16400276)=(0.54730000,0.21350000,0.03890000)
rgb(0.17438480)=(0.55120000,0.22130000,0.04200000)
rgb(0.18452443)=(0.55510000,0.22900000,0.04500000)
rgb(0.19443272)=(0.55900000,0.23660000,0.04800000)
rgb(0.20411998)=(0.56280000,0.24410000,0.05100000)
rgb(0.21359587)=(0.56650000,0.25150000,0.05400000)
rgb(0.22286942)=(0.57030000,0.25870000,0.05700000)
rgb(0.23194908)=(0.57390000,0.26590000,0.06010000)
rgb(0.24084279)=(0.57760000,0.27300000,0.06300000)
rgb(0.24955802)=(0.58120000,0.28000000,0.06590000)
rgb(0.25810180)=(0.58470000,0.28700000,0.06890000)
rgb(0.26648073)=(0.58820000,0.29390000,0.07170000)
rgb(0.27470106)=(0.59170000,0.30070000,0.07460000)
rgb(0.28276868)=(0.59520000,0.30740000,0.07740000)
rgb(0.29068916)=(0.59860000,0.31420000,0.08030000)
rgb(0.29846778)=(0.60190000,0.32080000,0.08310000)
rgb(0.30610952)=(0.60530000,0.32740000,0.08590000)
rgb(0.31361912)=(0.60860000,0.33390000,0.08870000)
rgb(0.32100108)=(0.61190000,0.34040000,0.09150000)
rgb(0.32825965)=(0.61510000,0.34680000,0.09430000)
rgb(0.33539890)=(0.61830000,0.35320000,0.09700000)
rgb(0.34242268)=(0.62150000,0.35960000,0.09980000)
rgb(0.34933468)=(0.62460000,0.36590000,0.10250000)
rgb(0.35613838)=(0.62780000,0.37220000,0.10520000)
rgb(0.36283715)=(0.63090000,0.37840000,0.10800000)
rgb(0.36943415)=(0.63400000,0.38470000,0.11070000)
rgb(0.37593244)=(0.63710000,0.39090000,0.11350000)
rgb(0.38233494)=(0.64010000,0.39710000,0.11620000)
rgb(0.38864441)=(0.64320000,0.40330000,0.11900000)
rgb(0.39486353)=(0.64620000,0.40940000,0.12170000)
rgb(0.40099485)=(0.64920000,0.41560000,0.12450000)
rgb(0.40704081)=(0.65230000,0.42180000,0.12730000)
rgb(0.41300376)=(0.65530000,0.42800000,0.13020000)
rgb(0.41888594)=(0.65830000,0.43410000,0.13300000)
rgb(0.42468951)=(0.66130000,0.44030000,0.13580000)
rgb(0.43041656)=(0.66430000,0.44650000,0.13880000)
rgb(0.43606906)=(0.66740000,0.45280000,0.14180000)
rgb(0.44164894)=(0.67040000,0.45900000,0.14480000)
rgb(0.44715803)=(0.67340000,0.46530000,0.14780000)
rgb(0.45259812)=(0.67650000,0.47160000,0.15100000)
rgb(0.45797090)=(0.67960000,0.47800000,0.15420000)
rgb(0.46327803)=(0.68260000,0.48430000,0.15740000)
rgb(0.46852108)=(0.68570000,0.49080000,0.16080000)
rgb(0.47370160)=(0.68880000,0.49720000,0.16420000)
rgb(0.47882104)=(0.69200000,0.50370000,0.16770000)
rgb(0.48388084)=(0.69510000,0.51030000,0.17140000)
rgb(0.48888237)=(0.69830000,0.51690000,0.17510000)
rgb(0.49382695)=(0.70150000,0.52360000,0.17900000)
rgb(0.49871587)=(0.70470000,0.53030000,0.18290000)
rgb(0.50355037)=(0.70790000,0.53700000,0.18700000)
rgb(0.50833164)=(0.71120000,0.54380000,0.19120000)
rgb(0.51306084)=(0.71440000,0.55070000,0.19560000)
rgb(0.51773911)=(0.71770000,0.55760000,0.20010000)
rgb(0.52236751)=(0.72100000,0.56460000,0.20480000)
rgb(0.52694711)=(0.72440000,0.57160000,0.20970000)
rgb(0.53147892)=(0.72770000,0.57870000,0.21470000)
rgb(0.53596393)=(0.73110000,0.58590000,0.21990000)
rgb(0.54040309)=(0.73450000,0.59310000,0.22530000)
rgb(0.54479734)=(0.73790000,0.60040000,0.23100000)
rgb(0.54914757)=(0.74130000,0.60770000,0.23680000)
rgb(0.55345466)=(0.74470000,0.61510000,0.24280000)
rgb(0.55771945)=(0.74820000,0.62250000,0.24910000)
rgb(0.56194277)=(0.75160000,0.63000000,0.25560000)
rgb(0.56612541)=(0.75510000,0.63760000,0.26230000)
rgb(0.57026816)=(0.75850000,0.64510000,0.26930000)
rgb(0.57437176)=(0.76200000,0.65280000,0.27650000)
rgb(0.57843695)=(0.76540000,0.66050000,0.28400000)
rgb(0.58246444)=(0.76880000,0.66820000,0.29180000)
rgb(0.58645492)=(0.77220000,0.67590000,0.29970000)
rgb(0.59040907)=(0.77560000,0.68370000,0.30800000)
rgb(0.59432754)=(0.77890000,0.69150000,0.31650000)
rgb(0.59821098)=(0.78220000,0.69940000,0.32530000)
rgb(0.60205999)=(0.78550000,0.70720000,0.33430000)
rgb(0.60587519)=(0.78870000,0.71500000,0.34360000)
rgb(0.60965717)=(0.79180000,0.72280000,0.35310000)
rgb(0.61340650)=(0.79490000,0.73070000,0.36280000)
rgb(0.61712374)=(0.79780000,0.73840000,0.37280000)
rgb(0.62080943)=(0.80070000,0.74620000,0.38310000)
rgb(0.62446410)=(0.80350000,0.75390000,0.39350000)
rgb(0.62808828)=(0.80610000,0.76150000,0.40410000)
rgb(0.63168246)=(0.80860000,0.76910000,0.41500000)
rgb(0.63524714)=(0.81100000,0.77660000,0.42590000)
rgb(0.63878280)=(0.81320000,0.78400000,0.43710000)
rgb(0.64228991)=(0.81530000,0.79130000,0.44840000)
rgb(0.64576892)=(0.81720000,0.79850000,0.45980000)
rgb(0.64922028)=(0.81890000,0.80560000,0.47130000)
rgb(0.65264444)=(0.82040000,0.81250000,0.48280000)
rgb(0.65604180)=(0.82170000,0.81920000,0.49440000)
rgb(0.65941280)=(0.82280000,0.82580000,0.50610000)
rgb(0.66275783)=(0.82360000,0.83230000,0.51780000)
rgb(0.66607730)=(0.82420000,0.83850000,0.52940000)
rgb(0.66937158)=(0.82460000,0.84460000,0.54100000)
rgb(0.67264107)=(0.82480000,0.85040000,0.55260000)
rgb(0.67588612)=(0.82460000,0.85600000,0.56400000)
rgb(0.67910711)=(0.82430000,0.86140000,0.57540000)
rgb(0.68230438)=(0.82360000,0.86660000,0.58670000)
rgb(0.68547829)=(0.82270000,0.87160000,0.59780000)
rgb(0.68862917)=(0.82150000,0.87630000,0.60870000)
rgb(0.69175736)=(0.82000000,0.88070000,0.61950000)
rgb(0.69486317)=(0.81830000,0.88500000,0.63010000)
rgb(0.69794693)=(0.81630000,0.88900000,0.64050000)
rgb(0.70100895)=(0.81400000,0.89270000,0.65070000)
rgb(0.70404953)=(0.81140000,0.89620000,0.66060000)
rgb(0.70706897)=(0.80850000,0.89940000,0.67030000)
rgb(0.71006756)=(0.80530000,0.90240000,0.67980000)
rgb(0.71304559)=(0.80190000,0.90510000,0.68900000)
rgb(0.71600334)=(0.79820000,0.90760000,0.69790000)
rgb(0.71894109)=(0.79420000,0.90980000,0.70660000)
rgb(0.72185909)=(0.79000000,0.91180000,0.71500000)
rgb(0.72475762)=(0.78540000,0.91350000,0.72310000)
rgb(0.72763693)=(0.78060000,0.91500000,0.73100000)
rgb(0.73049727)=(0.77560000,0.91620000,0.73850000)
rgb(0.73333891)=(0.77020000,0.91720000,0.74580000)
rgb(0.73616207)=(0.76460000,0.91800000,0.75280000)
rgb(0.73896699)=(0.75880000,0.91850000,0.75950000)
rgb(0.74175392)=(0.75270000,0.91880000,0.76600000)
rgb(0.74452307)=(0.74630000,0.91890000,0.77210000)
rgb(0.74727468)=(0.73970000,0.91870000,0.77800000)
rgb(0.75000897)=(0.73290000,0.91840000,0.78360000)
rgb(0.75272615)=(0.72580000,0.91770000,0.78890000)
rgb(0.75542644)=(0.71850000,0.91690000,0.79390000)
rgb(0.75811004)=(0.71100000,0.91580000,0.79870000)
rgb(0.76077715)=(0.70330000,0.91460000,0.80320000)
rgb(0.76342799)=(0.69540000,0.91310000,0.80750000)
rgb(0.76606275)=(0.68720000,0.91130000,0.81140000)
rgb(0.76868162)=(0.67890000,0.90940000,0.81520000)
rgb(0.77128479)=(0.67040000,0.90730000,0.81860000)
rgb(0.77387245)=(0.66160000,0.90490000,0.82190000)
rgb(0.77644479)=(0.65270000,0.90240000,0.82490000)
rgb(0.77900197)=(0.64370000,0.89960000,0.82760000)
rgb(0.78154419)=(0.63450000,0.89670000,0.83010000)
rgb(0.78407162)=(0.62510000,0.89350000,0.83240000)
rgb(0.78658442)=(0.61560000,0.89010000,0.83450000)
rgb(0.78908276)=(0.60590000,0.88660000,0.83630000)
rgb(0.79156682)=(0.59610000,0.88290000,0.83800000)
rgb(0.79403675)=(0.58630000,0.87890000,0.83940000)
rgb(0.79649271)=(0.57630000,0.87480000,0.84060000)
rgb(0.79893486)=(0.56620000,0.87050000,0.84160000)
rgb(0.80136335)=(0.55600000,0.86610000,0.84250000)
rgb(0.80377834)=(0.54580000,0.86140000,0.84310000)
rgb(0.80617997)=(0.53550000,0.85660000,0.84350000)
rgb(0.80856840)=(0.52520000,0.85170000,0.84380000)
rgb(0.81094376)=(0.51490000,0.84650000,0.84390000)
rgb(0.81330621)=(0.50450000,0.84130000,0.84380000)
rgb(0.81565587)=(0.49420000,0.83590000,0.84360000)
rgb(0.81799288)=(0.48380000,0.83040000,0.84320000)
rgb(0.82031739)=(0.47350000,0.82470000,0.84270000)
rgb(0.82262952)=(0.46330000,0.81890000,0.84200000)
rgb(0.82492941)=(0.45310000,0.81300000,0.84120000)
rgb(0.82721718)=(0.44300000,0.80700000,0.84020000)
rgb(0.82949297)=(0.43300000,0.80090000,0.83910000)
rgb(0.83175689)=(0.42310000,0.79470000,0.83790000)
rgb(0.83400907)=(0.41330000,0.78850000,0.83660000)
rgb(0.83624963)=(0.40360000,0.78210000,0.83510000)
rgb(0.83847869)=(0.39410000,0.77570000,0.83360000)
rgb(0.84069637)=(0.38470000,0.76930000,0.83200000)
rgb(0.84290278)=(0.37550000,0.76270000,0.83020000)
rgb(0.84509804)=(0.36640000,0.75620000,0.82840000)
rgb(0.84728226)=(0.35760000,0.74960000,0.82650000)
rgb(0.84945555)=(0.34890000,0.74300000,0.82450000)
rgb(0.85161801)=(0.34040000,0.73630000,0.82250000)
rgb(0.85376977)=(0.33220000,0.72960000,0.82040000)
rgb(0.85591091)=(0.32410000,0.72290000,0.81820000)
rgb(0.85804155)=(0.31630000,0.71620000,0.81600000)
rgb(0.86016179)=(0.30870000,0.70950000,0.81370000)
rgb(0.86227172)=(0.30120000,0.70280000,0.81140000)
rgb(0.86437146)=(0.29400000,0.69610000,0.80910000)
rgb(0.86646109)=(0.28710000,0.68940000,0.80670000)
rgb(0.86854072)=(0.28030000,0.68270000,0.80430000)
rgb(0.87061043)=(0.27380000,0.67600000,0.80180000)
rgb(0.87267033)=(0.26750000,0.66940000,0.79930000)
rgb(0.87472051)=(0.26140000,0.66270000,0.79680000)
rgb(0.87676105)=(0.25550000,0.65610000,0.79430000)
rgb(0.87879205)=(0.24990000,0.64950000,0.79180000)
rgb(0.88081359)=(0.24440000,0.64290000,0.78920000)
rgb(0.88282577)=(0.23920000,0.63640000,0.78670000)
rgb(0.88482867)=(0.23410000,0.62990000,0.78410000)
rgb(0.88682237)=(0.22920000,0.62340000,0.78150000)
rgb(0.88880697)=(0.22450000,0.61690000,0.77890000)
rgb(0.89078253)=(0.22010000,0.61050000,0.77630000)
rgb(0.89274915)=(0.21570000,0.60410000,0.77370000)
rgb(0.89470691)=(0.21160000,0.59770000,0.77110000)
rgb(0.89665588)=(0.20760000,0.59130000,0.76850000)
rgb(0.89859614)=(0.20370000,0.58500000,0.76590000)
rgb(0.90052777)=(0.20000000,0.57870000,0.76330000)
rgb(0.90245085)=(0.19640000,0.57240000,0.76070000)
rgb(0.90436545)=(0.19300000,0.56610000,0.75810000)
rgb(0.90627165)=(0.18970000,0.55990000,0.75550000)
rgb(0.90816951)=(0.18650000,0.55370000,0.75290000)
rgb(0.91005912)=(0.18340000,0.54750000,0.75030000)
rgb(0.91194054)=(0.18040000,0.54130000,0.74770000)
rgb(0.91381385)=(0.17760000,0.53520000,0.74510000)
rgb(0.91567911)=(0.17480000,0.52900000,0.74250000)
rgb(0.91753640)=(0.17210000,0.52290000,0.73990000)
rgb(0.91938578)=(0.16950000,0.51680000,0.73730000)
rgb(0.92122731)=(0.16700000,0.51060000,0.73470000)
rgb(0.92306107)=(0.16450000,0.50450000,0.73210000)
rgb(0.92488712)=(0.16210000,0.49840000,0.72950000)
rgb(0.92670552)=(0.15970000,0.49230000,0.72680000)
rgb(0.92851634)=(0.15740000,0.48610000,0.72420000)
rgb(0.93031964)=(0.15520000,0.48000000,0.72160000)
rgb(0.93211549)=(0.15300000,0.47380000,0.71890000)
rgb(0.93390393)=(0.15080000,0.46770000,0.71630000)
rgb(0.93568505)=(0.14870000,0.46150000,0.71360000)
rgb(0.93745889)=(0.14650000,0.45530000,0.71100000)
rgb(0.93922551)=(0.14450000,0.44900000,0.70830000)
rgb(0.94098498)=(0.14240000,0.44280000,0.70560000)
rgb(0.94273735)=(0.14030000,0.43650000,0.70290000)
rgb(0.94448267)=(0.13830000,0.43010000,0.70020000)
rgb(0.94622101)=(0.13620000,0.42380000,0.69740000)
rgb(0.94795242)=(0.13420000,0.41740000,0.69470000)
rgb(0.94967695)=(0.13210000,0.41100000,0.69190000)
rgb(0.95139467)=(0.13010000,0.40450000,0.68910000)
rgb(0.95310561)=(0.12800000,0.39800000,0.68630000)
rgb(0.95480984)=(0.12590000,0.39150000,0.68350000)
rgb(0.95650741)=(0.12370000,0.38490000,0.68060000)
rgb(0.95819837)=(0.12150000,0.37840000,0.67770000)
rgb(0.95988278)=(0.11930000,0.37170000,0.67490000)
rgb(0.96156067)=(0.11700000,0.36500000,0.67200000)
rgb(0.96323211)=(0.11470000,0.35830000,0.66900000)
rgb(0.96489714)=(0.11230000,0.35160000,0.66610000)
rgb(0.96655581)=(0.10990000,0.34480000,0.66320000)
rgb(0.96820817)=(0.10730000,0.33800000,0.66020000)
rgb(0.96985426)=(0.10460000,0.33110000,0.65720000)
rgb(0.97149414)=(0.10200000,0.32420000,0.65420000)
rgb(0.97312785)=(0.09910000,0.31730000,0.65120000)
rgb(0.97475544)=(0.09610000,0.31030000,0.64810000)
rgb(0.97637695)=(0.09300000,0.30330000,0.64500000)
rgb(0.97799244)=(0.08980000,0.29620000,0.64200000)
rgb(0.97960193)=(0.08640000,0.28910000,0.63880000)
rgb(0.98120548)=(0.08280000,0.28200000,0.63570000)
rgb(0.98280313)=(0.07890000,0.27480000,0.63260000)
rgb(0.98439493)=(0.07480000,0.26750000,0.62940000)
rgb(0.98598091)=(0.07040000,0.26020000,0.62620000)
rgb(0.98756112)=(0.06570000,0.25290000,0.62300000)
rgb(0.98913560)=(0.06060000,0.24550000,0.61980000)
rgb(0.99070440)=(0.05500000,0.23800000,0.61650000)
rgb(0.99226755)=(0.04890000,0.23050000,0.61330000)
rgb(0.99382509)=(0.04200000,0.22300000,0.61000000)
rgb(0.99537707)=(0.03410000,0.21530000,0.60670000)
rgb(0.99692352)=(0.02620000,0.20770000,0.60340000)
rgb(0.99846448)=(0.01820000,0.19990000,0.60000000)
rgb(1.00000000)=(0.00980000,0.19210000,0.59670000)},
}
\pgfplotsset{
colormap={plots6}{rgb(0.00000000)=(0.49230000,0.09080000,0.00010000)
rgb(0.01506375)=(0.49670000,0.10280000,0.00370000)
rgb(0.02962247)=(0.50110000,0.11400000,0.00710000)
rgb(0.04370893)=(0.50550000,0.12470000,0.01040000)
rgb(0.05735281)=(0.50980000,0.13490000,0.01380000)
rgb(0.07058107)=(0.51410000,0.14460000,0.01680000)
rgb(0.08341830)=(0.51840000,0.15400000,0.01970000)
rgb(0.09588694)=(0.52260000,0.16320000,0.02250000)
rgb(0.10800757)=(0.52680000,0.17200000,0.02500000)
rgb(0.11979910)=(0.53100000,0.18070000,0.02750000)
rgb(0.13127891)=(0.53510000,0.18910000,0.03010000)
rgb(0.14246308)=(0.53920000,0.19740000,0.03290000)
rgb(0.15336644)=(0.54330000,0.20550000,0.03590000)
rgb(0.16400276)=(0.54730000,0.21350000,0.03890000)
rgb(0.17438480)=(0.55120000,0.22130000,0.04200000)
rgb(0.18452443)=(0.55510000,0.22900000,0.04500000)
rgb(0.19443272)=(0.55900000,0.23660000,0.04800000)
rgb(0.20411998)=(0.56280000,0.24410000,0.05100000)
rgb(0.21359587)=(0.56650000,0.25150000,0.05400000)
rgb(0.22286942)=(0.57030000,0.25870000,0.05700000)
rgb(0.23194908)=(0.57390000,0.26590000,0.06010000)
rgb(0.24084279)=(0.57760000,0.27300000,0.06300000)
rgb(0.24955802)=(0.58120000,0.28000000,0.06590000)
rgb(0.25810180)=(0.58470000,0.28700000,0.06890000)
rgb(0.26648073)=(0.58820000,0.29390000,0.07170000)
rgb(0.27470106)=(0.59170000,0.30070000,0.07460000)
rgb(0.28276868)=(0.59520000,0.30740000,0.07740000)
rgb(0.29068916)=(0.59860000,0.31420000,0.08030000)
rgb(0.29846778)=(0.60190000,0.32080000,0.08310000)
rgb(0.30610952)=(0.60530000,0.32740000,0.08590000)
rgb(0.31361912)=(0.60860000,0.33390000,0.08870000)
rgb(0.32100108)=(0.61190000,0.34040000,0.09150000)
rgb(0.32825965)=(0.61510000,0.34680000,0.09430000)
rgb(0.33539890)=(0.61830000,0.35320000,0.09700000)
rgb(0.34242268)=(0.62150000,0.35960000,0.09980000)
rgb(0.34933468)=(0.62460000,0.36590000,0.10250000)
rgb(0.35613838)=(0.62780000,0.37220000,0.10520000)
rgb(0.36283715)=(0.63090000,0.37840000,0.10800000)
rgb(0.36943415)=(0.63400000,0.38470000,0.11070000)
rgb(0.37593244)=(0.63710000,0.39090000,0.11350000)
rgb(0.38233494)=(0.64010000,0.39710000,0.11620000)
rgb(0.38864441)=(0.64320000,0.40330000,0.11900000)
rgb(0.39486353)=(0.64620000,0.40940000,0.12170000)
rgb(0.40099485)=(0.64920000,0.41560000,0.12450000)
rgb(0.40704081)=(0.65230000,0.42180000,0.12730000)
rgb(0.41300376)=(0.65530000,0.42800000,0.13020000)
rgb(0.41888594)=(0.65830000,0.43410000,0.13300000)
rgb(0.42468951)=(0.66130000,0.44030000,0.13580000)
rgb(0.43041656)=(0.66430000,0.44650000,0.13880000)
rgb(0.43606906)=(0.66740000,0.45280000,0.14180000)
rgb(0.44164894)=(0.67040000,0.45900000,0.14480000)
rgb(0.44715803)=(0.67340000,0.46530000,0.14780000)
rgb(0.45259812)=(0.67650000,0.47160000,0.15100000)
rgb(0.45797090)=(0.67960000,0.47800000,0.15420000)
rgb(0.46327803)=(0.68260000,0.48430000,0.15740000)
rgb(0.46852108)=(0.68570000,0.49080000,0.16080000)
rgb(0.47370160)=(0.68880000,0.49720000,0.16420000)
rgb(0.47882104)=(0.69200000,0.50370000,0.16770000)
rgb(0.48388084)=(0.69510000,0.51030000,0.17140000)
rgb(0.48888237)=(0.69830000,0.51690000,0.17510000)
rgb(0.49382695)=(0.70150000,0.52360000,0.17900000)
rgb(0.49871587)=(0.70470000,0.53030000,0.18290000)
rgb(0.50355037)=(0.70790000,0.53700000,0.18700000)
rgb(0.50833164)=(0.71120000,0.54380000,0.19120000)
rgb(0.51306084)=(0.71440000,0.55070000,0.19560000)
rgb(0.51773911)=(0.71770000,0.55760000,0.20010000)
rgb(0.52236751)=(0.72100000,0.56460000,0.20480000)
rgb(0.52694711)=(0.72440000,0.57160000,0.20970000)
rgb(0.53147892)=(0.72770000,0.57870000,0.21470000)
rgb(0.53596393)=(0.73110000,0.58590000,0.21990000)
rgb(0.54040309)=(0.73450000,0.59310000,0.22530000)
rgb(0.54479734)=(0.73790000,0.60040000,0.23100000)
rgb(0.54914757)=(0.74130000,0.60770000,0.23680000)
rgb(0.55345466)=(0.74470000,0.61510000,0.24280000)
rgb(0.55771945)=(0.74820000,0.62250000,0.24910000)
rgb(0.56194277)=(0.75160000,0.63000000,0.25560000)
rgb(0.56612541)=(0.75510000,0.63760000,0.26230000)
rgb(0.57026816)=(0.75850000,0.64510000,0.26930000)
rgb(0.57437176)=(0.76200000,0.65280000,0.27650000)
rgb(0.57843695)=(0.76540000,0.66050000,0.28400000)
rgb(0.58246444)=(0.76880000,0.66820000,0.29180000)
rgb(0.58645492)=(0.77220000,0.67590000,0.29970000)
rgb(0.59040907)=(0.77560000,0.68370000,0.30800000)
rgb(0.59432754)=(0.77890000,0.69150000,0.31650000)
rgb(0.59821098)=(0.78220000,0.69940000,0.32530000)
rgb(0.60205999)=(0.78550000,0.70720000,0.33430000)
rgb(0.60587519)=(0.78870000,0.71500000,0.34360000)
rgb(0.60965717)=(0.79180000,0.72280000,0.35310000)
rgb(0.61340650)=(0.79490000,0.73070000,0.36280000)
rgb(0.61712374)=(0.79780000,0.73840000,0.37280000)
rgb(0.62080943)=(0.80070000,0.74620000,0.38310000)
rgb(0.62446410)=(0.80350000,0.75390000,0.39350000)
rgb(0.62808828)=(0.80610000,0.76150000,0.40410000)
rgb(0.63168246)=(0.80860000,0.76910000,0.41500000)
rgb(0.63524714)=(0.81100000,0.77660000,0.42590000)
rgb(0.63878280)=(0.81320000,0.78400000,0.43710000)
rgb(0.64228991)=(0.81530000,0.79130000,0.44840000)
rgb(0.64576892)=(0.81720000,0.79850000,0.45980000)
rgb(0.64922028)=(0.81890000,0.80560000,0.47130000)
rgb(0.65264444)=(0.82040000,0.81250000,0.48280000)
rgb(0.65604180)=(0.82170000,0.81920000,0.49440000)
rgb(0.65941280)=(0.82280000,0.82580000,0.50610000)
rgb(0.66275783)=(0.82360000,0.83230000,0.51780000)
rgb(0.66607730)=(0.82420000,0.83850000,0.52940000)
rgb(0.66937158)=(0.82460000,0.84460000,0.54100000)
rgb(0.67264107)=(0.82480000,0.85040000,0.55260000)
rgb(0.67588612)=(0.82460000,0.85600000,0.56400000)
rgb(0.67910711)=(0.82430000,0.86140000,0.57540000)
rgb(0.68230438)=(0.82360000,0.86660000,0.58670000)
rgb(0.68547829)=(0.82270000,0.87160000,0.59780000)
rgb(0.68862917)=(0.82150000,0.87630000,0.60870000)
rgb(0.69175736)=(0.82000000,0.88070000,0.61950000)
rgb(0.69486317)=(0.81830000,0.88500000,0.63010000)
rgb(0.69794693)=(0.81630000,0.88900000,0.64050000)
rgb(0.70100895)=(0.81400000,0.89270000,0.65070000)
rgb(0.70404953)=(0.81140000,0.89620000,0.66060000)
rgb(0.70706897)=(0.80850000,0.89940000,0.67030000)
rgb(0.71006756)=(0.80530000,0.90240000,0.67980000)
rgb(0.71304559)=(0.80190000,0.90510000,0.68900000)
rgb(0.71600334)=(0.79820000,0.90760000,0.69790000)
rgb(0.71894109)=(0.79420000,0.90980000,0.70660000)
rgb(0.72185909)=(0.79000000,0.91180000,0.71500000)
rgb(0.72475762)=(0.78540000,0.91350000,0.72310000)
rgb(0.72763693)=(0.78060000,0.91500000,0.73100000)
rgb(0.73049727)=(0.77560000,0.91620000,0.73850000)
rgb(0.73333891)=(0.77020000,0.91720000,0.74580000)
rgb(0.73616207)=(0.76460000,0.91800000,0.75280000)
rgb(0.73896699)=(0.75880000,0.91850000,0.75950000)
rgb(0.74175392)=(0.75270000,0.91880000,0.76600000)
rgb(0.74452307)=(0.74630000,0.91890000,0.77210000)
rgb(0.74727468)=(0.73970000,0.91870000,0.77800000)
rgb(0.75000897)=(0.73290000,0.91840000,0.78360000)
rgb(0.75272615)=(0.72580000,0.91770000,0.78890000)
rgb(0.75542644)=(0.71850000,0.91690000,0.79390000)
rgb(0.75811004)=(0.71100000,0.91580000,0.79870000)
rgb(0.76077715)=(0.70330000,0.91460000,0.80320000)
rgb(0.76342799)=(0.69540000,0.91310000,0.80750000)
rgb(0.76606275)=(0.68720000,0.91130000,0.81140000)
rgb(0.76868162)=(0.67890000,0.90940000,0.81520000)
rgb(0.77128479)=(0.67040000,0.90730000,0.81860000)
rgb(0.77387245)=(0.66160000,0.90490000,0.82190000)
rgb(0.77644479)=(0.65270000,0.90240000,0.82490000)
rgb(0.77900197)=(0.64370000,0.89960000,0.82760000)
rgb(0.78154419)=(0.63450000,0.89670000,0.83010000)
rgb(0.78407162)=(0.62510000,0.89350000,0.83240000)
rgb(0.78658442)=(0.61560000,0.89010000,0.83450000)
rgb(0.78908276)=(0.60590000,0.88660000,0.83630000)
rgb(0.79156682)=(0.59610000,0.88290000,0.83800000)
rgb(0.79403675)=(0.58630000,0.87890000,0.83940000)
rgb(0.79649271)=(0.57630000,0.87480000,0.84060000)
rgb(0.79893486)=(0.56620000,0.87050000,0.84160000)
rgb(0.80136335)=(0.55600000,0.86610000,0.84250000)
rgb(0.80377834)=(0.54580000,0.86140000,0.84310000)
rgb(0.80617997)=(0.53550000,0.85660000,0.84350000)
rgb(0.80856840)=(0.52520000,0.85170000,0.84380000)
rgb(0.81094376)=(0.51490000,0.84650000,0.84390000)
rgb(0.81330621)=(0.50450000,0.84130000,0.84380000)
rgb(0.81565587)=(0.49420000,0.83590000,0.84360000)
rgb(0.81799288)=(0.48380000,0.83040000,0.84320000)
rgb(0.82031739)=(0.47350000,0.82470000,0.84270000)
rgb(0.82262952)=(0.46330000,0.81890000,0.84200000)
rgb(0.82492941)=(0.45310000,0.81300000,0.84120000)
rgb(0.82721718)=(0.44300000,0.80700000,0.84020000)
rgb(0.82949297)=(0.43300000,0.80090000,0.83910000)
rgb(0.83175689)=(0.42310000,0.79470000,0.83790000)
rgb(0.83400907)=(0.41330000,0.78850000,0.83660000)
rgb(0.83624963)=(0.40360000,0.78210000,0.83510000)
rgb(0.83847869)=(0.39410000,0.77570000,0.83360000)
rgb(0.84069637)=(0.38470000,0.76930000,0.83200000)
rgb(0.84290278)=(0.37550000,0.76270000,0.83020000)
rgb(0.84509804)=(0.36640000,0.75620000,0.82840000)
rgb(0.84728226)=(0.35760000,0.74960000,0.82650000)
rgb(0.84945555)=(0.34890000,0.74300000,0.82450000)
rgb(0.85161801)=(0.34040000,0.73630000,0.82250000)
rgb(0.85376977)=(0.33220000,0.72960000,0.82040000)
rgb(0.85591091)=(0.32410000,0.72290000,0.81820000)
rgb(0.85804155)=(0.31630000,0.71620000,0.81600000)
rgb(0.86016179)=(0.30870000,0.70950000,0.81370000)
rgb(0.86227172)=(0.30120000,0.70280000,0.81140000)
rgb(0.86437146)=(0.29400000,0.69610000,0.80910000)
rgb(0.86646109)=(0.28710000,0.68940000,0.80670000)
rgb(0.86854072)=(0.28030000,0.68270000,0.80430000)
rgb(0.87061043)=(0.27380000,0.67600000,0.80180000)
rgb(0.87267033)=(0.26750000,0.66940000,0.79930000)
rgb(0.87472051)=(0.26140000,0.66270000,0.79680000)
rgb(0.87676105)=(0.25550000,0.65610000,0.79430000)
rgb(0.87879205)=(0.24990000,0.64950000,0.79180000)
rgb(0.88081359)=(0.24440000,0.64290000,0.78920000)
rgb(0.88282577)=(0.23920000,0.63640000,0.78670000)
rgb(0.88482867)=(0.23410000,0.62990000,0.78410000)
rgb(0.88682237)=(0.22920000,0.62340000,0.78150000)
rgb(0.88880697)=(0.22450000,0.61690000,0.77890000)
rgb(0.89078253)=(0.22010000,0.61050000,0.77630000)
rgb(0.89274915)=(0.21570000,0.60410000,0.77370000)
rgb(0.89470691)=(0.21160000,0.59770000,0.77110000)
rgb(0.89665588)=(0.20760000,0.59130000,0.76850000)
rgb(0.89859614)=(0.20370000,0.58500000,0.76590000)
rgb(0.90052777)=(0.20000000,0.57870000,0.76330000)
rgb(0.90245085)=(0.19640000,0.57240000,0.76070000)
rgb(0.90436545)=(0.19300000,0.56610000,0.75810000)
rgb(0.90627165)=(0.18970000,0.55990000,0.75550000)
rgb(0.90816951)=(0.18650000,0.55370000,0.75290000)
rgb(0.91005912)=(0.18340000,0.54750000,0.75030000)
rgb(0.91194054)=(0.18040000,0.54130000,0.74770000)
rgb(0.91381385)=(0.17760000,0.53520000,0.74510000)
rgb(0.91567911)=(0.17480000,0.52900000,0.74250000)
rgb(0.91753640)=(0.17210000,0.52290000,0.73990000)
rgb(0.91938578)=(0.16950000,0.51680000,0.73730000)
rgb(0.92122731)=(0.16700000,0.51060000,0.73470000)
rgb(0.92306107)=(0.16450000,0.50450000,0.73210000)
rgb(0.92488712)=(0.16210000,0.49840000,0.72950000)
rgb(0.92670552)=(0.15970000,0.49230000,0.72680000)
rgb(0.92851634)=(0.15740000,0.48610000,0.72420000)
rgb(0.93031964)=(0.15520000,0.48000000,0.72160000)
rgb(0.93211549)=(0.15300000,0.47380000,0.71890000)
rgb(0.93390393)=(0.15080000,0.46770000,0.71630000)
rgb(0.93568505)=(0.14870000,0.46150000,0.71360000)
rgb(0.93745889)=(0.14650000,0.45530000,0.71100000)
rgb(0.93922551)=(0.14450000,0.44900000,0.70830000)
rgb(0.94098498)=(0.14240000,0.44280000,0.70560000)
rgb(0.94273735)=(0.14030000,0.43650000,0.70290000)
rgb(0.94448267)=(0.13830000,0.43010000,0.70020000)
rgb(0.94622101)=(0.13620000,0.42380000,0.69740000)
rgb(0.94795242)=(0.13420000,0.41740000,0.69470000)
rgb(0.94967695)=(0.13210000,0.41100000,0.69190000)
rgb(0.95139467)=(0.13010000,0.40450000,0.68910000)
rgb(0.95310561)=(0.12800000,0.39800000,0.68630000)
rgb(0.95480984)=(0.12590000,0.39150000,0.68350000)
rgb(0.95650741)=(0.12370000,0.38490000,0.68060000)
rgb(0.95819837)=(0.12150000,0.37840000,0.67770000)
rgb(0.95988278)=(0.11930000,0.37170000,0.67490000)
rgb(0.96156067)=(0.11700000,0.36500000,0.67200000)
rgb(0.96323211)=(0.11470000,0.35830000,0.66900000)
rgb(0.96489714)=(0.11230000,0.35160000,0.66610000)
rgb(0.96655581)=(0.10990000,0.34480000,0.66320000)
rgb(0.96820817)=(0.10730000,0.33800000,0.66020000)
rgb(0.96985426)=(0.10460000,0.33110000,0.65720000)
rgb(0.97149414)=(0.10200000,0.32420000,0.65420000)
rgb(0.97312785)=(0.09910000,0.31730000,0.65120000)
rgb(0.97475544)=(0.09610000,0.31030000,0.64810000)
rgb(0.97637695)=(0.09300000,0.30330000,0.64500000)
rgb(0.97799244)=(0.08980000,0.29620000,0.64200000)
rgb(0.97960193)=(0.08640000,0.28910000,0.63880000)
rgb(0.98120548)=(0.08280000,0.28200000,0.63570000)
rgb(0.98280313)=(0.07890000,0.27480000,0.63260000)
rgb(0.98439493)=(0.07480000,0.26750000,0.62940000)
rgb(0.98598091)=(0.07040000,0.26020000,0.62620000)
rgb(0.98756112)=(0.06570000,0.25290000,0.62300000)
rgb(0.98913560)=(0.06060000,0.24550000,0.61980000)
rgb(0.99070440)=(0.05500000,0.23800000,0.61650000)
rgb(0.99226755)=(0.04890000,0.23050000,0.61330000)
rgb(0.99382509)=(0.04200000,0.22300000,0.61000000)
rgb(0.99537707)=(0.03410000,0.21530000,0.60670000)
rgb(0.99692352)=(0.02620000,0.20770000,0.60340000)
rgb(0.99846448)=(0.01820000,0.19990000,0.60000000)
rgb(1.00000000)=(0.00980000,0.19210000,0.59670000)},
}

%% file: figures/nn_structure.tex
\begin{tikzpicture}

\coordinate (r) at (0,2.73);
\node[inner sep=0pt] (img) at ($(r)+(0.0,-2.73)$)
{\includegraphics[width=1.0\columnwidth]{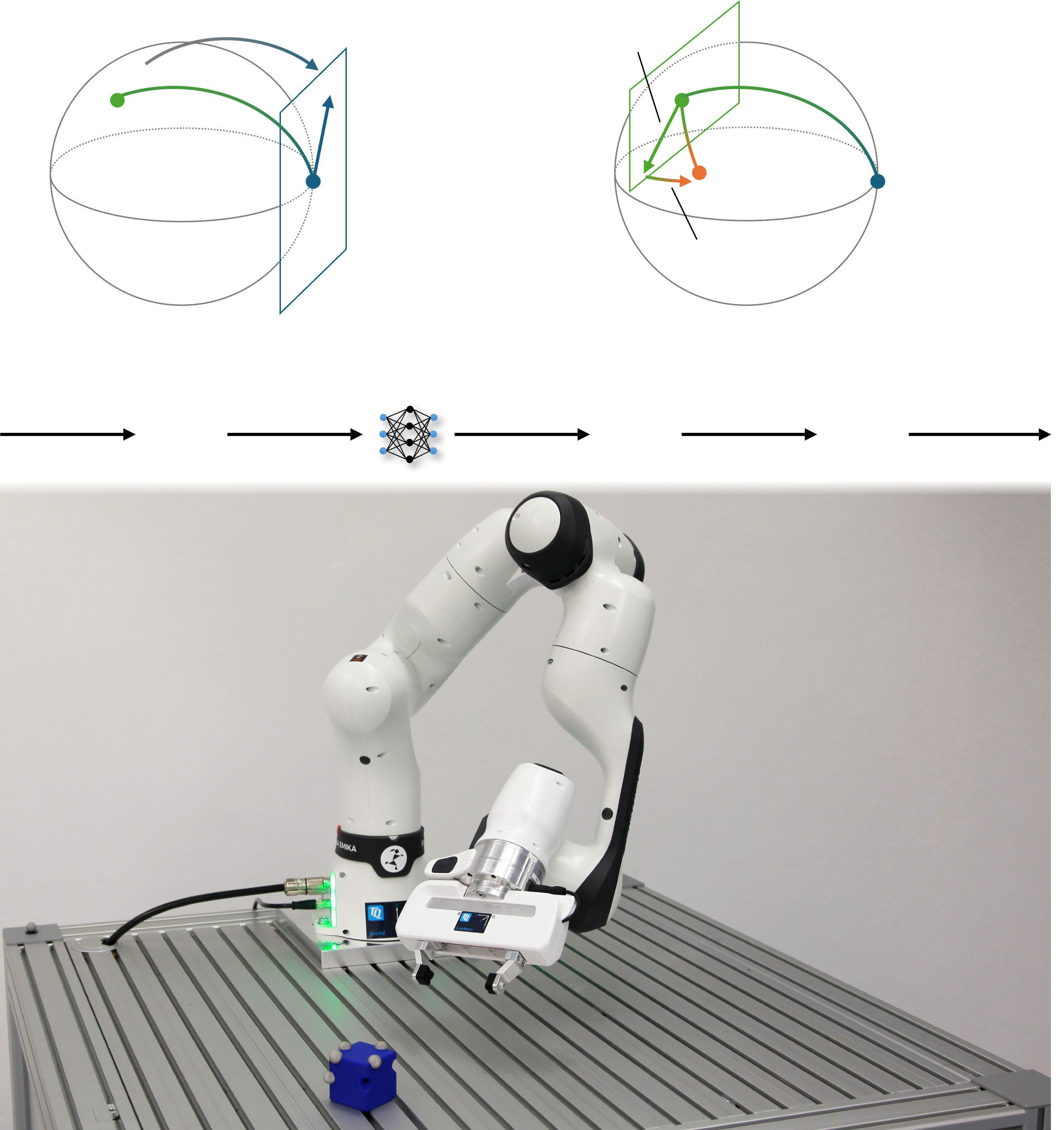}};

\node at ($(r)+(-1.0,-0.0)$) {\textcolor{myblue}{$T_{\bs{\mathcal{E}}}\mathcal{M}$}};
\node at ($(r)+(-3.25,1.25)$) {$\textcolor{mygray}{\bs{s}}$};
\node at ($(r)+(-2.2,1.7)$) {$\Log$};
\node at ($(r)+(-1.3,1.0)$) {\textcolor{myblue}{$\prescript{\bs{\mathcal{E}}}{}{\bs{\tau}}_{\bs{s}}$}};
\node at ($(r)+(-1.8,0.1)$) {$\textcolor{mygray}{\bs{\mathcal{E}}}$};
\node at ($(r)+(-3.8,-0.5)$) {$\textcolor{mygray}{\mathcal{M}}$};
\node at ($(r)+(2.2,1.7)$) {\textcolor{mygreen}{$T_{\bs{s}}\mathcal{M}$}};
\node at ($(r)+(0.8,1.6)$) {\textcolor{mygreen}{$\prescript{\bs{s}}{}{\bs{\tau}}_{\bs{a}}$}};
\node at ($(r)+(1.65,0.5)$) {$\textcolor{mygray}{\bs{s'}}$};
\node at ($(r)+(1.65,-0.3)$) {$\Exp$};
\node at ($(r)+(0.8,-0.5)$) {$\textcolor{mygray}{\mathcal{M}}$};

\node at ($(r)+(-3.8,-1.35)$) {$\textcolor{mygray}{\bs{s}}\in\textcolor{mygray}{\mathcal{M}}$};
\node at ($(r)+(-1.9,-1.35)$) {$\textcolor{myblue}{\prescript{\bs{\mathcal{E}}}{}{\bs{\tau}}_{\bs{s}}}\in\textcolor{myblue}{\mathbb{R}^3}$};
\node at ($(r)+(-0.1,-1.35)$) {$\textcolor{mygreen}{\prescript{\bs{s}}{}{\bs{\tau}}_{\bs{a}}}\in\textcolor{mygreen}{\mathbb{R}^3}$};
\node at ($(r)+(1.73,-1.35)$) {$\textcolor{mygray}{\bs{a}}\in\textcolor{mygray}{\mathcal{M}}$};
\node at ($(r)+(3.6,-1.35)$) {$\textcolor{mygray}{\bs{s}'}\in\textcolor{mygray}{\mathcal{M}}$};

\node at ($(r)+(-2.84,-1.6)$) {$\Log$};
\node at ($(r)+(0.84,-1.6)$) {$\Exp$};
\node at ($(r)+(2.67,-1.6)$) {$\bs{s}\cdot\bs{a}$};

\end{tikzpicture}

%% file: figures/orientation_comp.tex
\begin{tikzpicture}

\node[inner sep=0pt] (img) at (-0.05,0.0)
{\includegraphics[width=0.985\columnwidth]{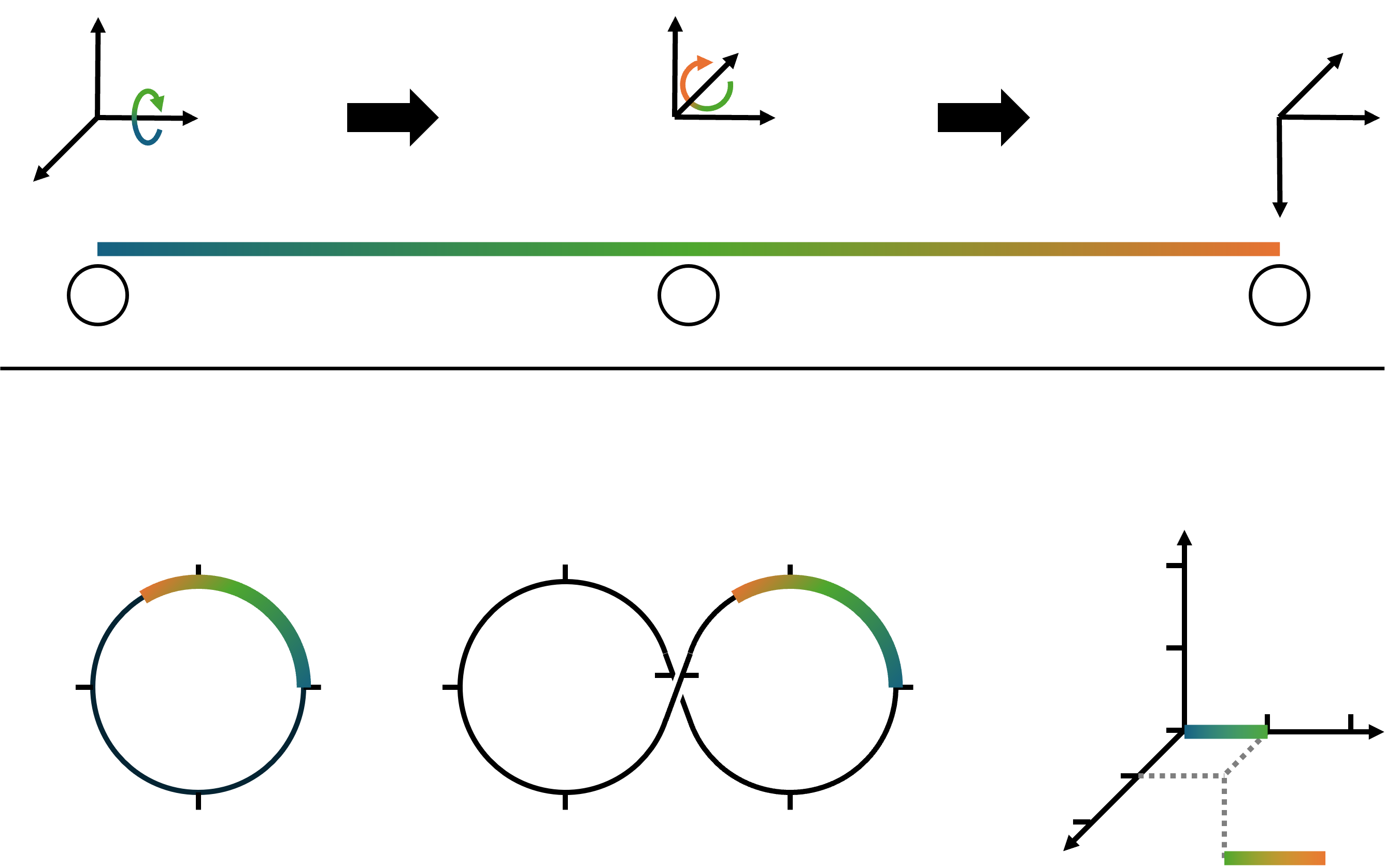}};

\coordinate (r) at (0,0);
\node at ($(r)+(-3.2,-0.1)$) {Rotation matrix};
\node at ($(r)+(-3.1,-0.6)$) {$\nicefrac{\pi}{2}$};
\node at ($(r)+(-3.2,-2.5)$) {$\text{--}\nicefrac{\pi}{2}$};
\node at ($(r)+(-4.1,-1.55)$) {$\pm\pi$};
\node at ($(r)+(-2.25,-1.55)$) {$0$};

\coordinate (q1) at (2.25,0);
\node at ($(q1)+(-2.45,-0.1)$) {Quaternion};
\node at ($(q1)+(-3.12,-0.6)$) {$\nicefrac{\pi}{2}$};
\node at ($(q1)+(-3.2,-2.5)$) {$\text{--}\nicefrac{\pi}{2}$};
\node at ($(q1)+(-4.0,-1.55)$) {$0$};
\node at ($(q1)+(-2.85,-1.47)$) {$\pm\pi$};

\coordinate (q2) at (3.6,0);
\node at ($(q2)+(-3.1,-0.6)$) {$\nicefrac{\pi}{2}$};
\node at ($(q2)+(-3.2,-2.5)$) {$\text{--}\nicefrac{\pi}{2}$};
\node at ($(q2)+(-2.25,-1.55)$) {$0$};
\node at ($(q2)+(-3.4,-1.47)$) {$\pm\pi$};

\coordinate (e) at (0,0);
\node at ($(e)+(3,-0.1)$) {Euler angles};
\node at ($(e)+(2.5,-0.8)$) {$\pm\pi$};
\node at ($(e)+(2.5,-1.3)$) {$\nicefrac{\pi}{2}$};
\node at ($(e)+(2.65,-1.75)$) {$0$};
\node at ($(e)+(2.22,-2.03)$) {$\nicefrac{\pi}{2}$};
\node at ($(e)+(1.9,-2.35)$) {$\pm\pi$};
\node at ($(e)+(3.4,-1.5)$) {$\nicefrac{\pi}{2}$};
\node at ($(e)+(3.95,-1.52)$) {$\pm\pi$};
\node at ($(e)+(3.05,-0.55)$) {$r$};
\node at ($(e)+(2.4,-2.6)$) {$y$};
\node at ($(e)+(4.05,-2.05)$) {$p$};

\coordinate (c) at (0,0);
\node at ($(c)+(-3.72,0.83)$) {1};
\node at ($(c)+(-0.12,0.83)$) {2};
\node at ($(c)+(3.47,0.83)$) {3};

\coordinate (f1) at (0,0);
\node at ($(f1)+(-4.25,1.6)$) {$x$};
\node at ($(f1)+(-3.0,1.85)$) {$y$};
\node at ($(f1)+(-3.55,2.53)$) {$z$};

\coordinate (f2) at (3.5,0);
\node at ($(f2)+(-3.55,2.53)$) {$x$};
\node at ($(f2)+(-3.0,1.85)$) {$y$};
\node at ($(f2)+(-3.2,2.3)$) {$z$};

\coordinate (f3) at (7.2,0);
\node at ($(f3)+(-3.2,1.7)$) {$x$};
\node at ($(f3)+(-3.55,1.35)$) {$y$};
\node at ($(f3)+(-3.2,2.3)$) {$z$};

\end{tikzpicture}

%% file: figures/group_tangent.tex
\begin{tikzpicture}

\node[inner sep=0pt] (img) at (0.0,0.0)
{\includegraphics[width=0.5\columnwidth]{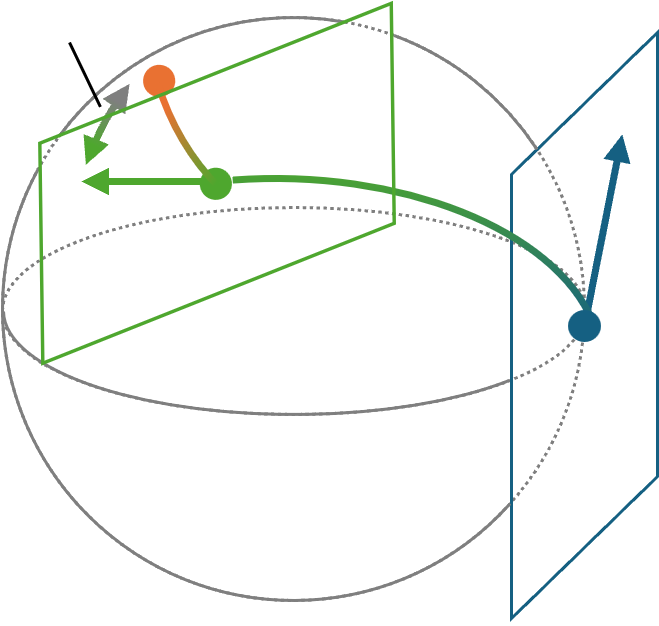}};

\coordinate (r) at (0,0);
\node at ($(r)+(3.0,-0.8)$) {\textcolor{myblue}{$T_{\bs{\mathcal{E}}}\mathcal{M}\cong\mathfrak{m}$}};
\node at ($(r)+(1.94,1.29)$) {\textcolor{myblue}{$\prescript{\bs{\mathcal{E}}}{}{\bs{\tau}}$}};
\node at ($(r)+(-2.45,0.9)$) {\textcolor{mygreen}{$T_{\bs{x}}\mathcal{M}$}};
\node at ($(r)+(-1.57,0.61)$) {\textcolor{mygreen}{$\prescript{\bs{x}}{}{\bs{\tau}}$}};
\node at ($(r)+(-1.9,2.0)$) {$\Exp,\Log$};
\node at ($(r)+(-2.1,-1.2)$) {$\textcolor{mygray}{\mathcal{M}}$};
\node at ($(r)+(1.9,-0.3)$) {$\textcolor{mygray}{\bs{\mathcal{E}}}$};
\node at ($(r)+(-0.55,1.1)$) {$\textcolor{mygray}{\bs{x}}$};
\node at ($(r)+(-0.85,1.68)$) {$\textcolor{mygray}{\bs{y}}$};

\end{tikzpicture}

%% file: figures/orientation_control.tex
\begin{tikzpicture}

\node[inner sep=0pt] (img) at (0.0,0.0)
{\includegraphics[width=1.0\columnwidth]{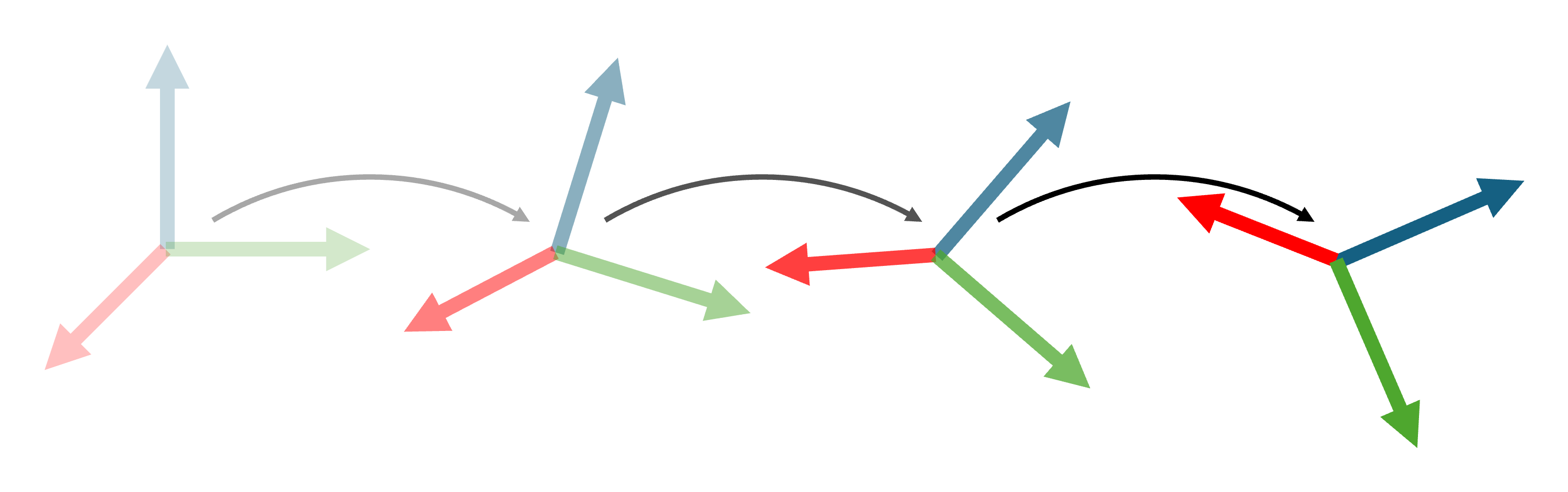}};

\coordinate (r) at (-3.2,-0.3);
\node at ($(r)+(-0.0,-0.0)$) {$\bs{s}_0$};
\node at ($(r)+(2.0,-0.0)$) {$\bs{s}_1$};
\node at ($(r)+(4.0,-0.0)$) {$\bs{s}_2$};
\node at ($(r)+(6.0,-0.0)$) {$\bs{s}_3$};

\node at ($(r)+(0.95,0.85)$) {$\bs{a}_0$};
\node at ($(r)+(3.15,0.85)$) {$\bs{a}_1$};
\node at ($(r)+(5.25,0.85)$) {$\bs{a}_2$};

\end{tikzpicture}

%% file: figures/pickplaceres.tex

\begin{tikzpicture}[/tikz/background rectangle/.style={fill={rgb,1:red,1.0;green,1.0;blue,1.0}, fill opacity={1.0}, draw opacity={1.0}}, show background rectangle]
\begin{axis}[point meta max={nan}, point meta min={nan}, legend cell align={left}, legend columns={1}, title={\textbf{Pick-and-place task}}, title style={yshift={-6}, at={{(0.5,1)}}, anchor={south}, font={{\fontsize{8 pt}{10.4 pt}\selectfont}}, color={rgb,1:red,0.0;green,0.0;blue,0.0}, draw opacity={1.0}, rotate={0.0}, align={center}}, legend columns={-1}, legend style={color={rgb,1:red,0.0;green,0.0;blue,0.0}, draw opacity={1.0}, line width={1}, solid, fill={rgb,1:red,1.0;green,1.0;blue,1.0}, fill opacity={1.0}, text opacity={1.0}, font={{\fontsize{8 pt}{10.4 pt}\selectfont}}, text={rgb,1:red,0.0;green,0.0;blue,0.0}, cells={anchor={center}}, at={(0.98, 0.02)}, anchor={south east}}, axis background/.style={fill={rgb,1:red,1.0;green,1.0;blue,1.0}, opacity={1.0}}, anchor={north west}, xshift={1.0mm}, yshift={-0.0mm}, width={85.55mm}, height={44.1mm}, scaled x ticks={false}, xlabel={simulation steps}, x tick style={color={rgb,1:red,0.0;green,0.0;blue,0.0}, opacity={1.0}}, x tick label style={color={rgb,1:red,0.0;green,0.0;blue,0.0}, opacity={1.0}, rotate={0}}, xlabel style={at={(ticklabel cs:0.5)}, anchor=near ticklabel, at={{(ticklabel cs:0.5)}}, anchor={near ticklabel}, font={{\fontsize{8 pt}{10.4 pt}\selectfont}}, color={rgb,1:red,0.0;green,0.0;blue,0.0}, draw opacity={1.0}, rotate={0.0}}, xmajorgrids={true}, xmin={-293504.0}, xmax={1.0303104e7}, xticklabels={{$0$,$4\times10^{^{6}}$,$8\times10^{^{6}}$}}, xtick={{0.0,4.0e6,8.0e6}}, xtick align={inside}, xticklabel style={font={{\fontsize{8 pt}{10.4 pt}\selectfont}}, color={rgb,1:red,0.0;green,0.0;blue,0.0}, draw opacity={1.0}, rotate={0.0}}, x grid style={color={rgb,1:red,0.0;green,0.0;blue,0.0}, draw opacity={0.1}, line width={0.5}, solid}, axis x line*={left}, x axis line style={color={rgb,1:red,0.0;green,0.0;blue,0.0}, draw opacity={1.0}, line width={1}, solid}, scaled y ticks={false}, ylabel={success rate}, y tick style={color={rgb,1:red,0.0;green,0.0;blue,0.0}, opacity={1.0}}, y tick label style={color={rgb,1:red,0.0;green,0.0;blue,0.0}, opacity={1.0}, rotate={0}}, ylabel style={at={(ticklabel cs:0.5)}, anchor=near ticklabel, at={{(ticklabel cs:0.5)}}, anchor={near ticklabel}, font={{\fontsize{8 pt}{10.4 pt}\selectfont}}, color={rgb,1:red,0.0;green,0.0;blue,0.0}, draw opacity={1.0}, rotate={0.0}}, ymajorgrids={true}, ymin={-0.05}, ymax={1.025}, yticklabels={{$0.0$,$0.5$,$1.0$}}, ytick={{0.0,0.5,1.0}}, ytick align={inside}, yticklabel style={font={{\fontsize{8 pt}{10.4 pt}\selectfont}}, color={rgb,1:red,0.0;green,0.0;blue,0.0}, draw opacity={1.0}, rotate={0.0}}, y grid style={color={rgb,1:red,0.0;green,0.0;blue,0.0}, draw opacity={0.1}, line width={0.5}, solid}, axis y line*={left}, y axis line style={color={rgb,1:red,0.0;green,0.0;blue,0.0}, draw opacity={1.0}, line width={1}, solid}, colorbar={false}]
    \addplot+[line width={0}, draw opacity={0}, fill={rgb,1:red,0.0;green,0.6056;blue,0.9787}, fill opacity={0.1}, mark={none}, forget plot]
        coordinates {
            (6400,0.0)
            (134400,0.0)
            (262400,0.0)
            (390400,0.0)
            (518400,0.0)
            (646400,0.0)
            (774400,0.0)
            (902400,0.0)
            (1030400,0.0)
            (1158400,0.009375)
            (1286400,0.021875)
            (1414400,0.05)
            (1542400,0.053125)
            (1670400,0.065625)
            (1798400,0.13125)
            (1926400,0.11875)
            (2054400,0.203125)
            (2182400,0.178125)
            (2310400,0.21875)
            (2438400,0.234375)
            (2566400,0.20625)
            (2694400,0.2375)
            (2822400,0.25)
            (2950400,0.209375)
            (3078400,0.321875)
            (3206400,0.328125)
            (3334400,0.33125)
            (3462400,0.34375)
            (3590400,0.35625)
            (3718400,0.35)
            (3846400,0.340625)
            (3974400,0.371875)
            (4102400,0.45)
            (4230400,0.38125)
            (4358400,0.34375)
            (4486400,0.45625)
            (4614400,0.45625)
            (4742400,0.41875)
            (4870400,0.48125)
            (4998400,0.496875)
            (5126400,0.453125)
            (5254400,0.4875)
            (5382400,0.45)
            (5510400,0.475)
            (5638400,0.565625)
            (5766400,0.55625)
            (5894400,0.484375)
            (6022400,0.546875)
            (6150400,0.528125)
            (6278400,0.565625)
            (6406400,0.6)
            (6534400,0.58125)
            (6662400,0.54375)
            (6790400,0.603125)
            (6918400,0.584375)
            (7046400,0.525)
            (7174400,0.565625)
            (7302400,0.575)
            (7430400,0.5375)
            (7558400,0.58125)
            (7686400,0.596875)
            (7814400,0.603125)
            (7942400,0.6)
            (8070400,0.578125)
            (8198400,0.609375)
            (8326400,0.58125)
            (8454400,0.665625)
            (8582400,0.625)
            (8710400,0.6375)
            (8838400,0.55)
            (8966400,0.6)
            (9094400,0.640625)
            (9222400,0.659375)
            (9350400,0.615625)
            (9478400,0.675)
            (9606400,0.646875)
            (9734400,0.653125)
            (9862400,0.625)
            (9990400,0.684375)
            (10003200,0.68125)
            (10003200,0.6221635299962843)
            (9990400,0.6186384876001167)
            (9862400,0.5525498555039399)
            (9734400,0.583773934308837)
            (9606400,0.5306812972338861)
            (9478400,0.5884256551425309)
            (9350400,0.555514424807277)
            (9222400,0.6094727518041522)
            (9094400,0.5811267726104382)
            (8966400,0.5388824774512252)
            (8838400,0.4387438555067632)
            (8710400,0.5941423846077761)
            (8582400,0.5665366033316571)
            (8454400,0.6206098115909753)
            (8326400,0.5289587483416203)
            (8198400,0.5421693338512295)
            (8070400,0.48904879491412984)
            (7942400,0.5440983005625052)
            (7814400,0.5420074774512252)
            (7686400,0.5390413889896196)
            (7558400,0.47795941324351)
            (7430400,0.45661138059034506)
            (7302400,0.4884256551425308)
            (7174400,0.47559462318195045)
            (7046400,0.42100274192316417)
            (6918400,0.510589256543015)
            (6790400,0.5747407779303361)
            (6662400,0.46930544219017745)
            (6534400,0.46890201242345286)
            (6406400,0.5563617498632221)
            (6278400,0.45827772476676465)
            (6150400,0.4360832720311053)
            (6022400,0.49388804700683386)
            (5894400,0.40394038952353906)
            (5766400,0.4448841894924659)
            (5638400,0.4611592834347077)
            (5510400,0.401214256543015)
            (5382400,0.3965542769484779)
            (5254400,0.45186951796565195)
            (5126400,0.3346425711866945)
            (4998400,0.4521318091772614)
            (4870400,0.38920827203110536)
            (4742400,0.324092929807119)
            (4614400,0.3860243592054868)
            (4486400,0.4261947124036385)
            (4358400,0.27654433385122945)
            (4230400,0.34050498496748344)
            (4102400,0.37805709729236664)
            (3974400,0.3314306902951726)
            (3846400,0.2583893914566202)
            (3718400,0.24253906959969124)
            (3590400,0.31894582379277087)
            (3462400,0.26641019580772135)
            (3334400,0.24538355209978696)
            (3206400,0.2739984122634726)
            (3078400,0.24808925654301506)
            (2950400,0.13726262025491048)
            (2822400,0.16370811705322452)
            (2694400,0.11526495490245033)
            (2566400,0.11758586900555557)
            (2438400,0.11692737204396166)
            (2310400,0.13388452741249832)
            (2182400,0.07066406959969126)
            (2054400,0.1368337392637612)
            (1926400,0.05966352999628425)
            (1798400,0.0823360129921921)
            (1670400,0.02518069029517255)
            (1542400,0.012379984967483444)
            (1414400,0.006642384607776226)
            (1286400,0.007899575140626313)
            (1158400,0.000816835038981778)
            (1030400,0.0)
            (902400,0.0)
            (774400,0.0)
            (646400,0.0)
            (518400,0.0)
            (390400,0.0)
            (262400,0.0)
            (134400,0.0)
            (6400,0.0)
            (6400,0.0)
        }
        ;
    \addplot+[line width={0}, draw opacity={0}, fill={rgb,1:red,0.0;green,0.6056;blue,0.9787}, fill opacity={0.1}, mark={none}, forget plot]
        coordinates {
            (6400,0.0)
            (134400,0.0)
            (262400,0.0)
            (390400,0.0)
            (518400,0.0)
            (646400,0.0)
            (774400,0.0)
            (902400,0.0)
            (1030400,0.0)
            (1158400,0.009375)
            (1286400,0.021875)
            (1414400,0.05)
            (1542400,0.053125)
            (1670400,0.065625)
            (1798400,0.13125)
            (1926400,0.11875)
            (2054400,0.203125)
            (2182400,0.178125)
            (2310400,0.21875)
            (2438400,0.234375)
            (2566400,0.20625)
            (2694400,0.2375)
            (2822400,0.25)
            (2950400,0.209375)
            (3078400,0.321875)
            (3206400,0.328125)
            (3334400,0.33125)
            (3462400,0.34375)
            (3590400,0.35625)
            (3718400,0.35)
            (3846400,0.340625)
            (3974400,0.371875)
            (4102400,0.45)
            (4230400,0.38125)
            (4358400,0.34375)
            (4486400,0.45625)
            (4614400,0.45625)
            (4742400,0.41875)
            (4870400,0.48125)
            (4998400,0.496875)
            (5126400,0.453125)
            (5254400,0.4875)
            (5382400,0.45)
            (5510400,0.475)
            (5638400,0.565625)
            (5766400,0.55625)
            (5894400,0.484375)
            (6022400,0.546875)
            (6150400,0.528125)
            (6278400,0.565625)
            (6406400,0.6)
            (6534400,0.58125)
            (6662400,0.54375)
            (6790400,0.603125)
            (6918400,0.584375)
            (7046400,0.525)
            (7174400,0.565625)
            (7302400,0.575)
            (7430400,0.5375)
            (7558400,0.58125)
            (7686400,0.596875)
            (7814400,0.603125)
            (7942400,0.6)
            (8070400,0.578125)
            (8198400,0.609375)
            (8326400,0.58125)
            (8454400,0.665625)
            (8582400,0.625)
            (8710400,0.6375)
            (8838400,0.55)
            (8966400,0.6)
            (9094400,0.640625)
            (9222400,0.659375)
            (9350400,0.615625)
            (9478400,0.675)
            (9606400,0.646875)
            (9734400,0.653125)
            (9862400,0.625)
            (9990400,0.684375)
            (10003200,0.68125)
            (10003200,0.7403364700037157)
            (9990400,0.7501115123998832)
            (9862400,0.6974501444960601)
            (9734400,0.722476065691163)
            (9606400,0.7630687027661138)
            (9478400,0.7615743448574692)
            (9350400,0.675735575192723)
            (9222400,0.7092772481958479)
            (9094400,0.7001232273895618)
            (8966400,0.6611175225487748)
            (8838400,0.6612561444932369)
            (8710400,0.6808576153922238)
            (8582400,0.6834633966683429)
            (8454400,0.7106401884090248)
            (8326400,0.6335412516583798)
            (8198400,0.6765806661487705)
            (8070400,0.6672012050858701)
            (7942400,0.6559016994374948)
            (7814400,0.6642425225487748)
            (7686400,0.6547086110103805)
            (7558400,0.6845405867564901)
            (7430400,0.6183886194096548)
            (7302400,0.6615743448574691)
            (7174400,0.6556553768180496)
            (7046400,0.6289972580768359)
            (6918400,0.6581607434569849)
            (6790400,0.631509222069664)
            (6662400,0.6181945578098225)
            (6534400,0.6935979875765472)
            (6406400,0.6436382501367779)
            (6278400,0.6729722752332354)
            (6150400,0.6201667279688946)
            (6022400,0.5998619529931661)
            (5894400,0.5648096104764609)
            (5766400,0.6676158105075342)
            (5638400,0.6700907165652924)
            (5510400,0.5487857434569849)
            (5382400,0.5034457230515221)
            (5254400,0.5231304820343481)
            (5126400,0.5716074288133055)
            (4998400,0.5416181908227387)
            (4870400,0.5732917279688946)
            (4742400,0.513407070192881)
            (4614400,0.5264756407945133)
            (4486400,0.4863052875963615)
            (4358400,0.41095566614877055)
            (4230400,0.4219950150325165)
            (4102400,0.5219429027076333)
            (3974400,0.41231930970482744)
            (3846400,0.4228606085433798)
            (3718400,0.4574609304003087)
            (3590400,0.39355417620722916)
            (3462400,0.42108980419227865)
            (3334400,0.417116447900213)
            (3206400,0.3822515877365274)
            (3078400,0.395660743456985)
            (2950400,0.2814873797450895)
            (2822400,0.3362918829467755)
            (2694400,0.35973504509754967)
            (2566400,0.2949141309944444)
            (2438400,0.35182262795603836)
            (2310400,0.30361547258750166)
            (2182400,0.2855859304003088)
            (2054400,0.2694162607362388)
            (1926400,0.17783647000371575)
            (1798400,0.1801639870078079)
            (1670400,0.10606930970482745)
            (1542400,0.09387001503251655)
            (1414400,0.09335761539222379)
            (1286400,0.03585042485937368)
            (1158400,0.01793316496101822)
            (1030400,0.0)
            (902400,0.0)
            (774400,0.0)
            (646400,0.0)
            (518400,0.0)
            (390400,0.0)
            (262400,0.0)
            (134400,0.0)
            (6400,0.0)
            (6400,0.0)
        }
        ;
    \addplot[color={rgb,1:red,0.0;green,0.6056;blue,0.9787}, name path={d9d5214c-aca2-4ab6-bbbd-828deecdebf8}, legend image code/.code={{
    \draw[fill={rgb,1:red,0.0;green,0.6056;blue,0.9787}, fill opacity={0.1}] (0cm,0cm) rectangle (0.3cm,0cm);
    }}, draw opacity={1.0}, line width={2}, solid]
        table[row sep={\\}]
        {
            \\
            6400.0  0.0  \\
            134400.0  0.0  \\
            262400.0  0.0  \\
            390400.0  0.0  \\
            518400.0  0.0  \\
            646400.0  0.0  \\
            774400.0  0.0  \\
            902400.0  0.0  \\
            1.0304e6  0.0  \\
            1.1584e6  0.009375  \\
            1.2864e6  0.021875  \\
            1.4144e6  0.05  \\
            1.5424e6  0.053125  \\
            1.6704e6  0.065625  \\
            1.7984e6  0.13125  \\
            1.9264e6  0.11875  \\
            2.0544e6  0.203125  \\
            2.1824e6  0.178125  \\
            2.3104e6  0.21875  \\
            2.4384e6  0.234375  \\
            2.5664e6  0.20625  \\
            2.6944e6  0.2375  \\
            2.8224e6  0.25  \\
            2.9504e6  0.209375  \\
            3.0784e6  0.321875  \\
            3.2064e6  0.328125  \\
            3.3344e6  0.33125  \\
            3.4624e6  0.34375  \\
            3.5904e6  0.35625  \\
            3.7184e6  0.35  \\
            3.8464e6  0.340625  \\
            3.9744e6  0.371875  \\
            4.1024e6  0.45  \\
            4.2304e6  0.38125  \\
            4.3584e6  0.34375  \\
            4.4864e6  0.45625  \\
            4.6144e6  0.45625  \\
            4.7424e6  0.41875  \\
            4.8704e6  0.48125  \\
            4.9984e6  0.496875  \\
            5.1264e6  0.453125  \\
            5.2544e6  0.4875  \\
            5.3824e6  0.45  \\
            5.5104e6  0.475  \\
            5.6384e6  0.565625  \\
            5.7664e6  0.55625  \\
            5.8944e6  0.484375  \\
            6.0224e6  0.546875  \\
            6.1504e6  0.528125  \\
            6.2784e6  0.565625  \\
            6.4064e6  0.6  \\
            6.5344e6  0.58125  \\
            6.6624e6  0.54375  \\
            6.7904e6  0.603125  \\
            6.9184e6  0.584375  \\
            7.0464e6  0.525  \\
            7.1744e6  0.565625  \\
            7.3024e6  0.575  \\
            7.4304e6  0.5375  \\
            7.5584e6  0.58125  \\
            7.6864e6  0.596875  \\
            7.8144e6  0.603125  \\
            7.9424e6  0.6  \\
            8.0704e6  0.578125  \\
            8.1984e6  0.609375  \\
            8.3264e6  0.58125  \\
            8.4544e6  0.665625  \\
            8.5824e6  0.625  \\
            8.7104e6  0.6375  \\
            8.8384e6  0.55  \\
            8.9664e6  0.6  \\
            9.0944e6  0.640625  \\
            9.2224e6  0.659375  \\
            9.3504e6  0.615625  \\
            9.4784e6  0.675  \\
            9.6064e6  0.646875  \\
            9.7344e6  0.653125  \\
            9.8624e6  0.625  \\
            9.9904e6  0.684375  \\
            1.00032e7  0.68125  \\
        }
        ;
    \addlegendentry {$\mathfrak{m}$}
    \addplot+[line width={0}, draw opacity={0}, fill={rgb,1:red,0.8889;green,0.4356;blue,0.2781}, fill opacity={0.1}, mark={none}, forget plot]
        coordinates {
            (6400,0.0)
            (134400,0.0)
            (262400,0.0)
            (390400,0.0)
            (518400,0.0)
            (646400,0.0)
            (774400,0.0)
            (902400,0.0)
            (1030400,0.0)
            (1158400,0.0)
            (1286400,0.0)
            (1414400,0.0)
            (1542400,0.0)
            (1670400,0.0)
            (1798400,0.0)
            (1926400,0.00625)
            (2054400,0.009375)
            (2182400,0.021875)
            (2310400,0.021875)
            (2438400,0.046875)
            (2566400,0.059375)
            (2694400,0.075)
            (2822400,0.059375)
            (2950400,0.071875)
            (3078400,0.08125)
            (3206400,0.128125)
            (3334400,0.1375)
            (3462400,0.1625)
            (3590400,0.153125)
            (3718400,0.13125)
            (3846400,0.15625)
            (3974400,0.209375)
            (4102400,0.240625)
            (4230400,0.20625)
            (4358400,0.1875)
            (4486400,0.209375)
            (4614400,0.23125)
            (4742400,0.253125)
            (4870400,0.21875)
            (4998400,0.259375)
            (5126400,0.25625)
            (5254400,0.284375)
            (5382400,0.3375)
            (5510400,0.309375)
            (5638400,0.30625)
            (5766400,0.315625)
            (5894400,0.359375)
            (6022400,0.346875)
            (6150400,0.315625)
            (6278400,0.353125)
            (6406400,0.36875)
            (6534400,0.41875)
            (6662400,0.403125)
            (6790400,0.378125)
            (6918400,0.421875)
            (7046400,0.440625)
            (7174400,0.415625)
            (7302400,0.409375)
            (7430400,0.4)
            (7558400,0.425)
            (7686400,0.39375)
            (7814400,0.396875)
            (7942400,0.475)
            (8070400,0.396875)
            (8198400,0.409375)
            (8326400,0.415625)
            (8454400,0.415625)
            (8582400,0.453125)
            (8710400,0.371875)
            (8838400,0.415625)
            (8966400,0.45)
            (9094400,0.371875)
            (9222400,0.409375)
            (9350400,0.3375)
            (9478400,0.325)
            (9606400,0.40625)
            (9734400,0.421875)
            (9862400,0.34375)
            (9990400,0.35625)
            (10003200,0.325)
            (10003200,0.15912459966137235)
            (9990400,0.19259723784029797)
            (9862400,0.21070791901619998)
            (9734400,0.27866950953263003)
            (9606400,0.22707273529406696)
            (9478400,0.19643748480797368)
            (9350400,0.20510174897303216)
            (9222400,0.23629682492440013)
            (9094400,0.295808365971012)
            (8966400,0.32335072419275346)
            (8838400,0.32083905694935566)
            (8710400,0.22722766486208604)
            (8582400,0.35680603120360976)
            (8454400,0.3545074774512252)
            (8326400,0.3435126202549105)
            (8198400,0.3309381225124304)
            (8070400,0.3443508240188387)
            (7942400,0.444541267672472)
            (7814400,0.31086650659673193)
            (7686400,0.37663367007796356)
            (7558400,0.3284278928727347)
            (7430400,0.3430169295974673)
            (7302400,0.3418070365298761)
            (7174400,0.34183925654301506)
            (7046400,0.3695355427120167)
            (6918400,0.3384603261560054)
            (6790400,0.3202913889896195)
            (6662400,0.30911494402192924)
            (6534400,0.36306709474623294)
            (6406400,0.2933280100865802)
            (6278400,0.27148530951491795)
            (6150400,0.27832082379277084)
            (6022400,0.2994820284746778)
            (5894400,0.27894038952353906)
            (5766400,0.24805703652987607)
            (5638400,0.24126051767785808)
            (5510400,0.2689306902951726)
            (5382400,0.24271405694935563)
            (5254400,0.22760655586067202)
            (5126400,0.19126051767785804)
            (4998400,0.19073161676097253)
            (4870400,0.15154433385122948)
            (4742400,0.18118209729236662)
            (4614400,0.13280009998735398)
            (4486400,0.14253359488685174)
            (4358400,0.09639137664304218)
            (4230400,0.16289238460777622)
            (4102400,0.18472330056250527)
            (3974400,0.12265976921555245)
            (3846400,0.07138452741249832)
            (3718400,0.07534830056250527)
            (3590400,0.06241921703937506)
            (3462400,0.0895461404379453)
            (3334400,0.09557372542187895)
            (3206400,0.08476738460777622)
            (3078400,0.03516287123827305)
            (2950400,0.054059758982825966)
            (2822400,0.02046906375628521)
            (2694400,0.03164238460777623)
            (2566400,0.0026065558606720232)
            (2438400,0.022169705779934535)
            (2310400,0.013316835038981779)
            (2182400,-0.004270625829189863)
            (2054400,0.000816835038981778)
            (1926400,-0.0023081649610182213)
            (1798400,0.0)
            (1670400,0.0)
            (1542400,0.0)
            (1414400,0.0)
            (1286400,0.0)
            (1158400,0.0)
            (1030400,0.0)
            (902400,0.0)
            (774400,0.0)
            (646400,0.0)
            (518400,0.0)
            (390400,0.0)
            (262400,0.0)
            (134400,0.0)
            (6400,0.0)
            (6400,0.0)
        }
        ;
    \addplot+[line width={0}, draw opacity={0}, fill={rgb,1:red,0.8889;green,0.4356;blue,0.2781}, fill opacity={0.1}, mark={none}, forget plot]
        coordinates {
            (6400,0.0)
            (134400,0.0)
            (262400,0.0)
            (390400,0.0)
            (518400,0.0)
            (646400,0.0)
            (774400,0.0)
            (902400,0.0)
            (1030400,0.0)
            (1158400,0.0)
            (1286400,0.0)
            (1414400,0.0)
            (1542400,0.0)
            (1670400,0.0)
            (1798400,0.0)
            (1926400,0.00625)
            (2054400,0.009375)
            (2182400,0.021875)
            (2310400,0.021875)
            (2438400,0.046875)
            (2566400,0.059375)
            (2694400,0.075)
            (2822400,0.059375)
            (2950400,0.071875)
            (3078400,0.08125)
            (3206400,0.128125)
            (3334400,0.1375)
            (3462400,0.1625)
            (3590400,0.153125)
            (3718400,0.13125)
            (3846400,0.15625)
            (3974400,0.209375)
            (4102400,0.240625)
            (4230400,0.20625)
            (4358400,0.1875)
            (4486400,0.209375)
            (4614400,0.23125)
            (4742400,0.253125)
            (4870400,0.21875)
            (4998400,0.259375)
            (5126400,0.25625)
            (5254400,0.284375)
            (5382400,0.3375)
            (5510400,0.309375)
            (5638400,0.30625)
            (5766400,0.315625)
            (5894400,0.359375)
            (6022400,0.346875)
            (6150400,0.315625)
            (6278400,0.353125)
            (6406400,0.36875)
            (6534400,0.41875)
            (6662400,0.403125)
            (6790400,0.378125)
            (6918400,0.421875)
            (7046400,0.440625)
            (7174400,0.415625)
            (7302400,0.409375)
            (7430400,0.4)
            (7558400,0.425)
            (7686400,0.39375)
            (7814400,0.396875)
            (7942400,0.475)
            (8070400,0.396875)
            (8198400,0.409375)
            (8326400,0.415625)
            (8454400,0.415625)
            (8582400,0.453125)
            (8710400,0.371875)
            (8838400,0.415625)
            (8966400,0.45)
            (9094400,0.371875)
            (9222400,0.409375)
            (9350400,0.3375)
            (9478400,0.325)
            (9606400,0.40625)
            (9734400,0.421875)
            (9862400,0.34375)
            (9990400,0.35625)
            (10003200,0.325)
            (10003200,0.49087540033862764)
            (9990400,0.5199027621597021)
            (9862400,0.4767920809838)
            (9734400,0.56508049046737)
            (9606400,0.5854272647059331)
            (9478400,0.45356251519202634)
            (9350400,0.4698982510269679)
            (9222400,0.5824531750755999)
            (9094400,0.447941634028988)
            (8966400,0.5766492758072466)
            (8838400,0.5104109430506444)
            (8710400,0.516522335137914)
            (8582400,0.5494439687963902)
            (8454400,0.47674252254877486)
            (8326400,0.48773737974508957)
            (8198400,0.4878118774875696)
            (8070400,0.44939917598116125)
            (7942400,0.505458732327528)
            (7814400,0.482883493403268)
            (7686400,0.4108663299220364)
            (7558400,0.5215721071272652)
            (7430400,0.4569830704025327)
            (7302400,0.4769429634701239)
            (7174400,0.489410743456985)
            (7046400,0.5117144572879833)
            (6918400,0.5052896738439946)
            (6790400,0.43595861101038047)
            (6662400,0.4971350559780708)
            (6534400,0.4744329052537671)
            (6406400,0.44417198991341983)
            (6278400,0.4347646904850821)
            (6150400,0.35292917620722913)
            (6022400,0.39426797152532217)
            (5894400,0.43980961047646094)
            (5766400,0.3831929634701239)
            (5638400,0.371239482322142)
            (5510400,0.34981930970482744)
            (5382400,0.4322859430506444)
            (5254400,0.34114344413932796)
            (5126400,0.32123948232214194)
            (4998400,0.32801838323902754)
            (4870400,0.28595566614877055)
            (4742400,0.32506790270763336)
            (4614400,0.32969990001264604)
            (4486400,0.27621640511314827)
            (4358400,0.2786086233569578)
            (4230400,0.24960761539222376)
            (4102400,0.29652669943749477)
            (3974400,0.2960902307844476)
            (3846400,0.24111547258750168)
            (3718400,0.18715169943749474)
            (3590400,0.24383078296062496)
            (3462400,0.23545385956205472)
            (3334400,0.17942627457812108)
            (3206400,0.17148261539222376)
            (3078400,0.12733712876172695)
            (2950400,0.08969024101717402)
            (2822400,0.09828093624371478)
            (2694400,0.11835761539222377)
            (2566400,0.11614344413932798)
            (2438400,0.07158029422006547)
            (2310400,0.03043316496101822)
            (2182400,0.04802062582918986)
            (2054400,0.01793316496101822)
            (1926400,0.014808164961018222)
            (1798400,0.0)
            (1670400,0.0)
            (1542400,0.0)
            (1414400,0.0)
            (1286400,0.0)
            (1158400,0.0)
            (1030400,0.0)
            (902400,0.0)
            (774400,0.0)
            (646400,0.0)
            (518400,0.0)
            (390400,0.0)
            (262400,0.0)
            (134400,0.0)
            (6400,0.0)
            (6400,0.0)
        }
        ;
    \addplot[color={rgb,1:red,0.8889;green,0.4356;blue,0.2781}, name path={43a695df-9a3e-4afd-9776-7b7770d7f709}, legend image code/.code={{
    \draw[fill={rgb,1:red,0.8889;green,0.4356;blue,0.2781}, fill opacity={0.1}] (0cm,0cm) rectangle (0.3cm,0cm);
    }}, draw opacity={1.0}, line width={2}, solid]
        table[row sep={\\}]
        {
            \\
            6400.0  0.0  \\
            134400.0  0.0  \\
            262400.0  0.0  \\
            390400.0  0.0  \\
            518400.0  0.0  \\
            646400.0  0.0  \\
            774400.0  0.0  \\
            902400.0  0.0  \\
            1.0304e6  0.0  \\
            1.1584e6  0.0  \\
            1.2864e6  0.0  \\
            1.4144e6  0.0  \\
            1.5424e6  0.0  \\
            1.6704e6  0.0  \\
            1.7984e6  0.0  \\
            1.9264e6  0.00625  \\
            2.0544e6  0.009375  \\
            2.1824e6  0.021875  \\
            2.3104e6  0.021875  \\
            2.4384e6  0.046875  \\
            2.5664e6  0.059375  \\
            2.6944e6  0.075  \\
            2.8224e6  0.059375  \\
            2.9504e6  0.071875  \\
            3.0784e6  0.08125  \\
            3.2064e6  0.128125  \\
            3.3344e6  0.1375  \\
            3.4624e6  0.1625  \\
            3.5904e6  0.153125  \\
            3.7184e6  0.13125  \\
            3.8464e6  0.15625  \\
            3.9744e6  0.209375  \\
            4.1024e6  0.240625  \\
            4.2304e6  0.20625  \\
            4.3584e6  0.1875  \\
            4.4864e6  0.209375  \\
            4.6144e6  0.23125  \\
            4.7424e6  0.253125  \\
            4.8704e6  0.21875  \\
            4.9984e6  0.259375  \\
            5.1264e6  0.25625  \\
            5.2544e6  0.284375  \\
            5.3824e6  0.3375  \\
            5.5104e6  0.309375  \\
            5.6384e6  0.30625  \\
            5.7664e6  0.315625  \\
            5.8944e6  0.359375  \\
            6.0224e6  0.346875  \\
            6.1504e6  0.315625  \\
            6.2784e6  0.353125  \\
            6.4064e6  0.36875  \\
            6.5344e6  0.41875  \\
            6.6624e6  0.403125  \\
            6.7904e6  0.378125  \\
            6.9184e6  0.421875  \\
            7.0464e6  0.440625  \\
            7.1744e6  0.415625  \\
            7.3024e6  0.409375  \\
            7.4304e6  0.4  \\
            7.5584e6  0.425  \\
            7.6864e6  0.39375  \\
            7.8144e6  0.396875  \\
            7.9424e6  0.475  \\
            8.0704e6  0.396875  \\
            8.1984e6  0.409375  \\
            8.3264e6  0.415625  \\
            8.4544e6  0.415625  \\
            8.5824e6  0.453125  \\
            8.7104e6  0.371875  \\
            8.8384e6  0.415625  \\
            8.9664e6  0.45  \\
            9.0944e6  0.371875  \\
            9.2224e6  0.409375  \\
            9.3504e6  0.3375  \\
            9.4784e6  0.325  \\
            9.6064e6  0.40625  \\
            9.7344e6  0.421875  \\
            9.8624e6  0.34375  \\
            9.9904e6  0.35625  \\
            1.00032e7  0.325  \\
        }
        ;
    \addlegendentry {$SO(3)$}
    \addplot+[line width={0}, draw opacity={0}, fill={rgb,1:red,0.2422;green,0.6433;blue,0.3044}, fill opacity={0.1}, mark={none}, forget plot]
        coordinates {
            (6400,0.0)
            (134400,0.0)
            (262400,0.0)
            (390400,0.0)
            (518400,0.0)
            (646400,0.0)
            (774400,0.0)
            (902400,0.0)
            (1030400,0.0)
            (1158400,0.0)
            (1286400,0.0)
            (1414400,0.009375)
            (1542400,0.009375)
            (1670400,0.0125)
            (1798400,0.01875)
            (1926400,0.021875)
            (2054400,0.028125)
            (2182400,0.053125)
            (2310400,0.0625)
            (2438400,0.05)
            (2566400,0.075)
            (2694400,0.134375)
            (2822400,0.140625)
            (2950400,0.15)
            (3078400,0.16875)
            (3206400,0.140625)
            (3334400,0.178125)
            (3462400,0.190625)
            (3590400,0.2375)
            (3718400,0.221875)
            (3846400,0.228125)
            (3974400,0.184375)
            (4102400,0.259375)
            (4230400,0.2625)
            (4358400,0.303125)
            (4486400,0.284375)
            (4614400,0.334375)
            (4742400,0.30625)
            (4870400,0.290625)
            (4998400,0.359375)
            (5126400,0.371875)
            (5254400,0.371875)
            (5382400,0.40625)
            (5510400,0.3375)
            (5638400,0.425)
            (5766400,0.403125)
            (5894400,0.40625)
            (6022400,0.403125)
            (6150400,0.35)
            (6278400,0.41875)
            (6406400,0.371875)
            (6534400,0.43125)
            (6662400,0.44375)
            (6790400,0.43125)
            (6918400,0.440625)
            (7046400,0.4875)
            (7174400,0.46875)
            (7302400,0.465625)
            (7430400,0.43125)
            (7558400,0.475)
            (7686400,0.4875)
            (7814400,0.5625)
            (7942400,0.4625)
            (8070400,0.490625)
            (8198400,0.509375)
            (8326400,0.475)
            (8454400,0.48125)
            (8582400,0.515625)
            (8710400,0.5375)
            (8838400,0.43125)
            (8966400,0.5125)
            (9094400,0.515625)
            (9222400,0.50625)
            (9350400,0.51875)
            (9478400,0.559375)
            (9606400,0.434375)
            (9734400,0.528125)
            (9862400,0.4875)
            (9990400,0.44375)
            (10003200,0.49375)
            (10003200,0.413619491219012)
            (9990400,0.3778280527820969)
            (9862400,0.41387988131631404)
            (9734400,0.4415506551425308)
            (9606400,0.4064241502812526)
            (9478400,0.5004954884955726)
            (9350400,0.45987048849557266)
            (9222400,0.408922428110016)
            (9094400,0.3776287364636274)
            (8966400,0.4476986207708508)
            (8838400,0.3701324774512252)
            (8710400,0.4645461404379453)
            (8582400,0.4662144115598691)
            (8454400,0.38035072318148166)
            (8326400,0.388991506596732)
            (8198400,0.4407316167609725)
            (8070400,0.4128131275125704)
            (7942400,0.4202835747309178)
            (7814400,0.4755036818307809)
            (7686400,0.4082888009022714)
            (7558400,0.38216593264054405)
            (7430400,0.3558280100865802)
            (7302400,0.37358327203110536)
            (7174400,0.3954622537527589)
            (7046400,0.35893748480797366)
            (6918400,0.3583893914566202)
            (6790400,0.31505629723388623)
            (6662400,0.30952042753178416)
            (6534400,0.26383630955026405)
            (6406400,0.24331248480797368)
            (6278400,0.302136824018467)
            (6150400,0.27377305676468455)
            (6022400,0.292971520424909)
            (5894400,0.28932320666331435)
            (5766400,0.29464653151431386)
            (5638400,0.2904071533290123)
            (5510400,0.2596881275125704)
            (5382400,0.2709335306586814)
            (5254400,0.2531866936572941)
            (5126400,0.25267355548274595)
            (4998400,0.24402481984192656)
            (4870400,0.12031966758201612)
            (4742400,0.19005629723388623)
            (4614400,0.20666989536435904)
            (4486400,0.1672395949125543)
            (4358400,0.14089565809478236)
            (4230400,0.09190819920054774)
            (4102400,0.10604286968316134)
            (3974400,0.052068980437396606)
            (3846400,0.08213364216502403)
            (3718400,0.09914164445840323)
            (3590400,0.08665602017316038)
            (3462400,0.07245205999256851)
            (3334400,0.057802354323884655)
            (3206400,0.04884890527757242)
            (3078400,0.05694660112501053)
            (2950400,0.03863418949246586)
            (2822400,0.0508662086478433)
            (2694400,0.042200742476546094)
            (2566400,0.020424221260709433)
            (2438400,0.009555690295172545)
            (2310400,0.016945688321521088)
            (2182400,0.009486749863222052)
            (2054400,-0.005749192794810631)
            (1926400,-0.0018214857626610986)
            (1798400,0.005677187085405069)
            (1670400,-0.00787250751625828)
            (1542400,-0.004600424859373686)
            (1414400,0.000816835038981778)
            (1286400,0.0)
            (1158400,0.0)
            (1030400,0.0)
            (902400,0.0)
            (774400,0.0)
            (646400,0.0)
            (518400,0.0)
            (390400,0.0)
            (262400,0.0)
            (134400,0.0)
            (6400,0.0)
            (6400,0.0)
        }
        ;
    \addplot+[line width={0}, draw opacity={0}, fill={rgb,1:red,0.2422;green,0.6433;blue,0.3044}, fill opacity={0.1}, mark={none}, forget plot]
        coordinates {
            (6400,0.0)
            (134400,0.0)
            (262400,0.0)
            (390400,0.0)
            (518400,0.0)
            (646400,0.0)
            (774400,0.0)
            (902400,0.0)
            (1030400,0.0)
            (1158400,0.0)
            (1286400,0.0)
            (1414400,0.009375)
            (1542400,0.009375)
            (1670400,0.0125)
            (1798400,0.01875)
            (1926400,0.021875)
            (2054400,0.028125)
            (2182400,0.053125)
            (2310400,0.0625)
            (2438400,0.05)
            (2566400,0.075)
            (2694400,0.134375)
            (2822400,0.140625)
            (2950400,0.15)
            (3078400,0.16875)
            (3206400,0.140625)
            (3334400,0.178125)
            (3462400,0.190625)
            (3590400,0.2375)
            (3718400,0.221875)
            (3846400,0.228125)
            (3974400,0.184375)
            (4102400,0.259375)
            (4230400,0.2625)
            (4358400,0.303125)
            (4486400,0.284375)
            (4614400,0.334375)
            (4742400,0.30625)
            (4870400,0.290625)
            (4998400,0.359375)
            (5126400,0.371875)
            (5254400,0.371875)
            (5382400,0.40625)
            (5510400,0.3375)
            (5638400,0.425)
            (5766400,0.403125)
            (5894400,0.40625)
            (6022400,0.403125)
            (6150400,0.35)
            (6278400,0.41875)
            (6406400,0.371875)
            (6534400,0.43125)
            (6662400,0.44375)
            (6790400,0.43125)
            (6918400,0.440625)
            (7046400,0.4875)
            (7174400,0.46875)
            (7302400,0.465625)
            (7430400,0.43125)
            (7558400,0.475)
            (7686400,0.4875)
            (7814400,0.5625)
            (7942400,0.4625)
            (8070400,0.490625)
            (8198400,0.509375)
            (8326400,0.475)
            (8454400,0.48125)
            (8582400,0.515625)
            (8710400,0.5375)
            (8838400,0.43125)
            (8966400,0.5125)
            (9094400,0.515625)
            (9222400,0.50625)
            (9350400,0.51875)
            (9478400,0.559375)
            (9606400,0.434375)
            (9734400,0.528125)
            (9862400,0.4875)
            (9990400,0.44375)
            (10003200,0.49375)
            (10003200,0.5738805087809881)
            (9990400,0.5096719472179031)
            (9862400,0.5611201186836859)
            (9734400,0.6146993448574691)
            (9606400,0.4623258497187474)
            (9478400,0.6182545115044273)
            (9350400,0.5776295115044274)
            (9222400,0.603577571889984)
            (9094400,0.6536212635363726)
            (8966400,0.577301379229149)
            (8838400,0.49236752254877486)
            (8710400,0.6104538595620547)
            (8582400,0.5650355884401309)
            (8454400,0.5821492768185184)
            (8326400,0.561008493403268)
            (8198400,0.5780183832390275)
            (8070400,0.5684368724874296)
            (7942400,0.5047164252690822)
            (7814400,0.649496318169219)
            (7686400,0.5667111990977286)
            (7558400,0.5678340673594559)
            (7430400,0.5066719899134198)
            (7302400,0.5576667279688946)
            (7174400,0.5420377462472411)
            (7046400,0.6160625151920263)
            (6918400,0.5228606085433798)
            (6790400,0.5474437027661139)
            (6662400,0.5779795724682157)
            (6534400,0.598663690449736)
            (6406400,0.5004375151920264)
            (6278400,0.535363175981533)
            (6150400,0.4262269432353154)
            (6022400,0.5132784795750911)
            (5894400,0.5231767933366857)
            (5766400,0.5116034684856862)
            (5638400,0.5595928466709876)
            (5510400,0.4153118724874296)
            (5382400,0.5415664693413186)
            (5254400,0.4905633063427059)
            (5126400,0.4910764445172541)
            (4998400,0.4747251801580734)
            (4870400,0.4609303324179839)
            (4742400,0.4224437027661138)
            (4614400,0.4620801046356409)
            (4486400,0.4015104050874457)
            (4358400,0.46535434190521763)
            (4230400,0.4330918007994523)
            (4102400,0.4127071303168387)
            (3974400,0.3166810195626034)
            (3846400,0.374116357834976)
            (3718400,0.34460835554159674)
            (3590400,0.3883439798268396)
            (3462400,0.30879794000743144)
            (3334400,0.29844764567611537)
            (3206400,0.2324010947224276)
            (3078400,0.2805533988749895)
            (2950400,0.26136581050753416)
            (2822400,0.2303837913521567)
            (2694400,0.2265492575234539)
            (2566400,0.12957577873929055)
            (2438400,0.09044430970482746)
            (2310400,0.10805431167847891)
            (2182400,0.09676325013677795)
            (2054400,0.06199919279481063)
            (1926400,0.045571485762661096)
            (1798400,0.03182281291459493)
            (1670400,0.03287250751625828)
            (1542400,0.023350424859373685)
            (1414400,0.01793316496101822)
            (1286400,0.0)
            (1158400,0.0)
            (1030400,0.0)
            (902400,0.0)
            (774400,0.0)
            (646400,0.0)
            (518400,0.0)
            (390400,0.0)
            (262400,0.0)
            (134400,0.0)
            (6400,0.0)
            (6400,0.0)
        }
        ;
    \addplot[color={rgb,1:red,0.2422;green,0.6433;blue,0.3044}, name path={ccbabee1-7093-4041-b859-b07d5cc475bd}, legend image code/.code={{
    \draw[fill={rgb,1:red,0.2422;green,0.6433;blue,0.3044}, fill opacity={0.1}] (0cm,0cm) rectangle (0.3cm,0cm);
    }}, draw opacity={1.0}, line width={2}, solid]
        table[row sep={\\}]
        {
            \\
            6400.0  0.0  \\
            134400.0  0.0  \\
            262400.0  0.0  \\
            390400.0  0.0  \\
            518400.0  0.0  \\
            646400.0  0.0  \\
            774400.0  0.0  \\
            902400.0  0.0  \\
            1.0304e6  0.0  \\
            1.1584e6  0.0  \\
            1.2864e6  0.0  \\
            1.4144e6  0.009375  \\
            1.5424e6  0.009375  \\
            1.6704e6  0.0125  \\
            1.7984e6  0.01875  \\
            1.9264e6  0.021875  \\
            2.0544e6  0.028125  \\
            2.1824e6  0.053125  \\
            2.3104e6  0.0625  \\
            2.4384e6  0.05  \\
            2.5664e6  0.075  \\
            2.6944e6  0.134375  \\
            2.8224e6  0.140625  \\
            2.9504e6  0.15  \\
            3.0784e6  0.16875  \\
            3.2064e6  0.140625  \\
            3.3344e6  0.178125  \\
            3.4624e6  0.190625  \\
            3.5904e6  0.2375  \\
            3.7184e6  0.221875  \\
            3.8464e6  0.228125  \\
            3.9744e6  0.184375  \\
            4.1024e6  0.259375  \\
            4.2304e6  0.2625  \\
            4.3584e6  0.303125  \\
            4.4864e6  0.284375  \\
            4.6144e6  0.334375  \\
            4.7424e6  0.30625  \\
            4.8704e6  0.290625  \\
            4.9984e6  0.359375  \\
            5.1264e6  0.371875  \\
            5.2544e6  0.371875  \\
            5.3824e6  0.40625  \\
            5.5104e6  0.3375  \\
            5.6384e6  0.425  \\
            5.7664e6  0.403125  \\
            5.8944e6  0.40625  \\
            6.0224e6  0.403125  \\
            6.1504e6  0.35  \\
            6.2784e6  0.41875  \\
            6.4064e6  0.371875  \\
            6.5344e6  0.43125  \\
            6.6624e6  0.44375  \\
            6.7904e6  0.43125  \\
            6.9184e6  0.440625  \\
            7.0464e6  0.4875  \\
            7.1744e6  0.46875  \\
            7.3024e6  0.465625  \\
            7.4304e6  0.43125  \\
            7.5584e6  0.475  \\
            7.6864e6  0.4875  \\
            7.8144e6  0.5625  \\
            7.9424e6  0.4625  \\
            8.0704e6  0.490625  \\
            8.1984e6  0.509375  \\
            8.3264e6  0.475  \\
            8.4544e6  0.48125  \\
            8.5824e6  0.515625  \\
            8.7104e6  0.5375  \\
            8.8384e6  0.43125  \\
            8.9664e6  0.5125  \\
            9.0944e6  0.515625  \\
            9.2224e6  0.50625  \\
            9.3504e6  0.51875  \\
            9.4784e6  0.559375  \\
            9.6064e6  0.434375  \\
            9.7344e6  0.528125  \\
            9.8624e6  0.4875  \\
            9.9904e6  0.44375  \\
            1.00032e7  0.49375  \\
        }
        ;
    \addlegendentry {$\mathcal{S}^{3+}$}
\end{axis}
\end{tikzpicture}